%% file: iclr2025_conference.tex
\documentclass{article} % For LaTeX2e
\usepackage{iclr2025_conference,times}

\input{math_commands.tex}

\usepackage[utf8]{inputenc} % allow utf-8 input
\usepackage[T1]{fontenc}    % use 8-bit T1 fonts
\usepackage{hyperref}       % hyperlinks
\usepackage{url}            % simple URL typesetting
\usepackage{xurl}
\usepackage{booktabs}       % professional-quality tables
\usepackage{amsfonts}       % blackboard math symbols
\usepackage{nicefrac}       % compact symbols for 1/2, etc.
\usepackage{microtype}      % microtypography
\usepackage{xcolor}         % colors
\usepackage{graphicx}
\usepackage{subcaption}
\usepackage{lipsum}
\usepackage{multirow}
\usepackage[most]{tcolorbox}
\usepackage{colortbl}
\usepackage{tabularx}
\usepackage{authblk}
\usepackage{fontawesome5}

\hypersetup{
    colorlinks=true,
    linkcolor=red,
    citecolor=blue,
    filecolor=magenta,      
    urlcolor=blue,
linktocpage}
\newcommand{\lmj}{LLM-as-a-Judge }
\newcommand{\OURS}{\textbf{J1-7B} }

\title{\raisebox{-0.3\height}{\includegraphics[height=2em]{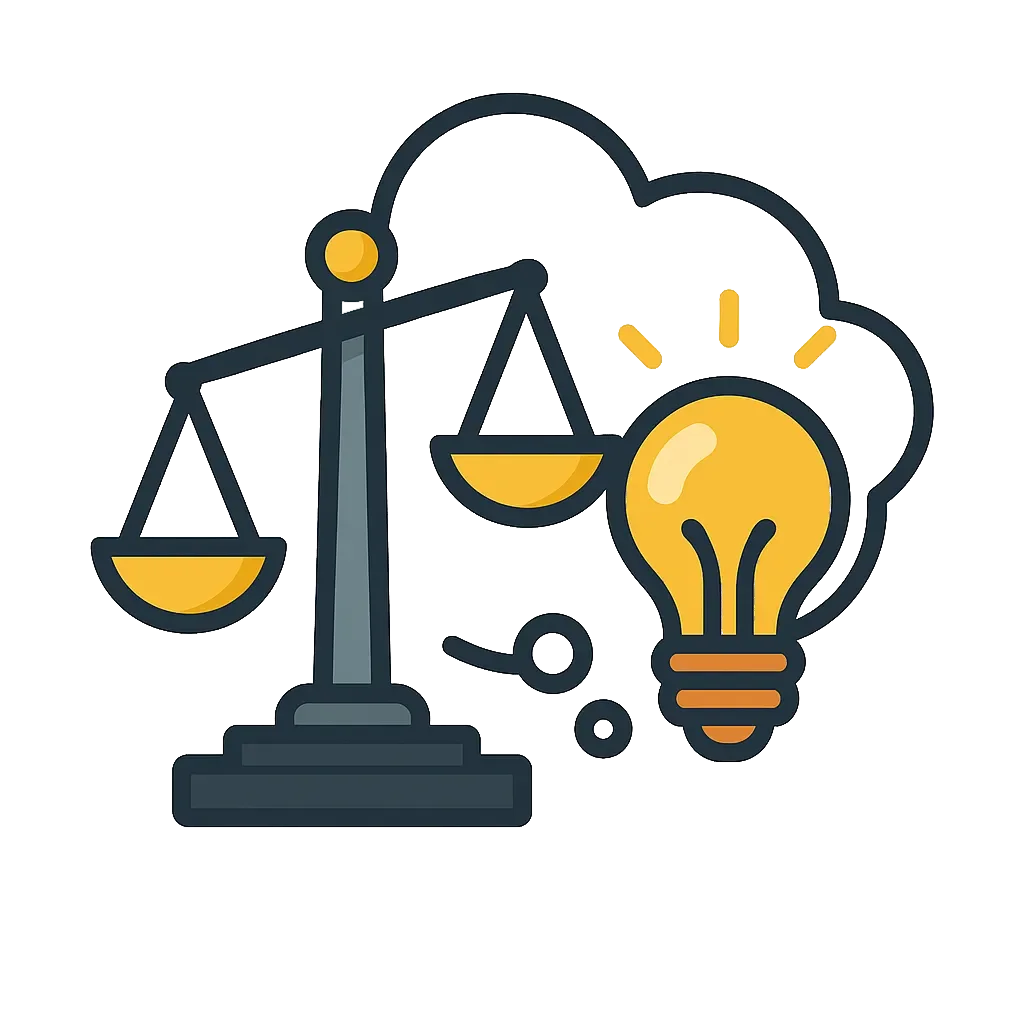}}~J1: Exploring Simple Test-Time Scaling for LLM-as-a-Judge}

\author{%
Chi-Min Chan$^{1}$, 
Chunpu Xu$^{3}$, 
Jiaming Ji$^{2}$, 
Zhen Ye$^{1}$, 
Pengcheng Wen$^{1}$, 
Chunyang Jiang$^{1}$, 
Yaodong Yang$^{2}$, 
Wei Xue$^{1}$, 
Sirui Han$^{1}$\thanks{Co-corresponding author}~~, 
Yike Guo$^{1}$\thanks{Corresponding author} \\
\small $^{1}$Hong Kong University of Science and Technology\\
\small $^{2}$Peking University\\
\small $^{3}$Hong Kong Polytechnic University\\
\texttt{\href{mailto:cchanbc@connect.ust.hk}{cchanbc@connect.ust.hk}}
}

\iclrfinalcopy % Uncomment for camera-ready version, but NOT for submission.
\begin{document}

\maketitle

\begin{abstract}
    The current focus of AI research is shifting from emphasizing model training towards enhancing evaluation quality, a transition that is crucial for driving further advancements in AI systems. Traditional evaluation methods typically rely on reward models assigning scalar preference scores to outputs. Although effective, such approaches lack interpretability, leaving users often uncertain about why a reward model rates a particular response as high or low. The advent of LLM-as-a-Judge provides a more scalable and interpretable method of supervision, offering insights into the decision-making process. Moreover, with the emergence of large reasoning models, which consume more tokens for deeper thinking and answer refinement, scaling test-time computation in the LLM-as-a-Judge paradigm presents an avenue for further boosting performance and providing more interpretability through reasoning traces.
    In this paper, we introduce $\textbf{J1-7B}$, which is first supervised fine-tuned on reflection-enhanced datasets collected via rejection-sampling and subsequently trained using Reinforcement Learning (RL) with verifiable rewards. At inference time, we apply Simple Test-Time Scaling (STTS) strategies for additional performance improvement. Experimental results demonstrate that $\textbf{J1-7B}$ surpasses the previous state-of-the-art LLM-as-a-Judge by $ \textbf{4.8}$\% and exhibits a $ \textbf{5.1}$\% stronger scaling trend under STTS.
    Additionally, we present three key findings: (1) Existing LLM-as-a-Judge does not inherently exhibit such scaling trend. (2) Model simply fine-tuned on reflection-enhanced datasets continues to demonstrate similarly weak scaling behavior. (3) Significant scaling trend emerges primarily during the RL phase, suggesting that effective STTS capability is acquired predominantly through RL training. We hope that our comprehensive analysis will provide insights to the community and contribute to the broader AI alignment community, especially in the context of developing more robust and scalable evaluation systems.
\end{abstract}

\section{Introduction}

\input{sections/introduction}

\label{sec:introduction}

\section{Related Works}
\input{sections/related_works}

\section{Simple Test-Time Scaling for \lmj}

\input{sections/methodology}

\section{Experiments}
\label{sec:experiments}
\input{sections/experiments}

\section{Conclusion}

In this paper, we investigate the effectiveness of STTS techniques for enhancing the capabilities and reliability of \lmj systems. To this end, we propose \textbf{J1-7B}, a novel \lmj trained via a two-stage paradigm, combining SFT on a reflection-enhanced dataset and RL with verifiable rewards. Our experiments demonstrate that existing \lmj methods, trained solely on general judgment datasets, exhibit limited scaling trends under STTS. In contrast, our approach significantly improves both baseline performance and STTS effectiveness, surpassing previous state-of-the-art \lmj by 4.8\% in performance and exhibiting a 5.1\% stronger scaling trend. Furthermore, detailed experiments reveal that models learn STTS during the RL process. These findings collectively suggest a promising research direction in developing more reliable, and scalable \lmj systems through structured reflective reasoning strategies.

\bibliography{iclr2025_conference}
\bibliographystyle{iclr2025_conference}

\clearpage

\appendix
\input{sections/appendix}

\end{document}

%% file: math_commands.tex
\usepackage{amsmath,amsfonts,bm}

\def\eqref#1{equation~\ref{#1}}
\def\1{\bm{1}}

\DeclareMathAlphabet{\mathsfit}{\encodingdefault}{\sfdefault}{m}{sl}
\SetMathAlphabet{\mathsfit}{bold}{\encodingdefault}{\sfdefault}{bx}{n}

%% file: sections/introduction.tex
The evaluation of AI models is fundamental to their success~\citep{hoffman2018metrics}, especially within the context of Reinforcement Learning from Human Feedback (RLHF)~\citep{ouyang2022training}, where accurately assessing model outputs directly guides policy improvement. The ability to assess model outputs effectively is also critical to ensuring their reliability, safety, and alignment with human goals. However, traditional evaluation methods, such as scalar reward models~\citep{gao2023scaling, bradley1952rank}, often lack interpretability and transparency, which limits their ability to explain why certain responses are deemed better than others. This lack of clarity can hinder trust in AI systems and reduce their overall effectiveness in real-world applications.

\begin{figure}[t]
  \centering
  \includegraphics[width=0.9\textwidth]{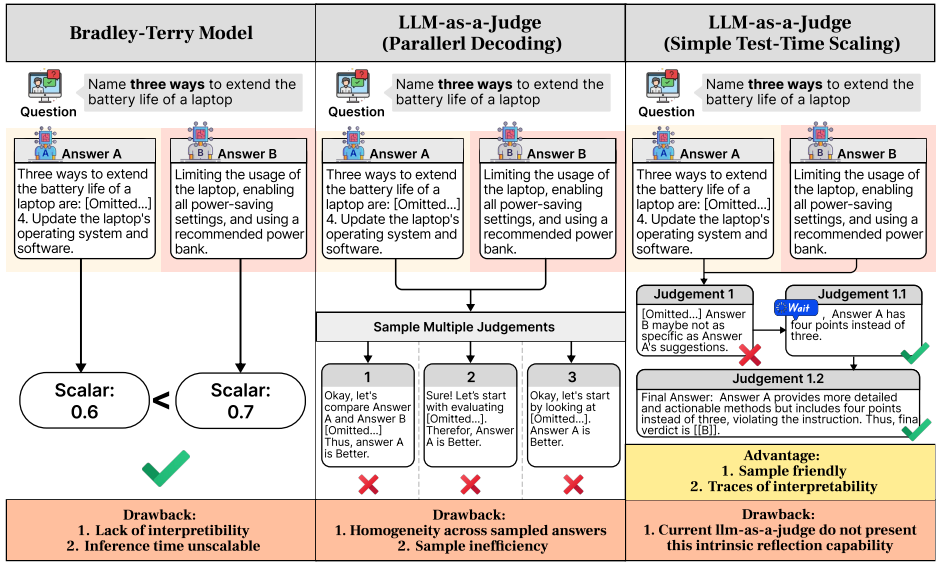}
  \caption{Comparison of Bradley-Terry model and \lmj under different scaling strategies.}
  \label{fig:teaser}
\end{figure}

In response to these challenges, the "LLM-as-a-Judge" paradigm has gained prominence~\citep{gu2024survey, li2024generation}. This approach leverages the capabilities of LLMs themselves to evaluate the outputs of other AI systems, often other LLMs. The rationale is compelling: LLMs, particularly those fine-tuned with methods like RLHF, demonstrate a strong ability to understand instructions and align with human preferences~\citep{zheng2023judging}, making them plausible candidates for automated evaluation. This method offers potential advantages in terms of speed, cost-efficiency, scalability, and adaptability compared to purely human-based assessment~\citep{hosking2023human}.

However, the \lmj approach is not a panacea. Ensuring the reliability and trustworthiness of \lmj systems remains a significant open challenge. Specifically, \lmj also tends to struggle when evaluating tasks requiring complex reasoning, such as mathematics or coding, particularly if the judge model itself lacks proficiency in those areas~\citep{kim2024evaluating, yu2025improve}. 
With the advent of large reasoning models~\citep{openai_learning_to_reason_2024, guo2025deepseek, team2025kimi}, a common paradigm for enhancing the reasoning capabilities of LLMs involves test-time scaling~\citep{snell2024scaling}. Existing methods include Best-of-N sampling, which relies on external validators to select the optimal response~\citep{lightman2023let, wu2024inference}, and sequential refine, which iteratively revises outputs by extending Chain-of-Thought(COT) reasoning with reflective steps~\citep{madaan2023self, openai_learning_to_reason_2024, muennighoff2025s1}. Nevertheless, these test-time scaling techniques have primarily focused on solving tasks, leaving their effectiveness in evaluation scenarios underexplored. This gap naturally leads to a scientific question: \textit{Can applying test-time scaling techniques to \lmj enhance its quality and reliability?}

To address the above question, we introduce \textbf{J1-7B}, an \lmj trained through a combination of Supervised Fine-Tuning (SFT) and Reinforcement Learning (RL), specifically optimized to benefit from Simple Test-Time Scaling (STTS)~\citep{muennighoff2025s1}, where we append thinking tokens like "Wait," multiple times to enforce model to think longer before providing final answers. We provide more discussion between parallel decoding and STTS in Appendix~\ref{appendix:comparison_between_pd_stts}. In contrast to previous approaches~\citep{yu2025improve, skyworkcritic2024} that train \lmj solely on general judgment datasets—resulting in limited scaling behavior under STTS, our method introduces a two-stage training paradigm that effectively enhances reflective reasoning. Specifically, we first curate a reflection-enhanced dataset augmented explicitly by STTS tokens through rejection sampling, enabling an effective cold-start initialization that teaches the model how to utilize reflective reasoning tokens optimally. Subsequently, we apply RL with verifiable reward to empower the model to autonomously refine and optimize its reflective capabilities. As a result of this combined training strategy, our model, \textbf{J1-7B}, achieves notable performance improvements and demonstrates an enhanced scaling trend under STTS during inference.

To sum up, the key contribution of our paper are as follows:
\begin{itemize}
    \item[$\bullet$] We identify that existing \lmj models trained solely on general judgment datasets exhibit limited scaling behavior under STTS, highlighting an important yet underexplored limitation in current frameworks.
    \item[$\bullet$] We introduce a novel two-stage training paradigm, starting with a reflection-enhanced SFT dataset curated through rejection sampling, effectively initializing the \lmj to optimally leverage reflective reasoning tokens introduced by STTS.
    \item[$\bullet$] We demonstrate that further enhancements in scaling behavior emerge explicitly during the RL stage, indicating that the capability for effective reflective reasoning under STTS is actively learned and reinforced through RL process.
    \item[$\bullet$] Empirically, we validate our proposed method by introducing \textbf{J1-7B} that surpasses previous state-of-the-art baselines. Specifically, \textbf{J1-7B} achieves a \textbf{4.8\%} improvement in overall judgment performance and exhibits a \textbf{5.1\%} stronger scaling trend under STTS, confirming the effectiveness and robustness of our proposed training methodology.
\end{itemize}

%% file: sections/related_works.tex
\subsection{LLM-As-A-Judge} 

The evaluation of Large Language Models (LLMs), particularly for complex, open-ended tasks, presents significant challenges, as traditional metrics often fall short and human feedback, while considered a standard, suffers from scalability issues, high costs, and inherent biases~\citep{kirk2023past}. This has spurred the development of the "LLM-as-a-judge" paradigm, which utilizes LLMs themselves as evaluators to offer a scalable and cost-effective alternative~\citep{hosking2023human, gu2024survey, li2024generation}. Research in this domain encompasses various approaches, including the development of fine-tuned judges~\citep{zhu2023judgelm, wang2023pandalm, li2023generative}, the creation of generative or reasoning-based judges that move beyond simple scores by incorporating critiques~\citep{ankner2024critique, zhang2024generative, yu2024self, ye2024improving} or structured evaluation plans~\citep{saha2025learning}, and efforts to enhance judging as a general LLM capability~\citep{yu2025improve}.  Furthermore, advanced applications are being explored, such as employing LLMs as critics for iterative refinement and bug detection~\citep{mcaleese2024llm}, enabling self-improvement through meta-judging~\citep{wu2024meta}, and facilitating scalable oversight where weaker judges evaluate stronger models~\citep{kenton2024scalable}.

\subsection{Test-Time Scaling}

Recent advances in scaling inference-time computation, such as OpenAI's o1/o3 models~\citep{openai_learning_to_reason_2024}, have demonstrated the benefits of generating additional intermediate tokens before producing a final answer to enhance reasoning capabilities. This approach has inspired several community-driven replications, including Deepseek R1~\citep{guo2025deepseek}, Kimi 1.5~\citep{team2025kimi}, and other related efforts~\citep{wang2024openr, zhang2024llama}. In contrast to sequential refinement strategies for inference-time scaling, parallel decoding techniques~\citep{snell2024scaling, wu2024inference, brown2024large} offer an alternative paradigm. By generating multiple candidate sequences in parallel—via repetitive attempts or value-guided search—these methods improve performance given higher computational budgets, with proven efficacy in various domains~\citep{zhangopenprm, chan2024rq, ehrlich2025codemonkeys, liu2025video, xu2024llava, liu2025inference, ye2025llasa}.

In recent developments, a more lightweight approach for inference-time scaling has emerged, known as ``simple test-time scaling.''~\citep{muennighoff2025s1}. This method builds upon large reasoning models and leverages the inherent reasoning capabilities of existing models. During the testing phase, it forces the model to engage in additional reflection by incorporating special tokens such as "wait" during the model's reasoning process. This method demonstrates superior scaling effects compared to repetitive sampling approaches. Notable works in this area have sparked considerable interest within the research community, showing great potential. However, current studies have primarily focused on mathematical and code-related tasks~\citep{yu2025z1, shah2025rethinking}. Our work fits within this paradigm and extends its application to the realm of LLM-as-a-judge, specifically in the context of alignment. We believe this approach represents a key step toward achieving scalable oversight in AI systems. Furthermore, it offers a more interpretable evaluation framework, as it traces the model's thought process and how its reasoning evolves, providing valuable insight into the underlying decision-making process.

%% file: sections/methodology.tex
\begin{figure}[t]
  \centering
  \includegraphics[width=\textwidth]{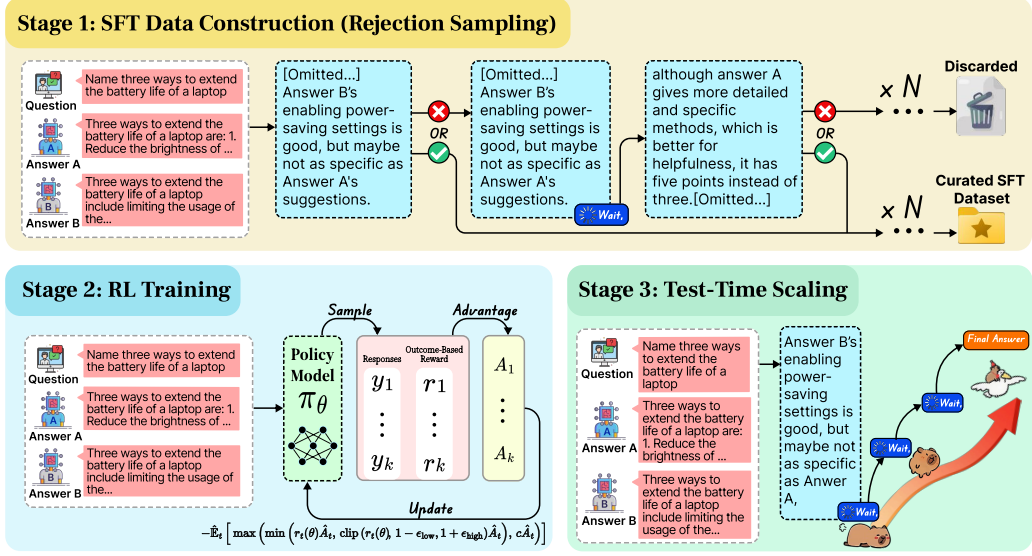}
  \caption{\textbf{Pipeline of J1-7B.} We first curate the SFT dataset through rejection sampling and subsequently apply RL training to integrate STTS capabilities into \OURS.}
  \label{fig:methodology}
\end{figure}

In this section, we outline our methodological framework in detail. In Section~\ref{sec:overview}, we introduce key concepts, including task definitions and comparison between traditional reward models and \lmj.
In Section~\ref{sec:sft_data}, we describe the collection of the SFT dataset via rejection sampling; In Section~\ref{sec:rl_training}, we detail our RL training approach using a verifiable reward signal; In Section~\ref{sec:test_time_scaling}, we introduce the budget-forcing mechanism designed for STTS.

\subsection{Overview}
\label{sec:overview}

The core objective of our work is to improve the capability of an \lmj in distinguishing the quality of responses. Formally, the task consists of a query \( q \), two candidate responses \( a_1 \) and \( a_2 \) generated by different models, and a corresponding human-provided preference \( r_{\text{true}} \). Specifically, if the first response \( a_1 \) is preferred, \( r_{\text{true}} = 0 \); conversely, if the second response \( a_2 \) is preferred, \( r_{\text{true}} = 1 \).

Traditionally, a reward model predicts scalar scores for each response by evaluating the pairs \((q, a_1)\) and \((q, a_2)\) separately, denoted as \( r_{\text{predict}}^{(1)} \) and \( r_{\text{predict}}^{(2)} \), respectively. The predicted preference \( r_{\text{predict}} \) is determined as:
\begin{equation}
    r_{\text{predict}} = 
    \begin{cases}
        0, & \text{if } r_{\text{predict}}^{(1)} > r_{\text{predict}}^{(2)}\\[6pt]
        1, & \text{otherwise}
    \end{cases}
\end{equation}

In contrast, an \lmj typically takes pair-wise input and are prompterd to directly outputs a preference decision, formally represented as:

\begin{equation}
    r_{\text{predict}} = \text{LLM}(q, a_1, a_2, c), \quad r_{\text{predict}} \in \{0, 1\}
\end{equation}

where $c$ is the prompt template that ask LLM to chose the prefered response and formulate the output in the corresponding format. The detailed templates for the pair-wise input method are provided in Appendix~\ref{appendix:prompt_template}.
The accuracy metric \( \text{acc}(r_{\text{true}}, r_{\text{predict}}) \) are used to quantifies the model's effectiveness; higher accuracy indicates a greater ability to distinguish between responses.

\subsection{Supervised-Finetuning Dataset Curation via Rejection Sampling}
\label{sec:sft_data}

To endow the model with foundational reasoning capabilities, a supervised dataset for cold-start training is essential. Specifically, our initial dataset is curated from publicly available sources, including HelpSteer2~\citep{wang2024helpsteer2}, Offsetbias~\citep{park2024offsetbias}, Wildguard~\citep{han2024wildguard} and Magpie~\citep{xu2024magpie}. To augment this dataset with intermediate reasoning steps, we utilize Deepseek-R1~\citep{guo2025deepseek}, a powerful, open-source reasoning model, to generate intermediate thought processes along with the final answers.

To ensure the quality of the collected reasoning trajectories, we adopt a \textit{rejection sampling} strategy: we retain only those trajectories whose final answers align with the provided correct answers in the original dataset. Specifically, for trajectories that initially yield incorrect final answers, we utilize the method as specified in Section~\ref{sec:test_time_scaling} to enforce the model to think again. This reflection procedure is iteratively performed for three cycles to enrich our dataset with reflective reasoning patterns.

The detailed workflow for this iterative reflective data collection process is illustrated in the upper part of Figure~\ref{fig:methodology}. The statistical breakdown of the dataset after each filtering and reflection step is presented in Appendix~\ref{appendix:training_dataset_statistics}.

\subsection{Reinforcement Learning}
\label{sec:rl_training}

Following the SFT cold-start phase, we further optimize the LLM using RL, guided by an outcome-based reward strategy. Specifically, we define the reward as follows:
\begin{equation}
    \text{reward} = \mathbb{I}(r_{\text{predict}} = r_{\text{true}}),
\end{equation}
where $\mathbb{I}(\cdot)$ is the indicator function. This reward structure assigns a reward of 1 when the model's predicted preference aligns with the human ground-truth preference, and 0 otherwise.

We explore several policy gradient algorithms during RL optimization, including Proximal Policy Optimization (PPO)~\citep{schulman2017proximal}, Reinforce++~\citep{hu2025reinforce++}, and Group Relative Policy Optimization (GRPO)~\citep{shao2024deepseekmath}. Additionally, we employ a dual-clip PPO objective~\citep{ye2020mastering} to stabilize RL training. The objective function is given by:

\begin{equation}
    L_t^{\text{clip}}(\theta) = -\hat{\mathbb{E}}_t\left[
        \max\left(
            \min\left(
                r_t(\theta)\hat{A}_t,\,
                \text{clip}\left(r_t(\theta), 1-\epsilon_{\text{low}}, 1+\epsilon_{\text{high}}\right)\hat{A}_t
            \right),\,
            c\hat{A}_t
        \right)\right],
\end{equation}

where $c > 1$ is a constant indicating the lower bound, $\hat{A}_t$ is the advantage estimate at token $t$, and $(\epsilon_{\text{low}}, \epsilon_{\text{high}})$ represent clipping thresholds, $r_t(\theta)$ is policy ratio for token $y_t$ given context $x_t$, and is computed as:

\begin{equation}
    r_t(\theta) = \frac{\pi_{\theta}(y_t|x_t)}{\pi_{\theta_{\text{old}}}(y_t|x_t)}
    = \exp\left(\log \pi_{\theta}(y_t|x_t) - \log \pi_{\theta_{\text{old}}}(y_t|x_t)\right),
\end{equation}

where $\pi_{\theta}$ is the current policy parameterized by $\theta$, and $\pi_{\theta_{\text{old}}}$ denote the previous policy. We provide details of different RL algorithms and their hyperparameters in Appendix~\ref{appendix:training_details}.

By default, we utilize the english DPO subset from the RISE dataset~\citep{yu2025improve} in the RL phase. We specifically employ the $(q, a_1, a_2, r_{\text{true}})$ tuples, excluding the generated CoT trajectories provided by the original dataset. In Appendix~\ref{appendix:different_data_mixture}, we further investigate the impact of employing various data mixtures during the RL optimization phase.

\subsection{Simple Test-Time Scaling}
\label{sec:test_time_scaling}

During the inference stage, we leverage STTS~\citep{muennighoff2025s1} to further enhance the model's performance. More concretely, current reasoning models typically enclose their intermediate thinking processes within special tokens, such as \texttt{<think>} and \texttt{</think>}, generating the final response only after the closing token (\texttt{</think>}). Benefiting from this structured generation pattern, we can explicitly detect when a model completes its initial reasoning phase. Subsequently, instead of allowing the model to finalize its answer immediately upon encountering the \texttt{</think>} token, we replace this token with an additional reflective prompt, such as \texttt{``wait''}, thereby encouraging the model to engage in further reflection before providing its ultimate decision.

Formally, the reflective extension at inference time can be represented as follows:
\begin{equation}
    r_{predict} = \text{LLM}(q, a_1, a_2,c, \{\texttt{<think>}, \dots, \texttt{``wait''}, \dots, \texttt{</think>}\}), \quad r_{\text{predict}} \in \{0, 1\}
\end{equation}

The workflow of this process is illustrated on the right most panel of Figure~\ref{fig:teaser}. This simple yet effective scaling approach has garnered attention in recent literature due to its ease of implementation~\citep{muennighoff2025s1, yu2025z1} and computational efficiency compared to parallel sampling approaches.

%% file: sections/experiments.tex
In this section, we introduce the benchmark datasets in Section~\ref{subsec:benchmark}, followed by the description of the baselines in Section~\ref{subsec:baselines}. The subsequent sections aim to address the following research questions:

\begin{itemize}
    \item[$\bullet$] \textbf{RQ1:} How does \OURS perform compared to existing state-of-the-art models? (Section~\ref{subsec:overal_results})
    \item[$\bullet$] \textbf{RQ2:} How does \OURS perform under STTS settings? (Section~\ref{subsec:STTS})
    \item[$\bullet$] \textbf{RQ3:} What factors influence the performance of \OURS during STTS, such as the composition of cold-start dataset, the choice of RL algorithm, and the stage of training progression?(Section ~\ref{subsec:cold_start}, ~\ref{subsec:different_rl_algorithms},~\ref{subsec:step_increase})
\end{itemize}

We provide more experimental results and case study showcasing the reflective behaviors during STTS in Appendix~\ref{appendix:case_study} and~\ref{appendix:more_experimental_results},respectively.

\input{tables/main}

\subsection{Benchmark}
\label{subsec:benchmark}

We evaluate \OURS using four diverse preference datasets: \textit{Anthropic Harmless}~\citep{bai2022training}, \textit{RewardMath}~\citep{kim2024evaluating}, \textit{CodePrefBench}~\citep{liu2024learning} and \textit{RewardBench}~\citep{lambert2024rewardbench}. Collectively, these datasets comprehensively cover evaluation aspects including safety, mathematics, code, and open-domain queries. Detailed descriptions of these datasets can be found in the Appendix~\ref{appendix:benchmark}.

The primary metric used for evaluation across these datasets is accuracy between $r_\text{predict}$ and $r_\text{true}$. When performing STTS, we measure the relative improvement. Given that baseline models differ in their initial accuracy scores, we specifically focus on the relative improvement with respect to the remaining potential for improvement (i.e., the gap to 100\% accuracy), defined formally as:

\begin{equation}
\Delta~\text{Relative} \% = \frac{\text{Accuracy}_{\text{STTS}} - \text{Accuracy}_{\text{INIT}}}{1 - \text{Accuracy}_{\text{INIT}}} \times 100\% .
\end{equation}

\subsection{Baselines}
\label{subsec:baselines}

We benchmark \OURS against a diverse set of powerful closed-source and open-source \lmj as well as scalar reward models. Closed-source \lmj include \textit{GPT-4o}, \textit{o1-mini}~\citep{openai_learning_to_reason_2024}, and \textit{Gemini-2.0-flash}~\citep{google_gemini_2024}, while open-source \lmj encompass \textit{DeepSeek-R1}~\citep{guo2025deepseek}, \textit{LLaMA3.1-8B-Instruct}~\citep{grattafiori2024llama}, \textit{Qwen2.5-7B-Instruct}~\citep{yang2024qwen2}, as well as previous state-of-the-art \lmj \textit{Skywork-Critic}~\citep{skyworkcritic2024} and \textit{Rise-Judge}~\citep{yu2025improve}, which have shown competitive performance on the RewardBench leaderboard. For the scalar reward models, we compare with \textit{IntermLM2-7B-Reward}~\citep{cai2024internlm2}, \textit{Skywork-Reward-Llama3.1-8B}~\citep{liu2024skywork} and \textit{FsfairX-LLaMA3-8B}~\citep{xiong2024iterative}.

\subsection{Overal Results}
\label{subsec:overal_results}

\begin{figure}[t]
  \centering
  \resizebox{1.\textwidth}{!}{
    \begin{tabular}{cc}
      \begin{subfigure}[b]{0.48\textwidth}
        \centering
        \includegraphics[width=\linewidth]{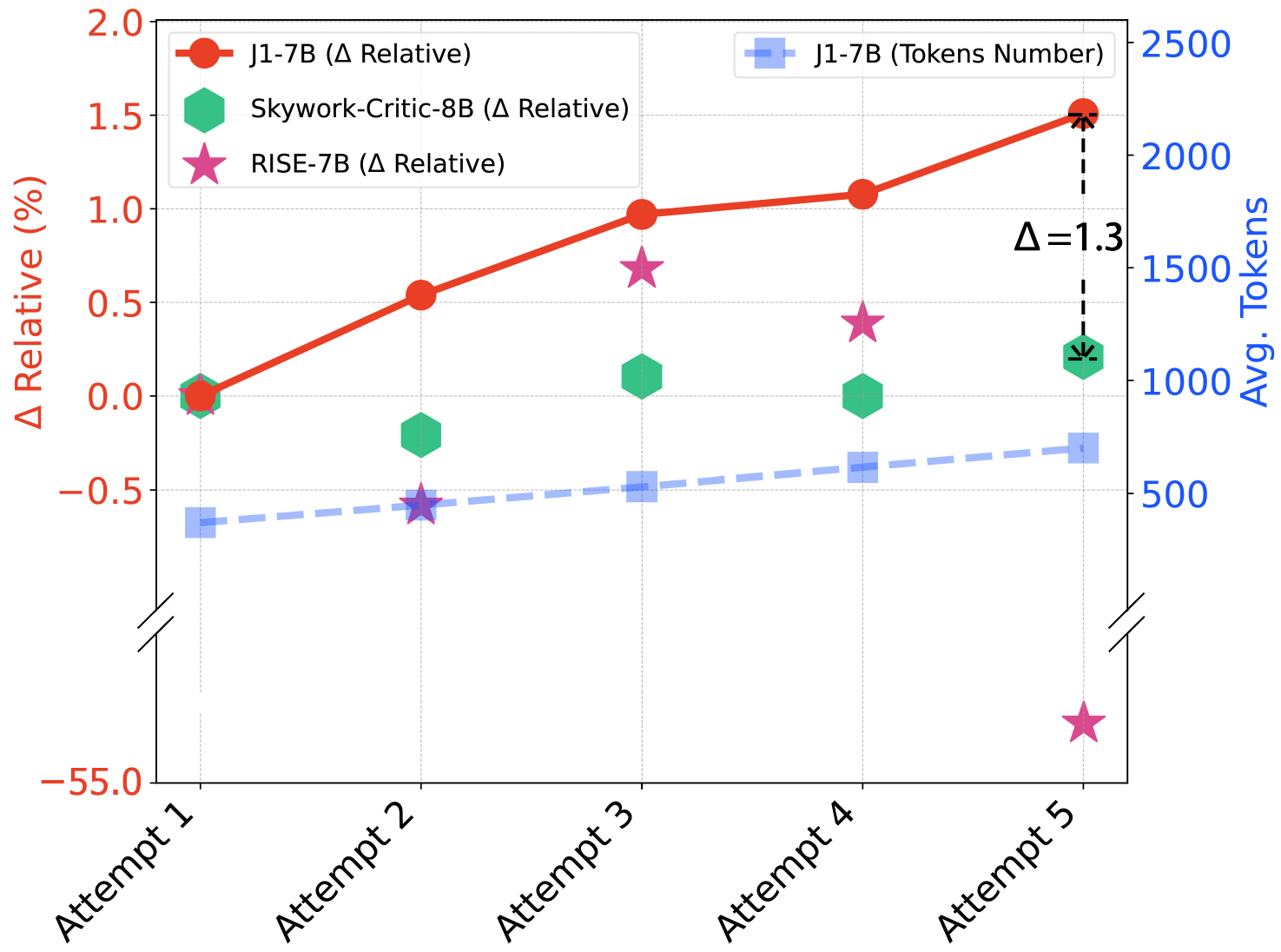}
        \caption{Anthropic Harmless}
        \label{fig:scaling_trend_sub1}
      \end{subfigure} &
      \begin{subfigure}[b]{0.48\textwidth}
        \centering
        \includegraphics[width=\linewidth]{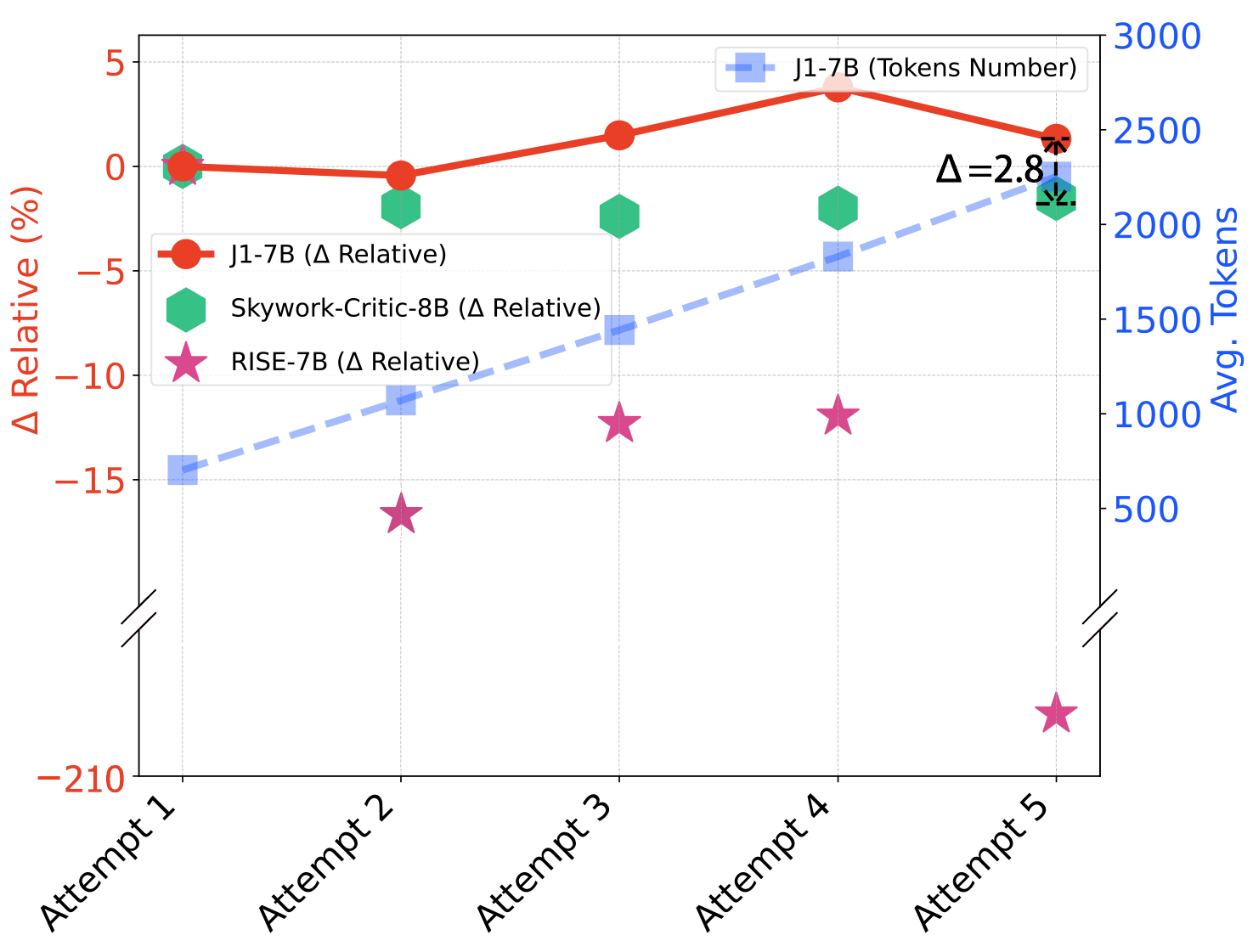}
        \caption{RewardBench}
        \label{fig:scaling_trend_sub2}
      \end{subfigure} \\[10pt]  % 增加行间距
      \begin{subfigure}[b]{0.48\textwidth}
        \centering
        \includegraphics[width=\linewidth]{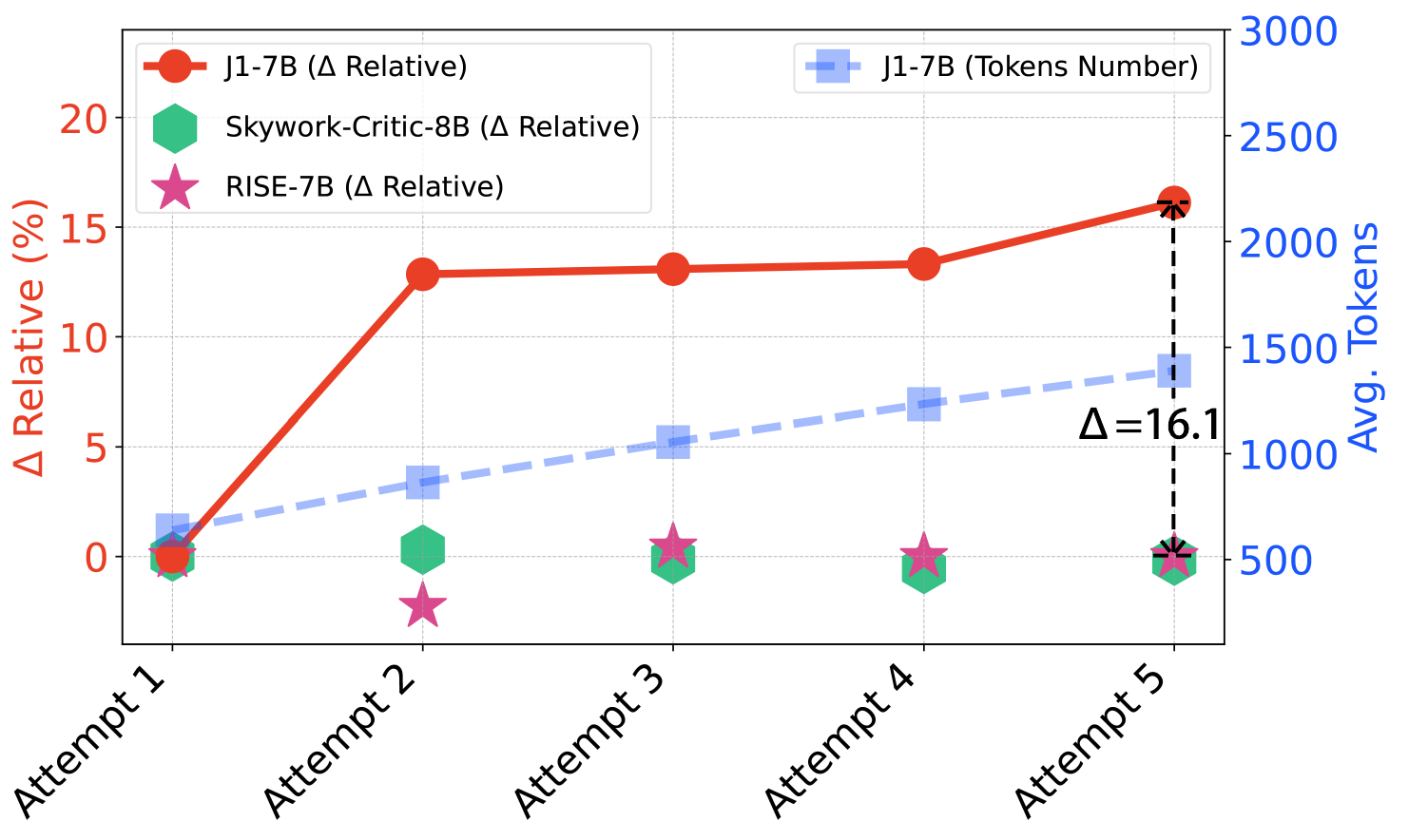}
        \caption{RewardMATH}
        \label{fig:scaling_trend_sub3}
      \end{subfigure} &
      \begin{subfigure}[b]{0.48\textwidth}
        \centering
        \includegraphics[width=\linewidth]{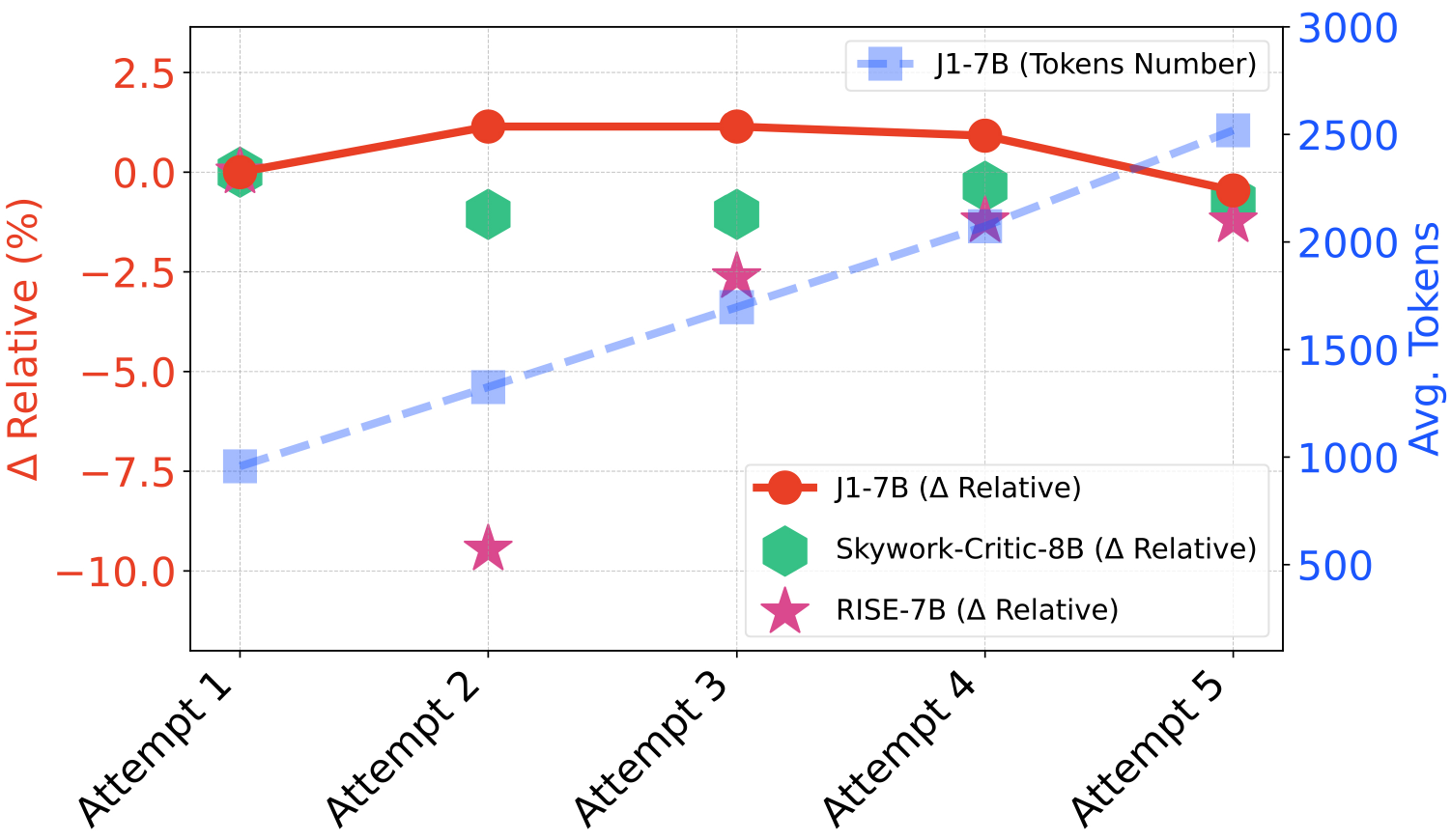}
        \caption{CodePrefBench}
        \label{fig:scaling_trend_sub4}
      \end{subfigure}
    \end{tabular}
  }
  \caption{Scaling trend for STTS on four different tasks. }
  
  \label{fig:scaling_trend}
\end{figure}

\begin{figure}[b]
  \centering

   \begin{subfigure}[b]{0.24\textwidth}
    \centering
    \includegraphics[width=\linewidth]{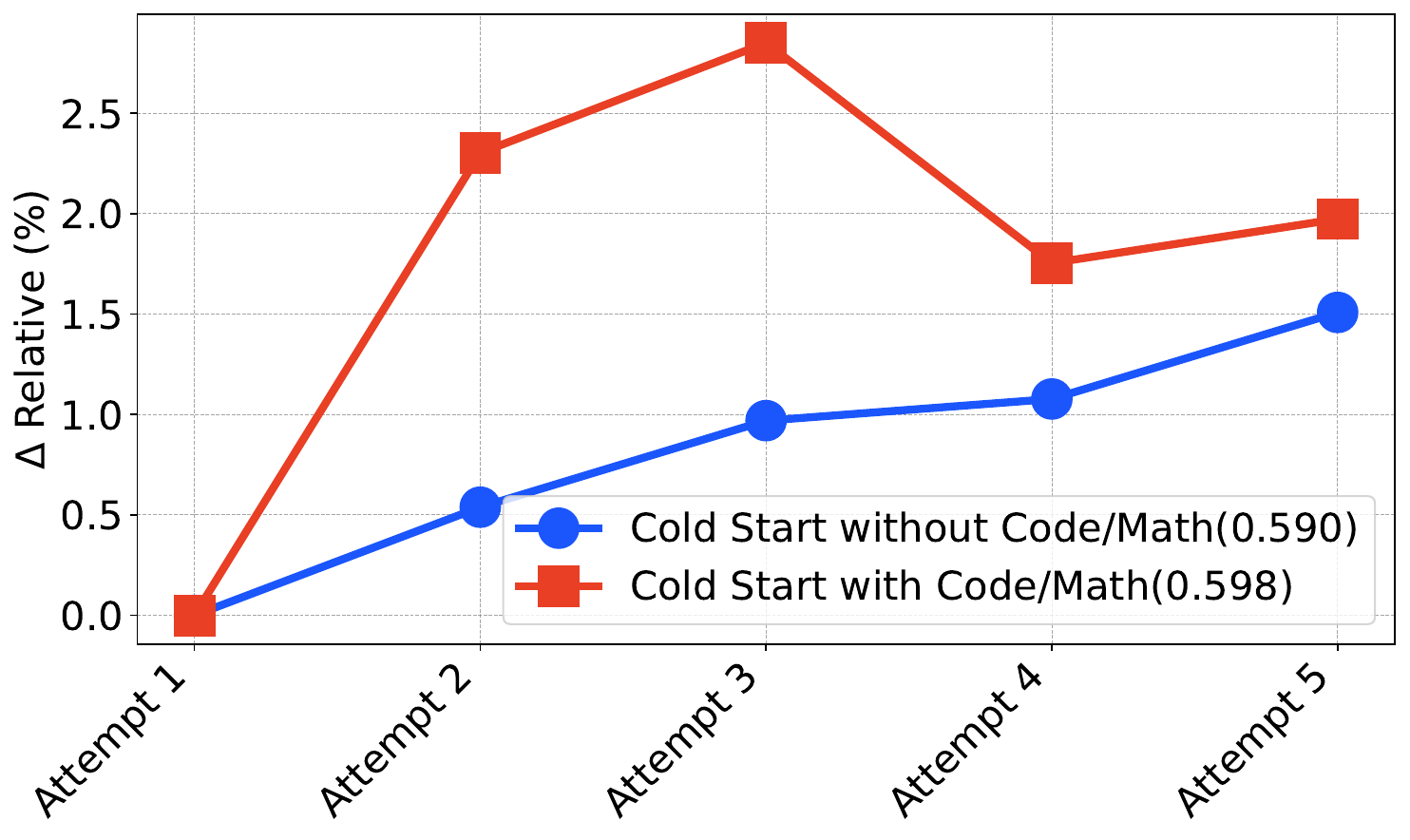}
    \caption{Anthropic Harmless}
    \label{fig:cold_start_anthropic_harmless}
  \end{subfigure}
  \hfill
  \begin{subfigure}[b]{0.24\textwidth}
    \centering
    \includegraphics[width=\linewidth]{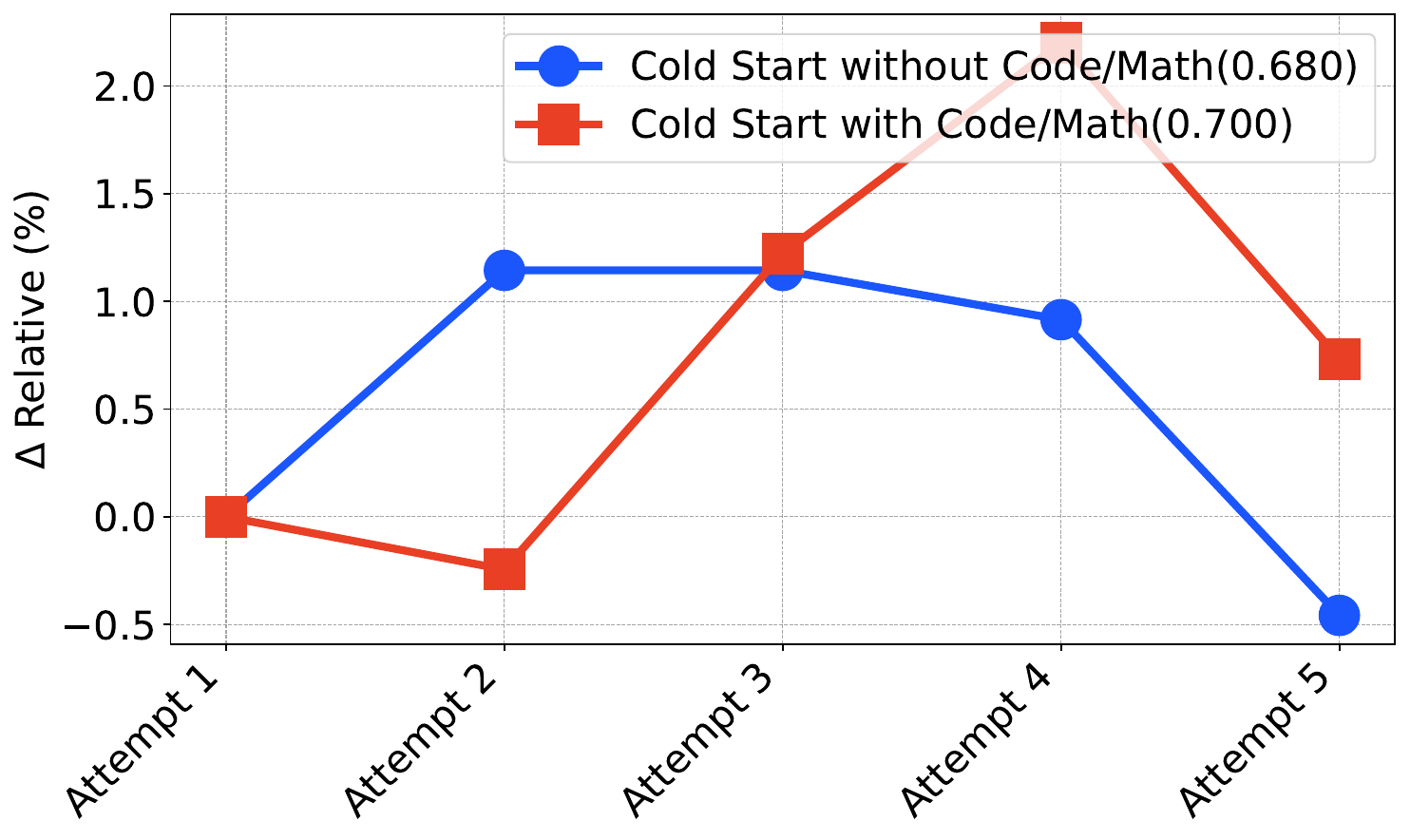}
    \caption{CodePrefBench}
    \label{fig:cold_start_codeprefbench}
  \end{subfigure}
  \begin{subfigure}[b]{0.24\textwidth}
    \centering
    \includegraphics[width=\linewidth]{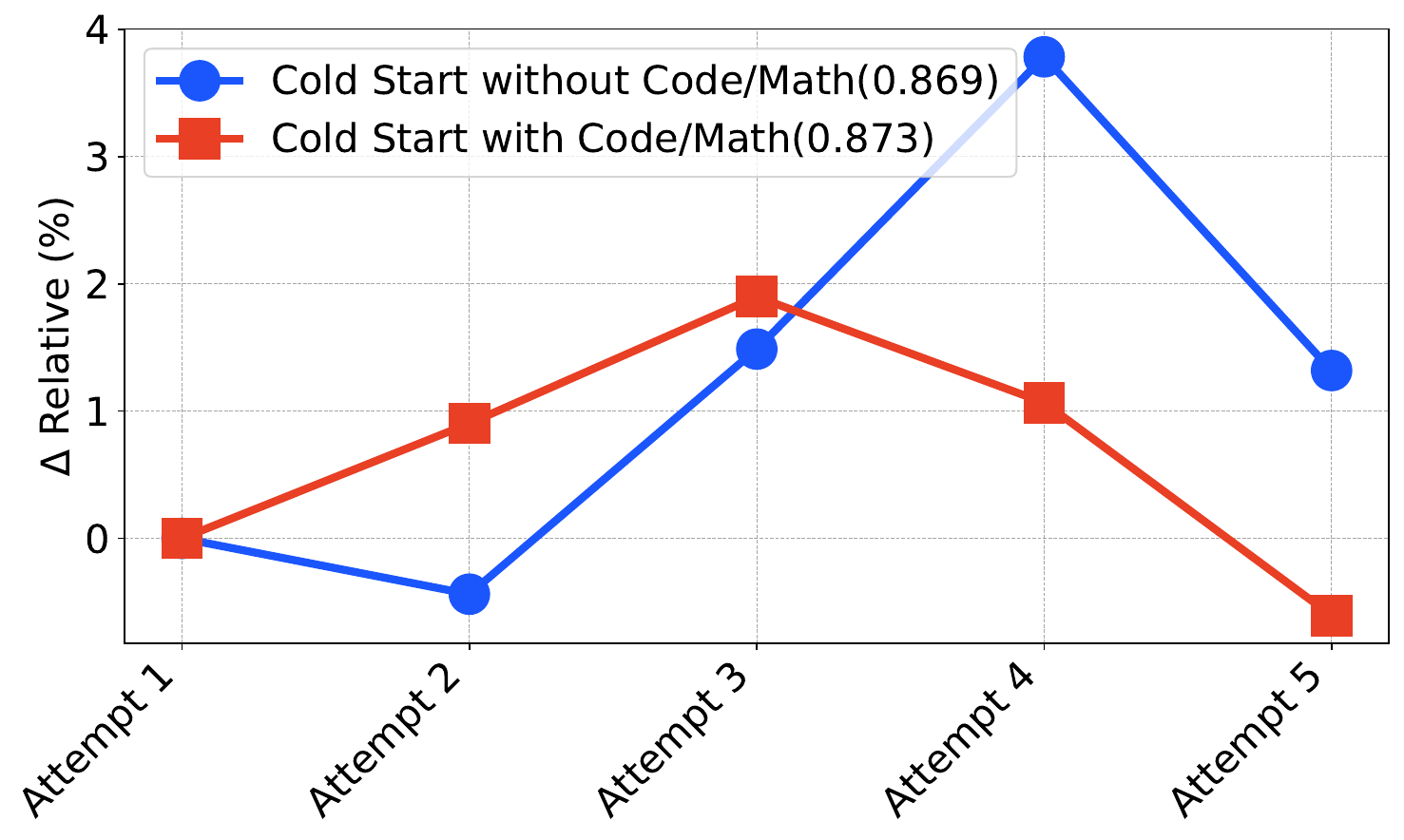}
    \caption{RewardBench}
    \label{fig:cold_start_rewardbench}
  \end{subfigure}
  \hfill
  \begin{subfigure}[b]{0.24\textwidth}
    \centering
    \includegraphics[width=\linewidth]{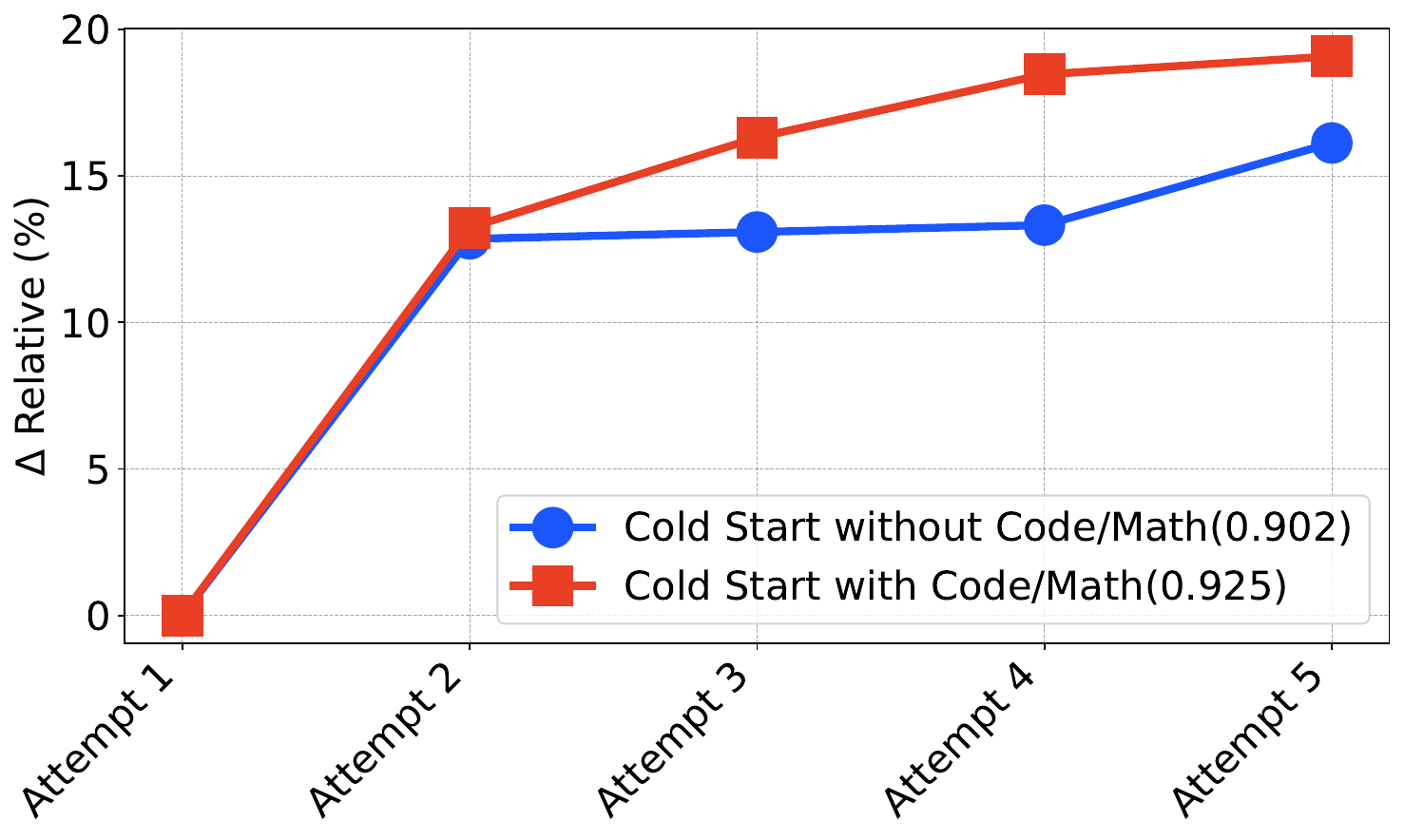}
    \caption{RewardMATH}
    \label{fig:cold_start_rewardmath}
  \end{subfigure}
  
  \caption{Cold start on reasoning-intensive data improves STTS.}
  \label{fig:algorithms_ablation/cold_start}
\end{figure}

\begin{figure}[t]
  \centering
  \begin{subfigure}[b]{0.48\textwidth}
    \centering
    \includegraphics[width=0.9\linewidth]{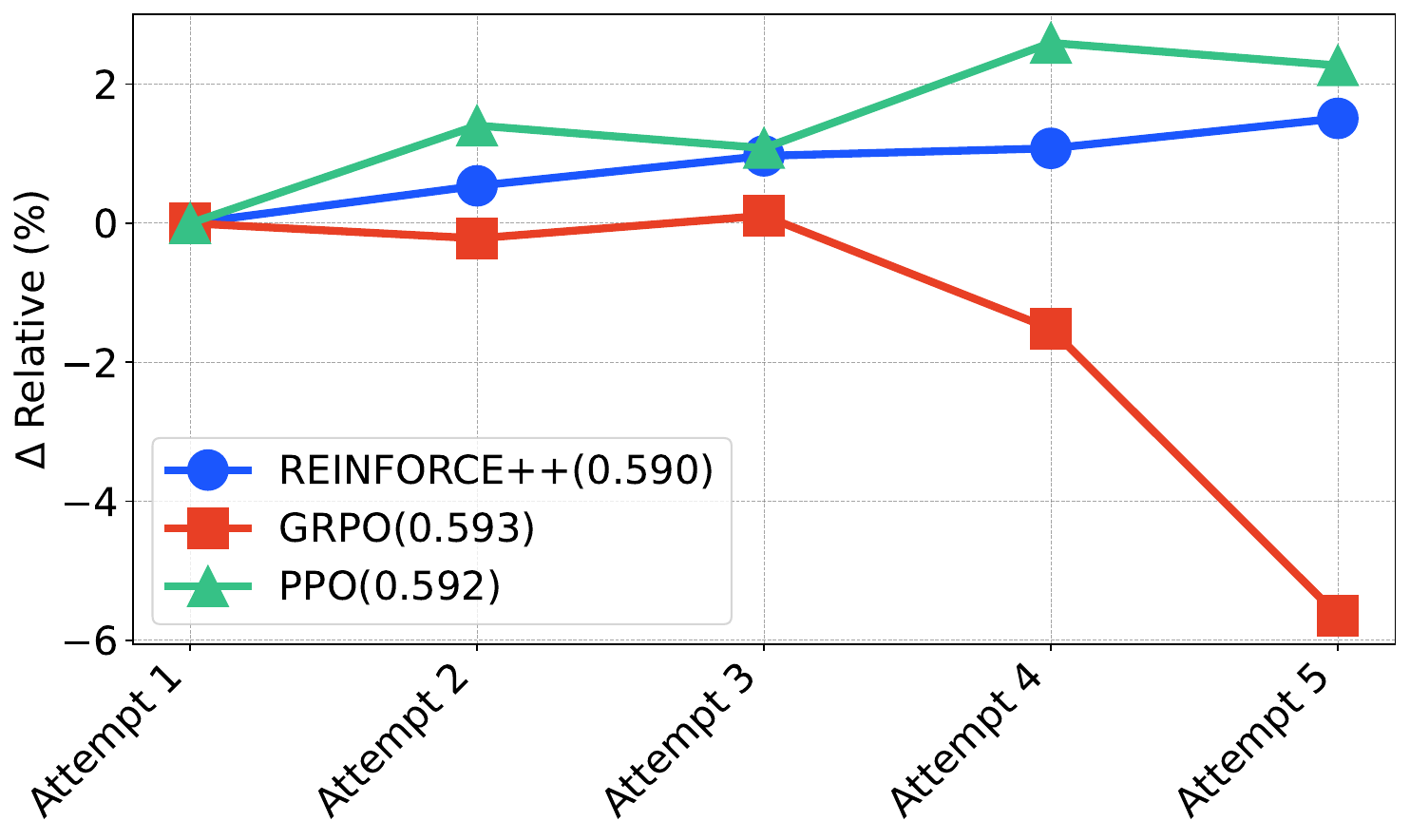}
    \caption{Anthropic Harmless ($\Delta$ Relative)}
    \label{fig:algorithms_ablation/anthropic_harmless/plot_exp2_algorithms}
  \end{subfigure}
  \hfill
  \begin{subfigure}[b]{0.48\textwidth}
    \centering
    \includegraphics[width=0.9\linewidth]{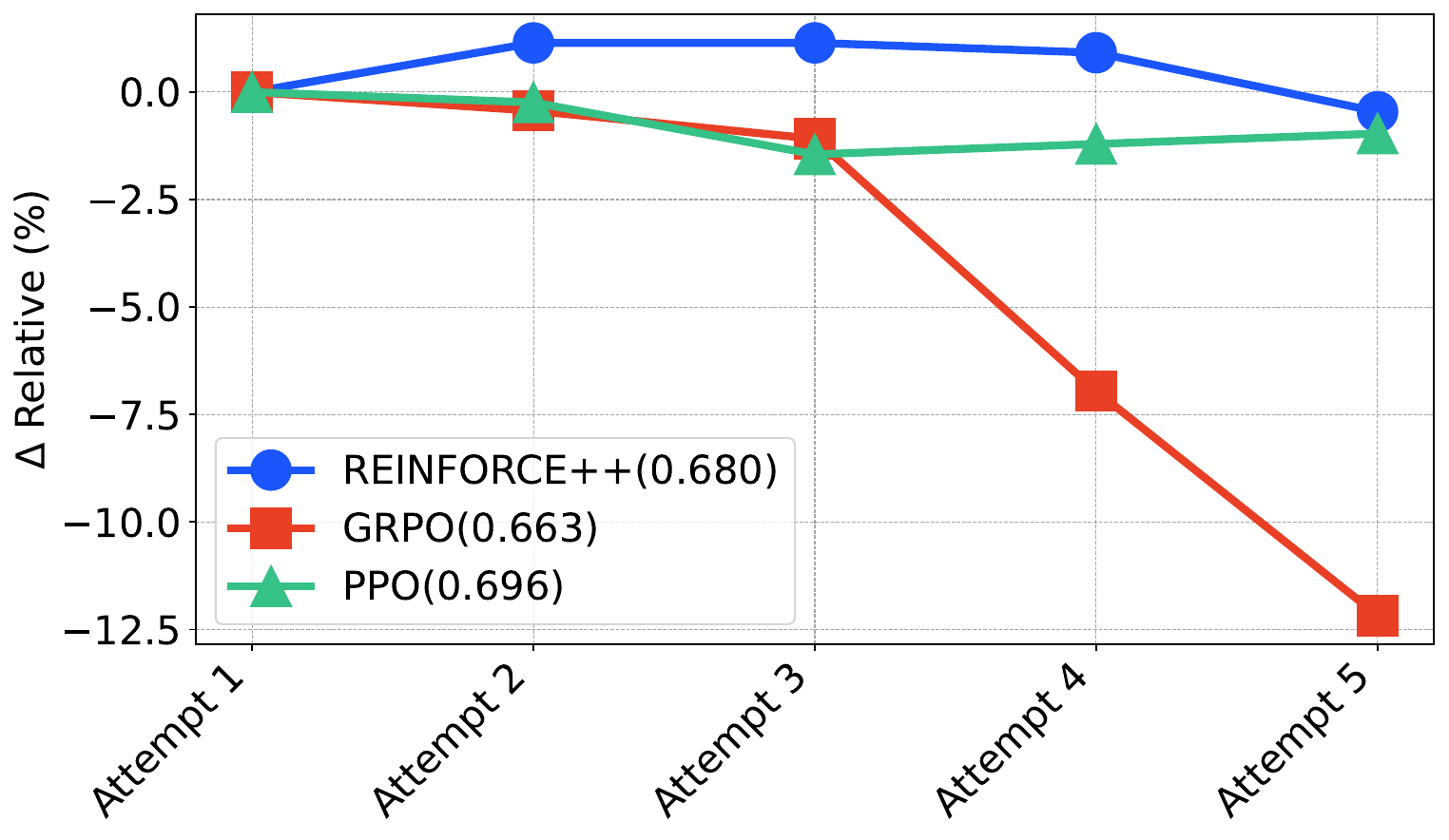}
    \caption{CodePrefBench ($\Delta$ Relative)}
    \label{fig:algorithms_ablation/codeprefbench/plot_exp2_algorithms}
  \end{subfigure}
  
  \vspace{0.5cm}
  
  \begin{subfigure}[b]{0.48\textwidth}
    \centering
    \includegraphics[width=0.9\linewidth]{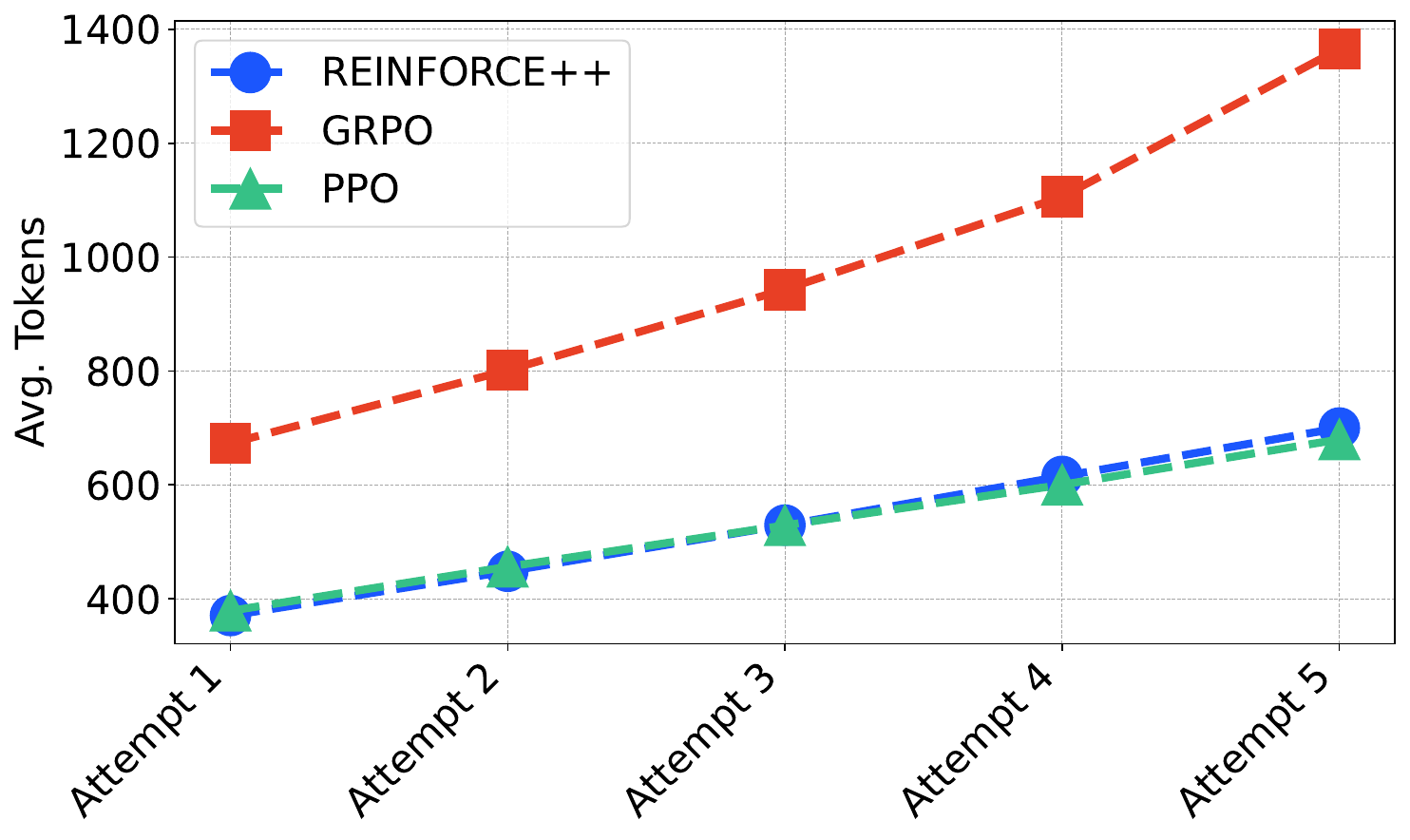}
    \caption{Anthropic Harmless (Avg. Tokens)}
    \label{fig:algorithms_ablation/anthropic_harmless/plot_exp2_algorithms_token_num}
  \end{subfigure}
  \hfill
  \begin{subfigure}[b]{0.48\textwidth}
    \centering
    \includegraphics[width=0.9\linewidth]{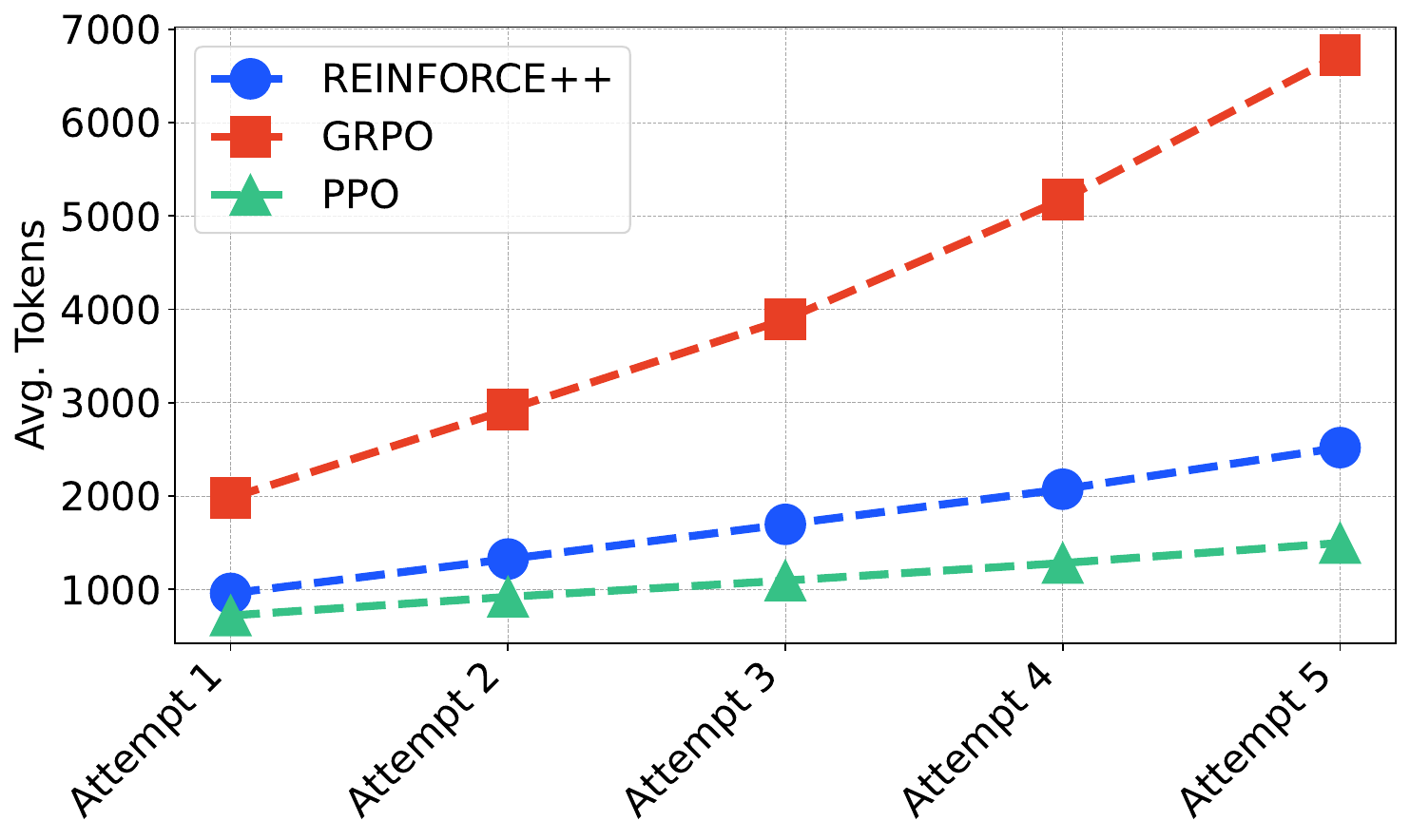}
    \caption{CodePrefBench (Avg. Tokens)}
    \label{fig:algorithms_ablation/codeprefbench/plot_exp2_algorithms_token_num}
  \end{subfigure}
  
  \caption{Scaling trend for STTS on Anthropic Harmless and CodePrefBench with different RL algorithms.}
  \label{fig:algorithms_ablation_1}
\end{figure}

The overall experimental results are summarized in Table~\ref{tab:main}. Our approach demonstrates superior performance compared to both advanced closed-source models and previously established open-source \lmj. Although closed-source models such as \textit{o1-mini} and \textit{Gemini-2.0-flash} exhibit exceptional capabilities across various reasoning benchmarks, their discriminative performance on preference-based datasets is relatively weaker. This highlights a notable distribution shift and indicates substantial potential for improvement in preference discrimination tasks. Furthermore, while existing state-of-the-art open-source models perform competitively on the RewardBench dataset, our model consistently surpasses them across other benchmarks, including mathematics and coding scenarios, culminating in superior overall performance.
Additionally, we investigate the performance of \OURS under two conditions: SFT alone and with subsequent RL. The results underscore the importance of integrating an RL training stage to enhance overall performance significantly.

\subsection{\OURS Demonstrate Superior STTS Scaling Trend.}
\label{subsec:STTS}

Furthermore, we evaluate the effectiveness of STTS across the same tasks, comparing \OURS's performance with that of previous models. By default, we use ``\textbf{Attempt 1}'' to denote the original response without additional reflection, and ``\textbf{Attempt 2-4}'' to denote that we append reflective tokens 2-4 times. As is shown in Figure~\ref{fig:scaling_trend}, we observe positive gains from applying STTS to \OURS across most tasks, typically showing improvements when adding two reflective tokens. However, adding further tokens in tasks like \textit{RewardBench} results in decreased scores. We hypothesize that this decline is due to the significantly increased token length, which surpasses the model's long-context comprehension capabilities, leading to performance degradation. It is noteworthy that previous models also output COT processes before their final answers, but unlike \OURS, they do not consistently exhibit positive gains from STTS. We attribute this difference primarily to the benefits derived from our RL training phase, as further discussed in Section~\ref{subsec:step_increase}.

\subsection{Cold Start on Reasoning-Intensive Data Improve STTS}
\label{subsec:cold_start}

 A natural question is whether incorporating datasets requiring strong reasoning abilities—such as mathematical and coding tasks—during cold-start training could further enhance the STTS capability in downstream tasks. In this section, we explore the effectiveness of using reasoning-intensive data, distinct from the general judgment data, during cold-start training.

Specifically, we employ publicly available datasets OpenR1-Math~\citep{open_r1_math_220k} and CodeForces CoTs~\citep{penedo2025codeforces} and incorporate them alongside \lmj data during the cold-start training phase. Experimental results, as presented in Figure~\ref{fig:algorithms_ablation/cold_start}, demonstrate that introducing mathematical and coding data not only benefits the corresponding tasks but also transfers effectively to safety and open-ended scenarios. Moreover, incorporating these data in cold start stage not only enhances the performance of the initial checkpoint but also boosts the effectiveness of the subsequent STTS.

\subsection{Different RL Algorithms Might Elicit Different STTS behavior}
\label{subsec:different_rl_algorithms}

In this section, we investigate whether different RL algorithms result in varying effectiveness of STTS. Specifically, we take the checkpoints from Section~\ref{subsec:cold_start} that incorporate reasoning-intensive data in cold start stage as a start point and utilize different RL algorithms during RL training. As illustrated in Figure~\ref{fig:algorithms_ablation_1}, different RL algorithms, namely Reinforce++, PPO, and GRPO, achieve comparable scores at the trained checkpoints (scores shown next to the legend). However, we further observe that in most scenarios, the STTS performance trend for GRPO is inferior; specifically, increasing the number of reflective tokens does not yield improved results; instead, it leads to degraded results.
Upon analyzing the number of generated tokens, we find that models optimized using GRPO tend to produce significantly more tokens. Moreover, with increasing STTS, the rate of growth in token generation for GRPO models accelerates faster compared to other algorithms. We attribute this STTS failure to a decreased model inference capability resulting from excessively long contexts.
Additional analyses provided in the Appendix~\ref{appendix:reflective_words_frequency} demonstrate that, even without test-time intervention, models trained with GRPO already exhibit greater usage of reflective tokens. Consequently, the incremental benefit from further application of STTS is markedly diminished.

\begin{figure}[t]
  \centering
  \begin{subfigure}[b]{0.32\textwidth}
    \centering
    \includegraphics[width=\linewidth]{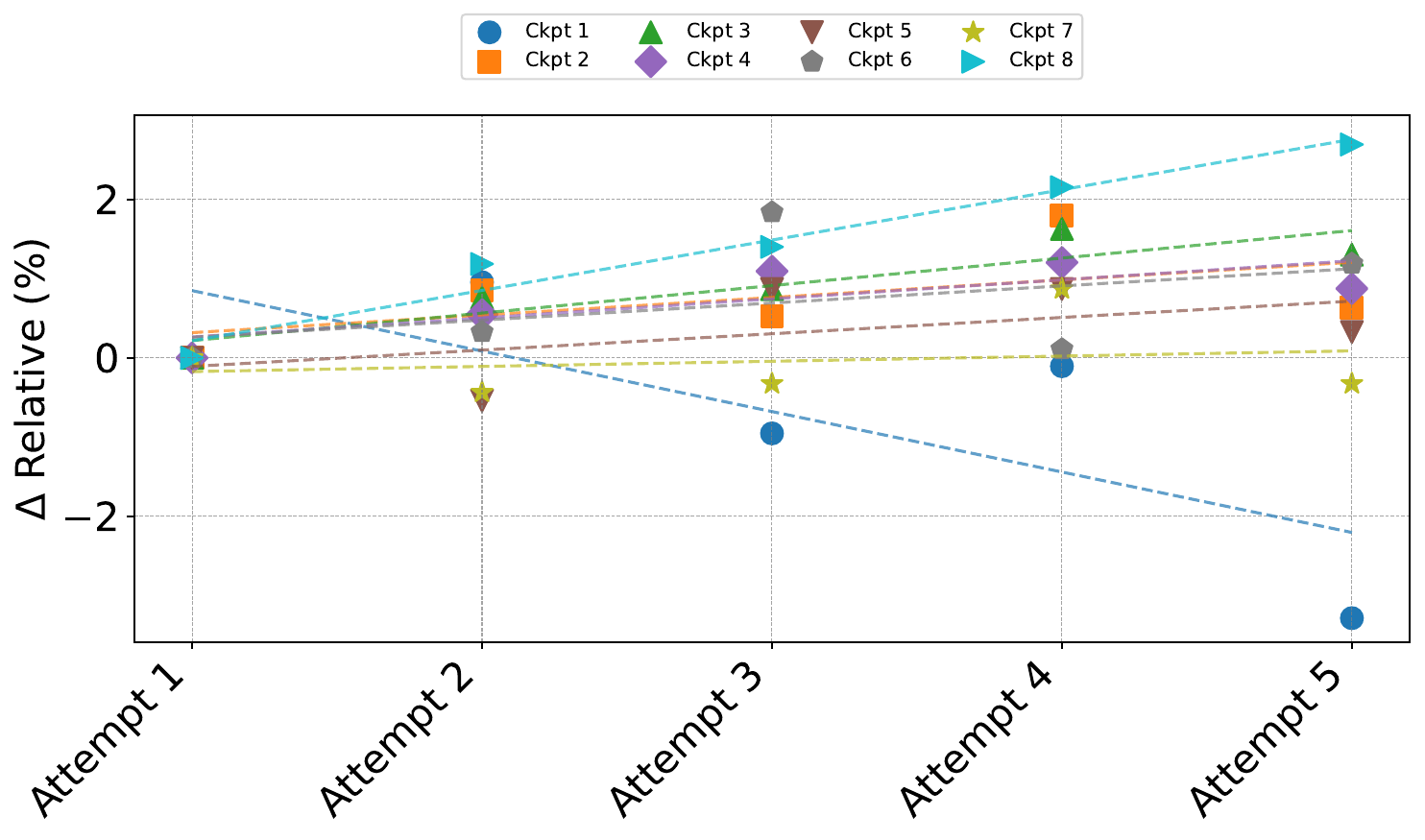}
    \caption{}
    \label{fig:step_progress_ahthropic_1}
  \end{subfigure}
  \hfill
  \begin{subfigure}[b]{0.32\textwidth}
    \centering
    \includegraphics[width=\linewidth]{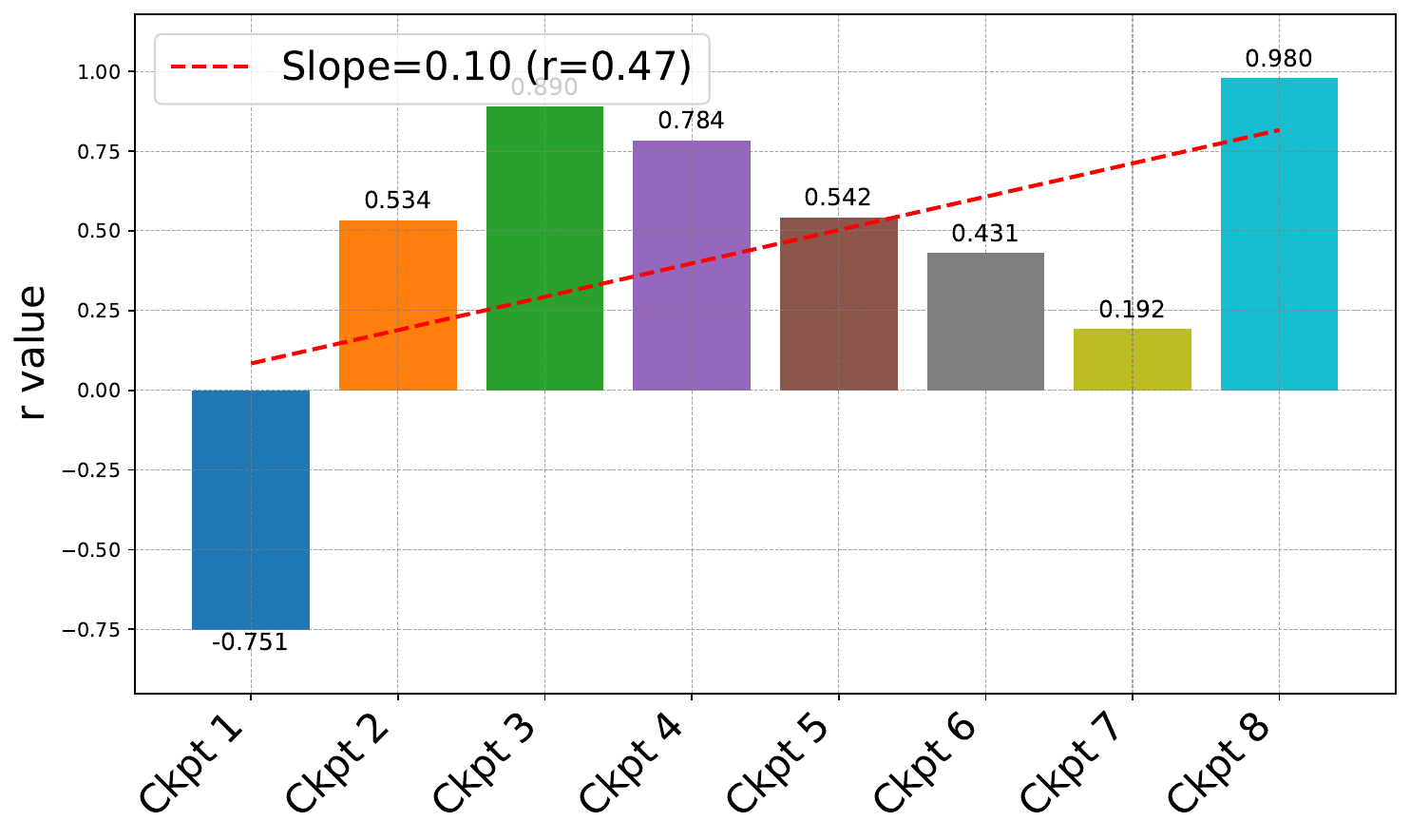}
    \caption{}
    \label{fig:step_progress_ahthropic_2}
  \end{subfigure}
  \hfill
  \begin{subfigure}[b]{0.32\textwidth}
    \centering
    \includegraphics[width=\linewidth]{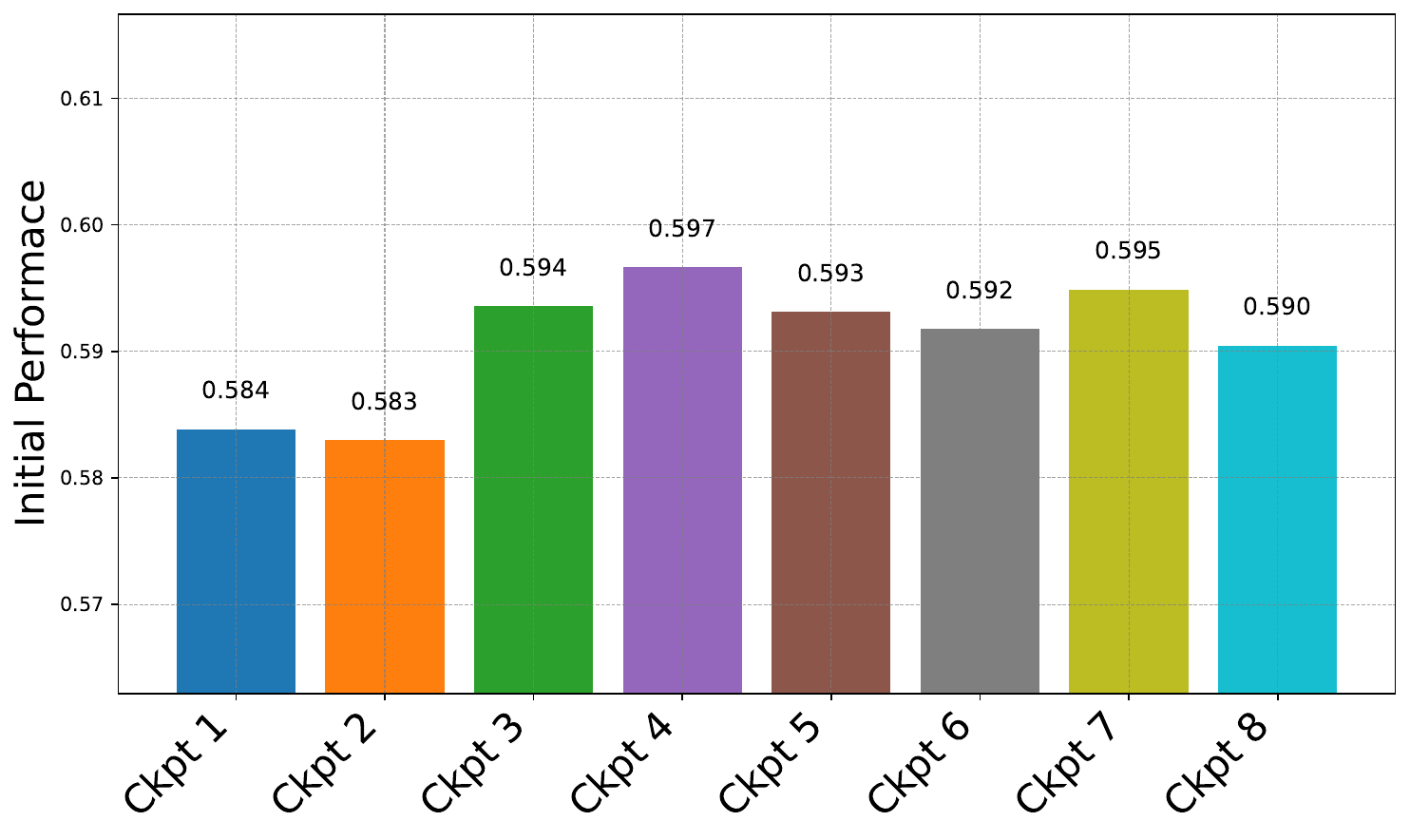}
    \caption{}
    \label{fig:step_progress_ahthropic_3}
  \end{subfigure}
  
  \caption{Scaling behaviour of different checkpoints on Anthropic Harmless.}
  \label{fig:step_progress_ahthropic}
\end{figure}

\subsection{LLM Learns STTS as RL Steps Increase}
\label{subsec:step_increase}

In this section, we delve deeper into the origins of the STTS capability. We hypothesize that models are more likely to acquire reflection skills during the exploration phase of RL. Prior studies primarily emphasize that reflective abilities are intrinsically acquired during pre-training~\citep{shah2025rethinking}. Distinct from this intrinsic perspective, our hypothesis suggests that RL could stimulate STTS through enforced reflection.
To empirically validate this, we evaluate multiple checkpoints from different stages of the RL training phase. Figure~\ref{fig:step_progress_ahthropic_3} first illustrates that the initial performance of these checkpoints improves over the course of training. Subsequently, using the same experimental settings as in previous STTS experiments, we measure the effectiveness of STTS across checkpoints, as depicted in Figure~\ref{fig:step_progress_ahthropic_1}. To quantify the trend of improvement with respect to STTS attempts, we perform linear regression on the results and compute the corresponding $r$-values, where higher $r$-values indicate a more pronounced upward trend. Figure~\ref{fig:step_progress_ahthropic_2} clearly demonstrates an increasing pattern in $r$-values throughout the RL training process.
These findings collectively suggest that the capability of enforced reflection via STTS likely emerges progressively during reinforcement learning. We anticipate that this insight will inspire future investigations and developments within the community.

%% file: tables/main.tex
\begin{table}[t]
\centering
\caption{Performance comparison between \OURS and baseline methods.}
\renewcommand{\arraystretch}{1.1}
\label{tab:main}
\resizebox{\textwidth}{!}{
% \begin{tabular}{p{4.2cm}p{1.5cm}p{1.5cm}p{1.8cm}p{1.5cm}p{1.2cm}}
\begin{tabular}{lccccc}
% \begin{tabular}{XXXXXX}
\toprule
% \multirow{2}{*}{\textbf{Model}} 
% \multirow{2}{*}{\textbf{Overall}}

\textbf{Model} & \textbf{RewardBench} & \textbf{RewardMath} & \textbf{Anthropic Harmless} & \textbf{CodePrefBench} & \textbf{Overall} \\
\hline
\hline
\multicolumn{6}{c}{\textit{Results of LLM-as-a-Judge (Proprietary or Larger Models)}} \\
\hline
GPT-4o &85.40 &76.81 &51.37 &75.09 &72.17  \\
o1-mini  &88.56 &95.70 &45.76 &76.13 &76.54  \\
Gemini-2.0-flash  &83.68 &85.30 &48.98 &69.21 &71.79  \\
Deepseek-R1 &82.83 &98.21 &53.31 &79.49 &78.46  \\

\hline
\multicolumn{6}{c}{\textit{Results of Scalar Reward Models (Open Source with Comparable Size)}} \\
\hline

InternLM2-7B-Reward  &88.18 &78.05 &\textbf{71.76} &62.48 &75.12  \\
Skywork-Reward-Llama3.1-8B  &\textbf{93.37} &72.33 &59.44 &59.89 &71.26  \\
FsfairX-LLaMA3-8B &84.45 &66.23 &53.75 &58.20 &65.66  \\

\hline
\multicolumn{6}{c}{\textit{Results of LLM-as-a-Judge (Open Source with Comparable Size)}} \\
\hline

Llama3.1-8B-Instruct &70.47 &61.12 &46.43 &67.10 &61.28  \\
Qwen2.5-7B-Instruct &78.50 &69.70 &49.56 &67.59 &66.34  \\
Skywork-Critic-Llama3.1-8B &88.86 &66.51 &58.61 &60.57 &68.64  \\
RISE-Judge-Qwen2.5-7B &87.42 &81.69 &56.35 &59.22 &71.17 \\

\hline
\multicolumn{6}{c}{\textit{Results of Our Models}} \\
\hline
\textbf{J1-7B (SFT Only)} & 85.01 & 82.40 & 53.88 & 49.20 & 67.62  \\
\textbf{J1-7B (SFT + RL)} & 86.91 & \textbf{90.15} & 59.05 & \textbf{67.80} & \textbf{75.98} \\
\bottomrule

% \hline
\end{tabular}}

\end{table}

%% file: sections/appendix.tex
\section{Comparision between Parallel Decoding and STTS}
\label{appendix:comparison_between_pd_stts}

\begin{figure}[ht]
  \centering

   \begin{subfigure}[b]{0.48\textwidth}
    \centering
    \includegraphics[width=\linewidth]{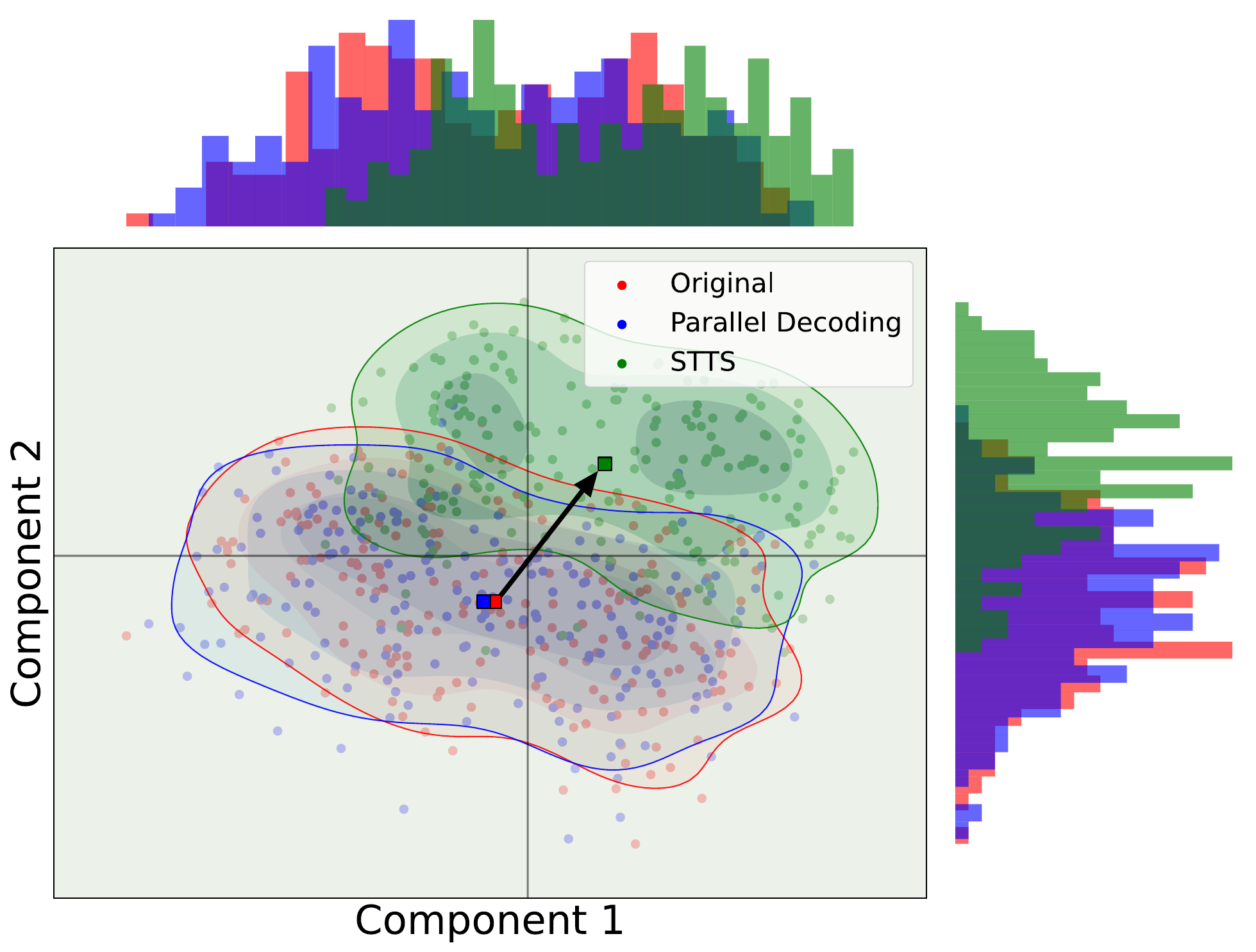}
    \caption{Anthropic Harmless}
    \label{fig:exp0_anthropic_harmless}
  \end{subfigure}
  \hfill
  \begin{subfigure}[b]{0.48\textwidth}
    \centering
    \includegraphics[width=\linewidth]{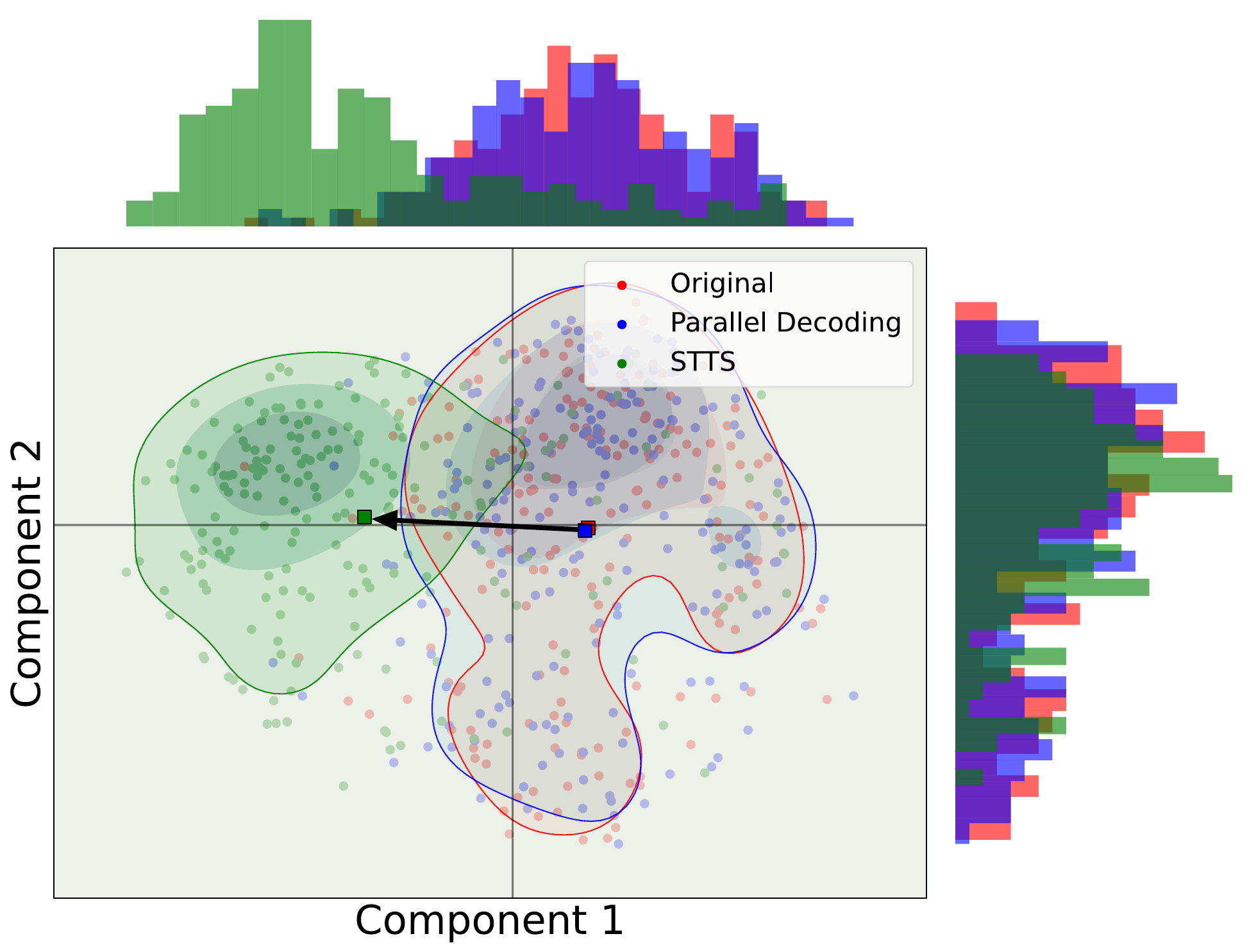}
    \caption{CodePrefBench}
    \label{fig:exp0_codeprefbench}
  \end{subfigure}

  \vspace{0.5cm} 
  
  \begin{subfigure}[b]{0.48\textwidth}
    \centering
    \includegraphics[width=\linewidth]{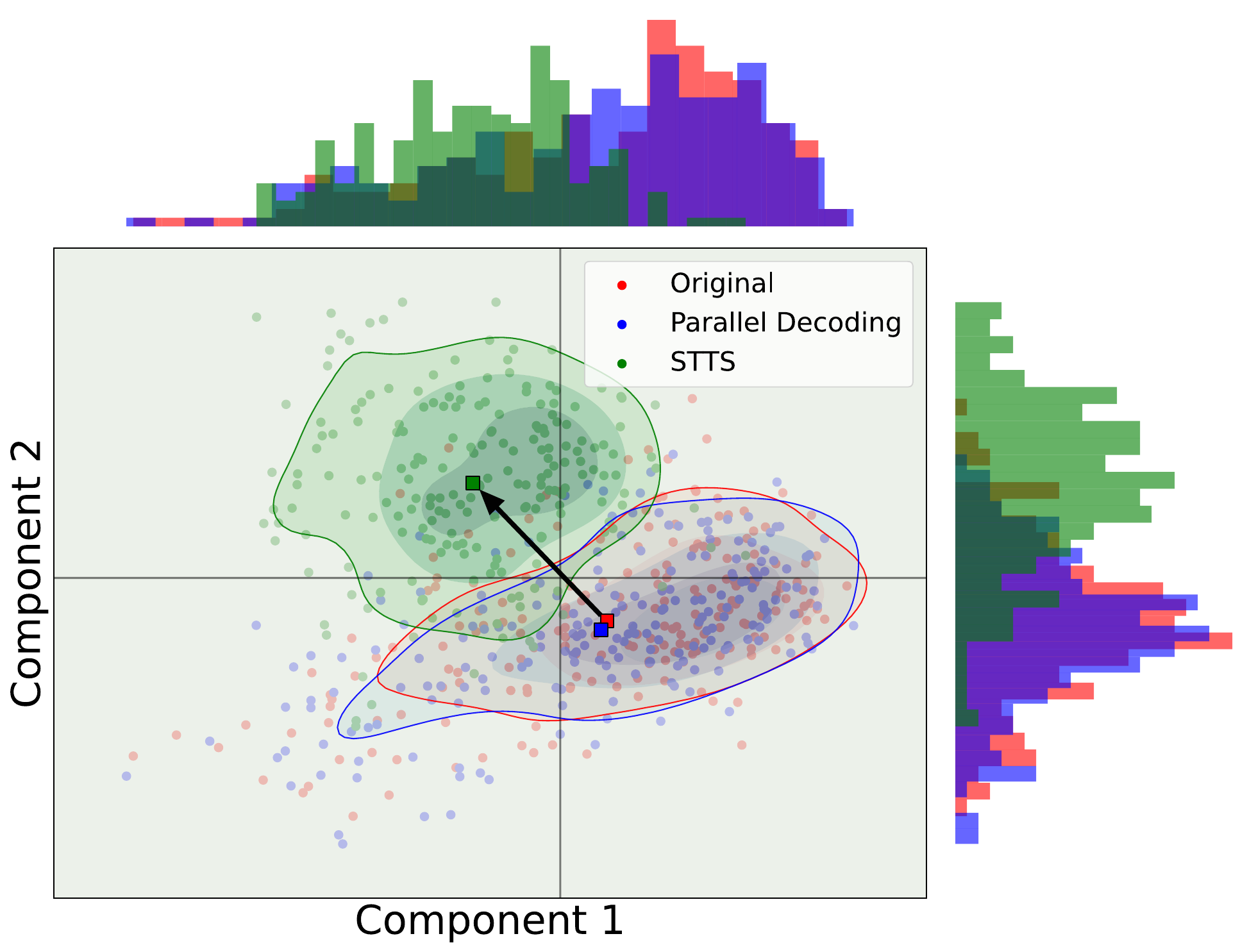}
    \caption{RewardBench}
    \label{fig:exp0_rewardbench}
  \end{subfigure}
  \hfill
  \begin{subfigure}[b]{0.48\textwidth}
    \centering
    \includegraphics[width=\linewidth]{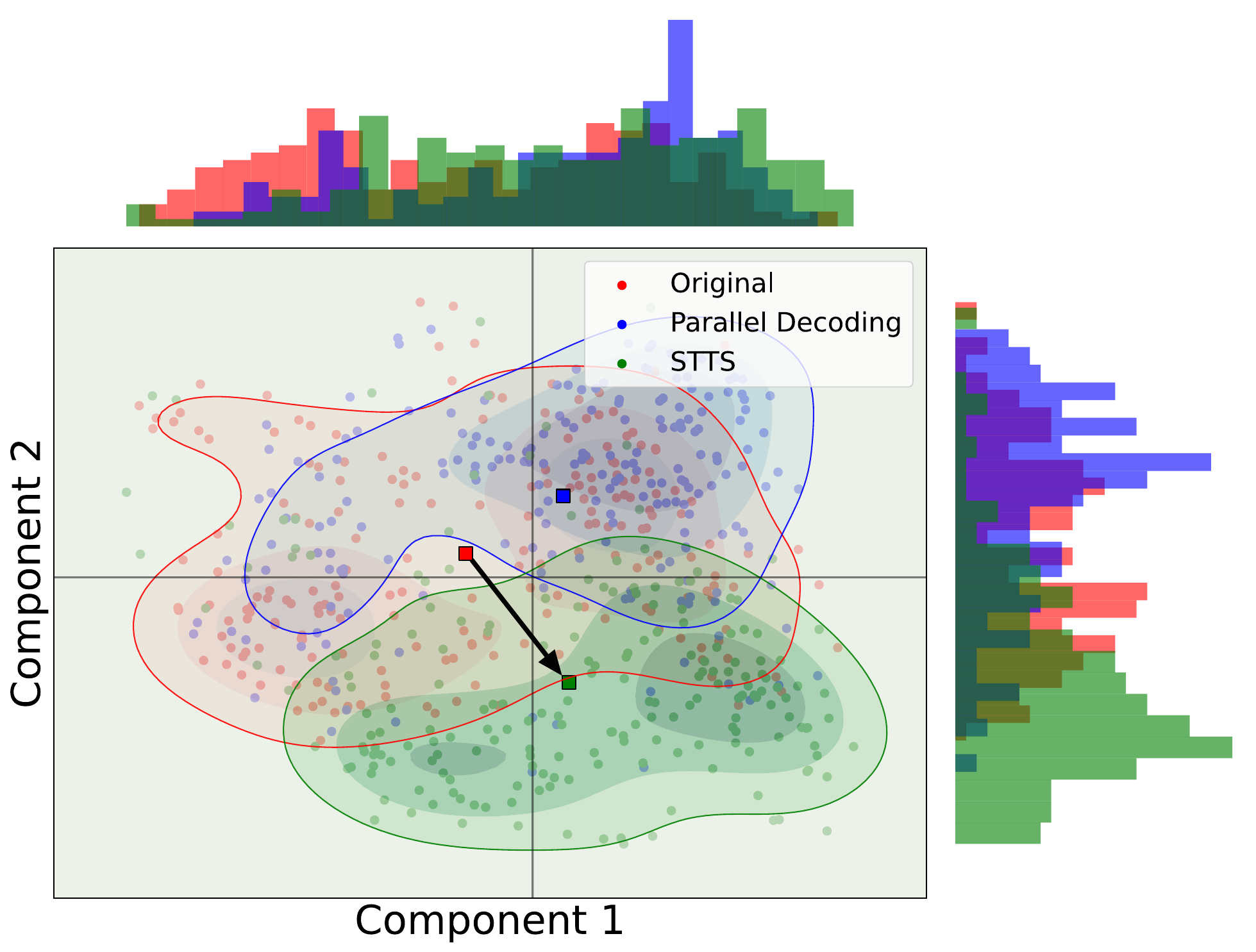}
    \caption{RewardMATH}
    \label{fig:exp0_rewardmath}
  \end{subfigure}
  
  \caption{Distribution patterns between STTS and Parallel Decoding.}
  \label{fig:exp0}
\end{figure}

In this section, we present empirical evidence that STTS generates outputs with greater diversity compared to parallel decoding. We randomly select 100 instances from four benchmarks and visualize the results in Figure~\ref{fig:exp0}. 

The ``\textbf{Original}'' and ``\textbf{Parallel Decoding}'' outputs represent two distinct responses generated with a temperature parameter of 0.8, while the ``\textbf{STTS}'' outputs were obtained after four append-wait iterations based on ``\textbf{Original}''. We employ the all-MiniLM-L6-v2 model~\citep{wang2020minilm} for embedding generation and use PCA~\citep{mackiewicz1993principal} for dimensionality reduction to plot the distribution trajectories.

Figure~\ref{fig:exp0} clearly demonstrates that while the Original and Parallel Decoding outputs exhibit similar distribution patterns, STTS outputs show significantly greater diversity in their spatial distribution. Combined with Figure~\ref{fig:teaser}, our quantitative and qualitative analyses confirm that STTS exhibits greater diversity compared to parallel decoding, demonstrating stronger potential to yield superior solutions through iterative refinement. Moreover, parallel decoding incurs significant token consumption, whereas STTS is notably more sample-efficient. Motivated by these findings, we primarily focus on exploring STTS for \lmj throughout this paper.

\section{Training Dataset Statistics}
\label{appendix:training_dataset_statistics}

Below is a brief introduction to each training dataset.
\begin{itemize}
\item[$\bullet$] HelpSteer2~\citep{wang2024helpsteer2} is an open-source dataset designed to train reward models for improved helpfulness. It helps align models to be more accurate and coherent while allowing adjustments in response complexity and verbosity.
% The dataset includes 10,681 annotated prompt-response pairs.
\item[$\bullet$] Magpie~\citep{xu2024magpie} is a large scale synthesized instruction dataset. During synthesis, rather than relying on prompt engineering or seed questions, instruction data is generated from open-weight LLMs by directly using a pre-query template.
\item[$\bullet$] OffsetBias~\citep{park2024offsetbias} is a dataset of paired preferences designed to mitigate typical biases found within judge models. This dataset is constructed by GPT-4 and Claude-3, and incorporates prompting techniques like the Off-topic response approach and the Erroneous response approach.
% OffsetBias comprises 8,504 samples for training.
\item[$\bullet$] WildGuard~\citep{han2024wildguard} is a meticulously balanced multitask moderation dataset in 13 risk categories. Data are selected from four sources (i.e., synthetic
vanilla/adversarial, in-the-wild, and annotator-written data) to ensure completeness.

\end{itemize}

\subsection{Statistics During Rejection Sampling}
\label{appendix:statistics_during_rejection}

Table~\ref{appendix:curating_process} illustrates our data construction procedure using DeepSeek-R1 via rejection sampling. Each cell's left value indicates the number of correctly answered samples, while the right value represents the total samples at the current phase. A notable observation is that the model initially achieves an overall accuracy of approximately 70\%. However, when employing STTS in subsequent attempts, only around 4\% of previously incorrect responses are corrected. Moreover, this conversion rate further diminishes with additional attempts, suggesting that while STTS provides benefits, the model predominantly maintains its initial decisions.

\begin{table}[ht]
  \centering
  \caption{Statistics of curating process.}
  \label{appendix:curating_process}
  \resizebox{\textwidth}{!}{
  \begin{tabular}{l||c|c|c|c|c} % 使用 tabularx 并指定列类型
    \toprule
    \textbf{Dataset} & \textbf{Original Number} & \textbf{Attempt 0} & \textbf{Attempt 1} & \textbf{Attempt 2} & \textbf{Attempt 3} \\ \midrule
    HelpSteer2 & 6766 & 3366/6766 & 99/3400 & 59/3301 & 36/3242 \\ \midrule
    OffsetBias & 8504 & 7327/8504 & 85/1177 & 44/1092 & 25/1048 \\ \midrule
    WildGuard & 6709 & 5828/6709 & 110/881 & 29/771 & 19/742 \\ \midrule
    Magpie & 54582 & 36846/54582 & 599/17736 & 260/17137 & 194/16877 \\
    \bottomrule
  \end{tabular}}
\end{table}

\subsection{Other Statistics of Curated Dataset}
\label{appendix:other_statistics}

\paragraph{Length Distribution}

\begin{figure}[ht]
  \centering
  \begin{subfigure}[b]{0.48\textwidth}
    \centering
    \includegraphics[width=\linewidth]{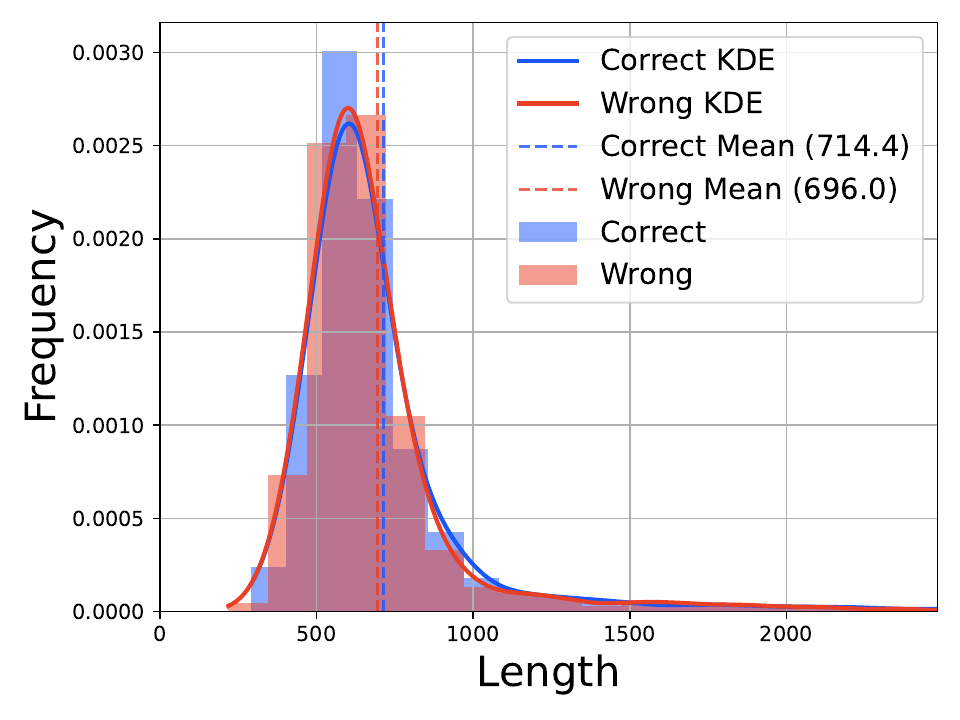}
    \caption{HelpSteer2}
    \label{appendix:length_distribution_1}
  \end{subfigure}
  \hfill
  \begin{subfigure}[b]{0.48\textwidth}
    \centering
    \includegraphics[width=\linewidth]{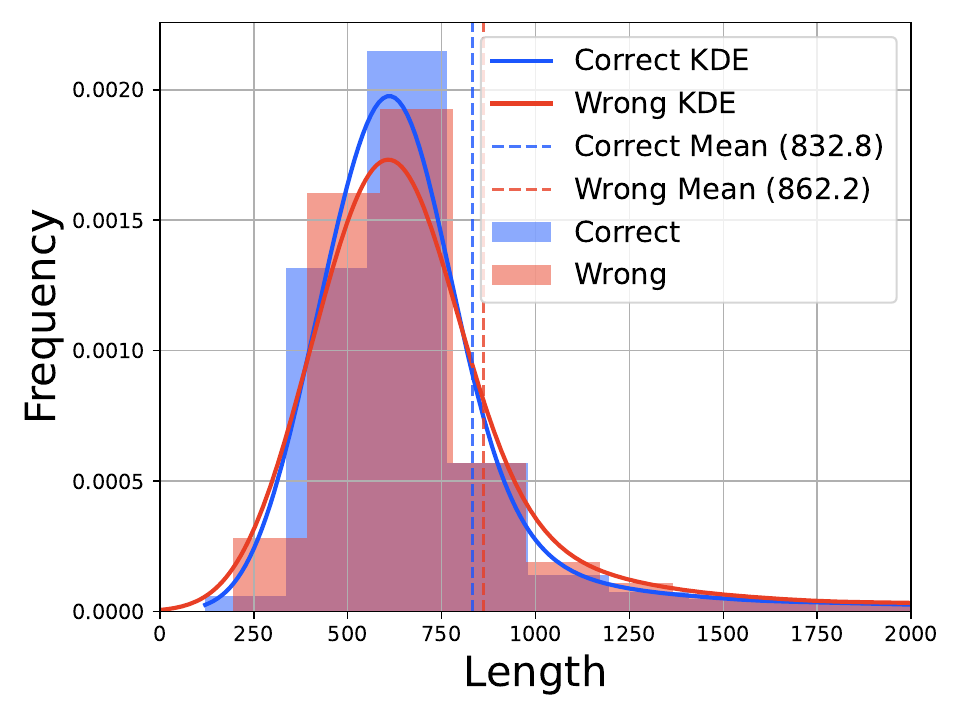}
    \caption{Magpie}
    \label{appendix:length_distribution_2}
  \end{subfigure}
  
  \vspace{0.5cm} 
  
  \begin{subfigure}[b]{0.48\textwidth}
    \centering
    \includegraphics[width=\linewidth]{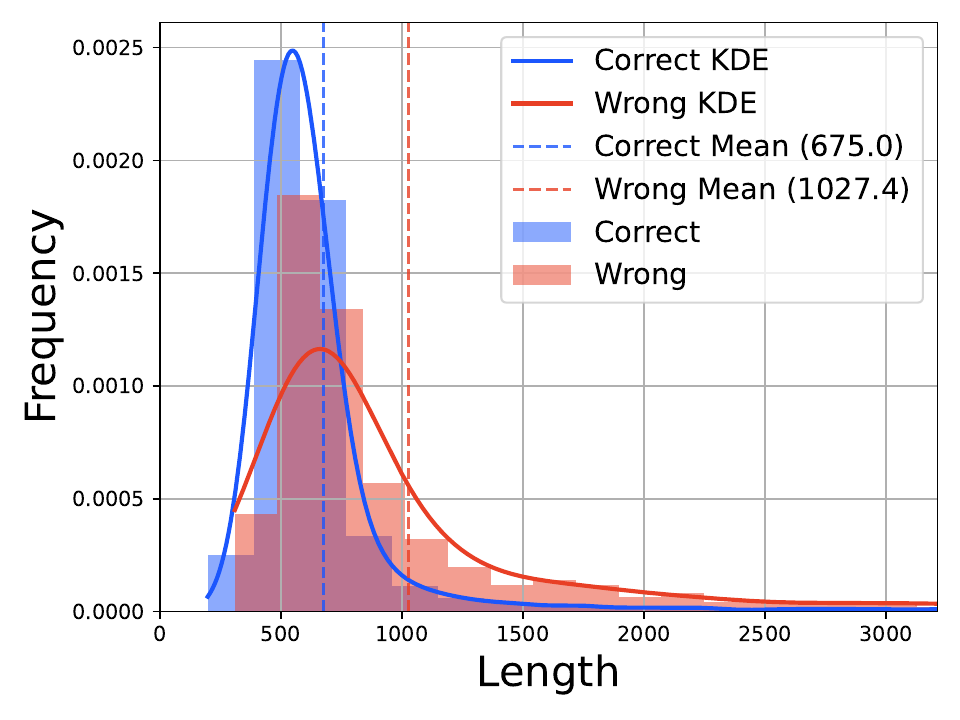}
    \caption{OffsetBias}
    \label{appendix:length_distribution_3}
  \end{subfigure}
  \hfill
  \begin{subfigure}[b]{0.48\textwidth}
    \centering
    \includegraphics[width=\linewidth]{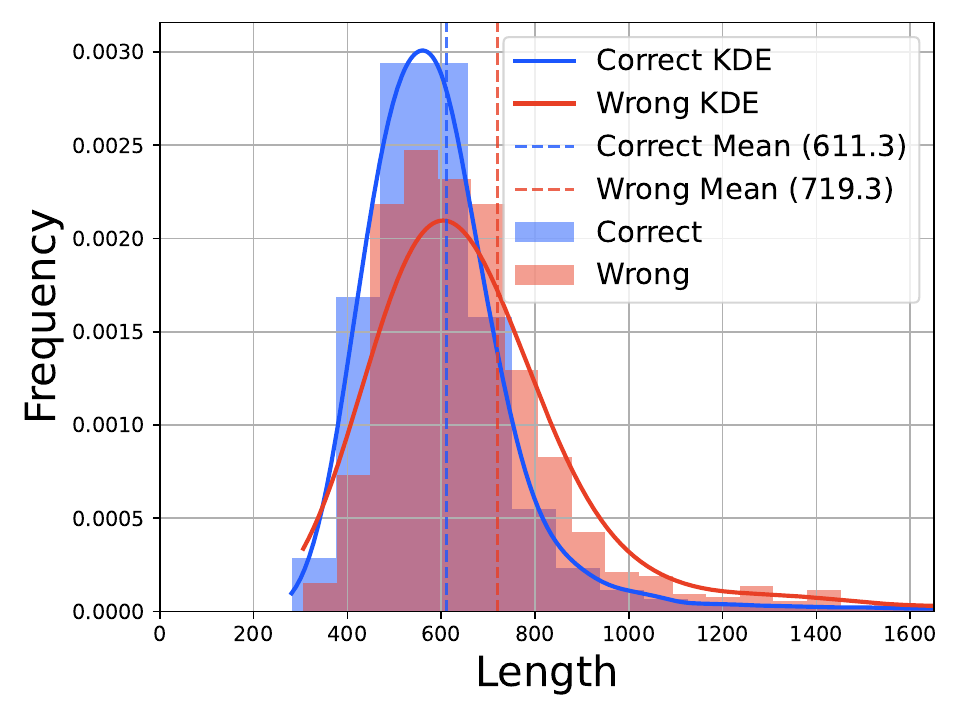}
    \caption{WildGuard}
    \label{appendix:length_distribution_4}
  \end{subfigure}
  
  \caption{Length distribution of the curated dataset.}
  \label{appendix:length_distribution}
\end{figure}

We present statistical analyses of sequence lengths for the constructed training data in Figure~\ref{appendix:length_distribution}, detailing the distributions across different datasets. We further partition the data into two categories based on whether the model correctly or incorrectly answered each instance. Notably, sequences in the incorrect category consistently exhibit greater lengths compared to those in the correct category. This observation suggests that the model may engage in redundant or ineffective reasoning, employing additional tokens without successfully arriving at the correct final answer.

\paragraph{Reflective Words Frequency}

\begin{figure}[ht]
  \centering
  \begin{subfigure}[b]{0.48\textwidth}
    \centering
    \includegraphics[width=\linewidth]{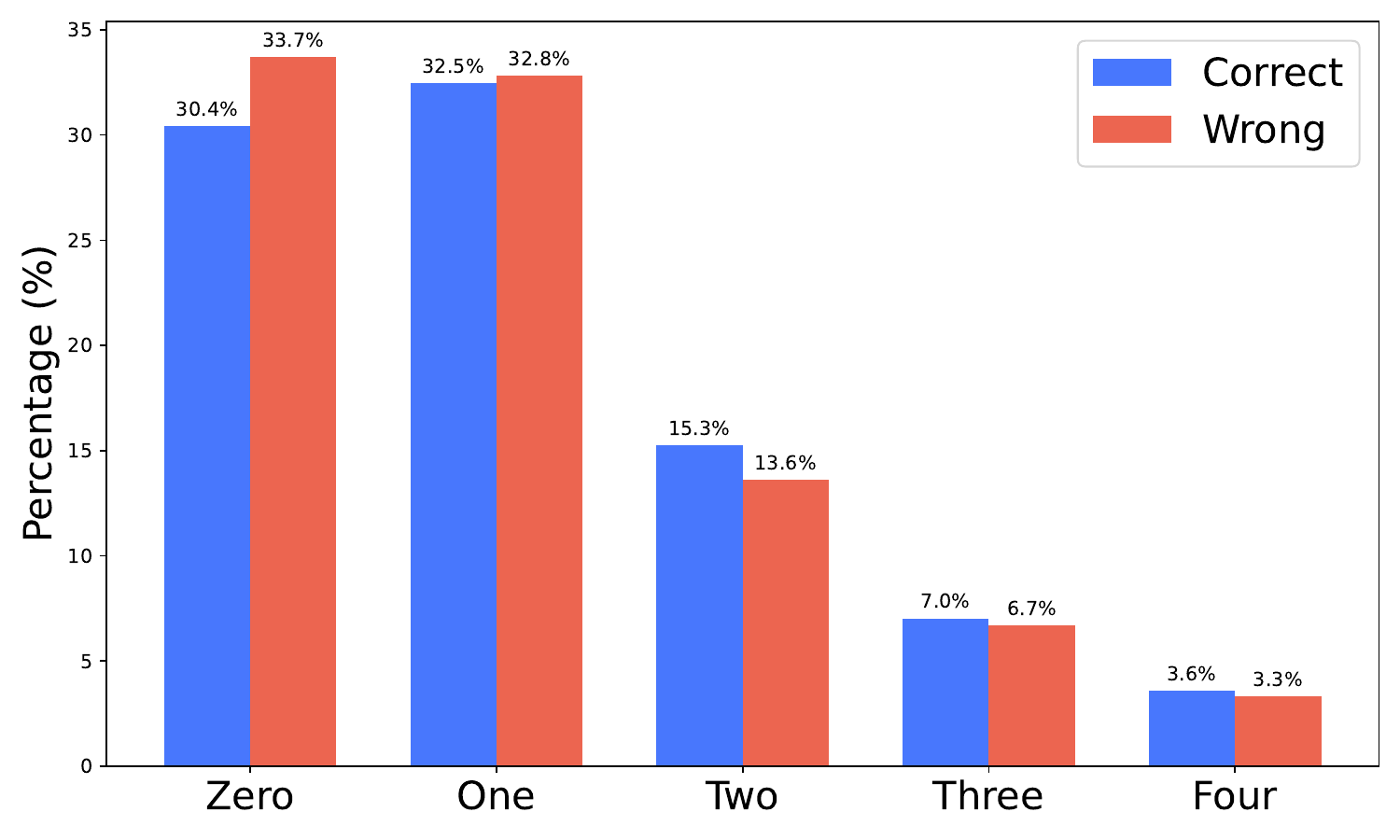}
    \caption{HelpSteer2}
    \label{appendix:wait_freq_1}
  \end{subfigure}
  \hfill
  \begin{subfigure}[b]{0.48\textwidth}
    \centering
    \includegraphics[width=\linewidth]{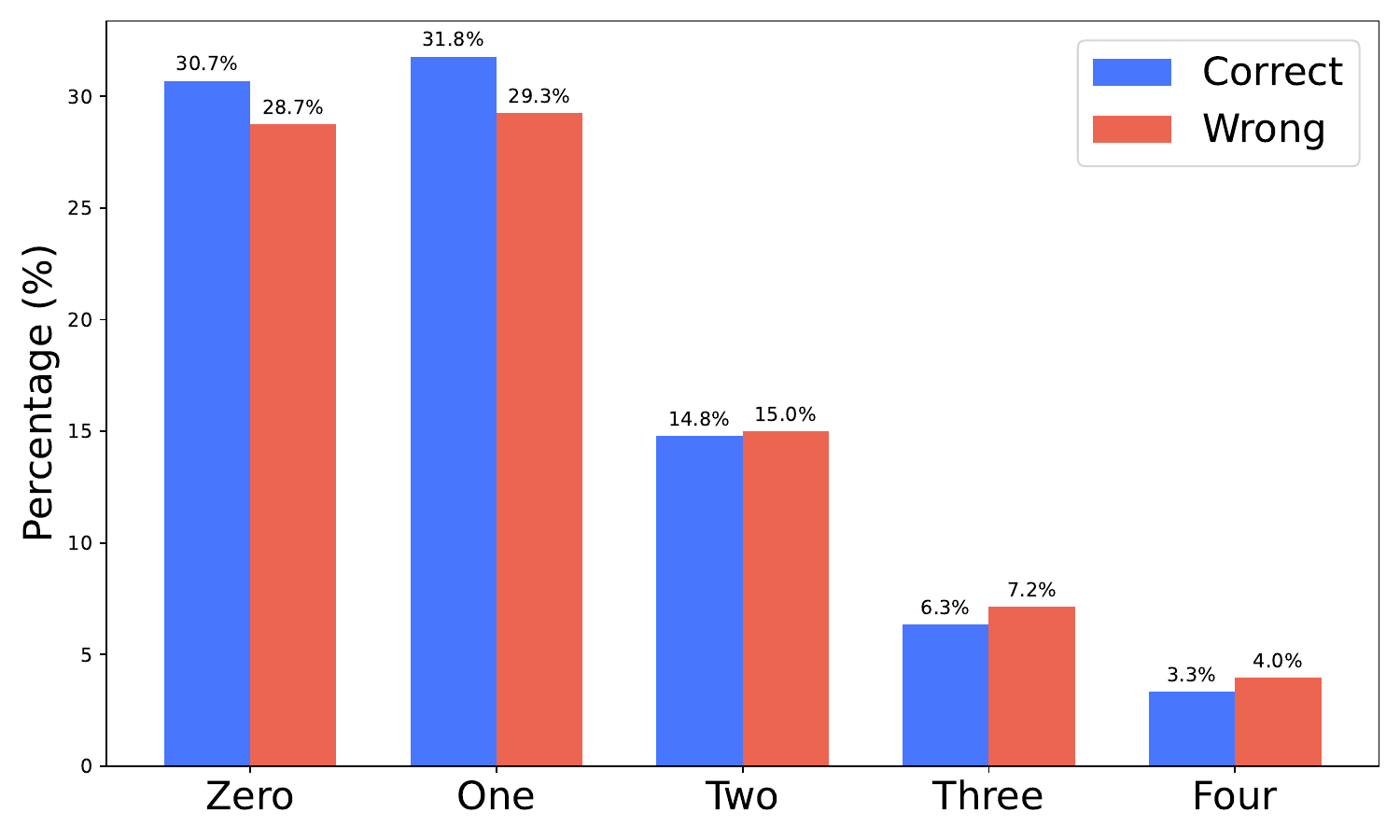}
    \caption{Magpie}
    \label{appendix:wait_freq_2}
  \end{subfigure}
  
  \vspace{0.5cm} 
  
  \begin{subfigure}[b]{0.48\textwidth}
    \centering
    \includegraphics[width=\linewidth]{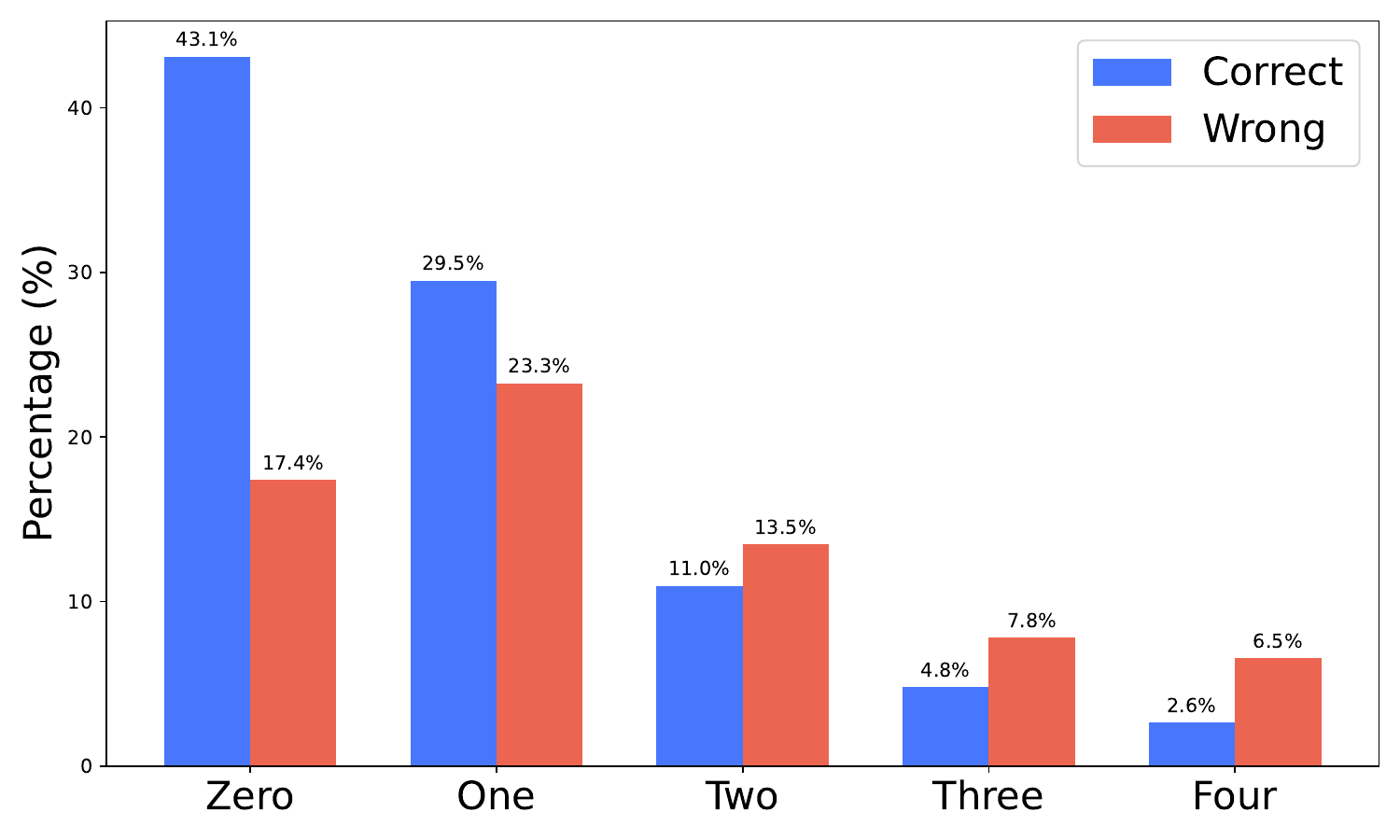}
    \caption{OffsetBias}
    \label{appendix:wait_freq_3}
  \end{subfigure}
  \hfill
  \begin{subfigure}[b]{0.48\textwidth}
    \centering
    \includegraphics[width=\linewidth]{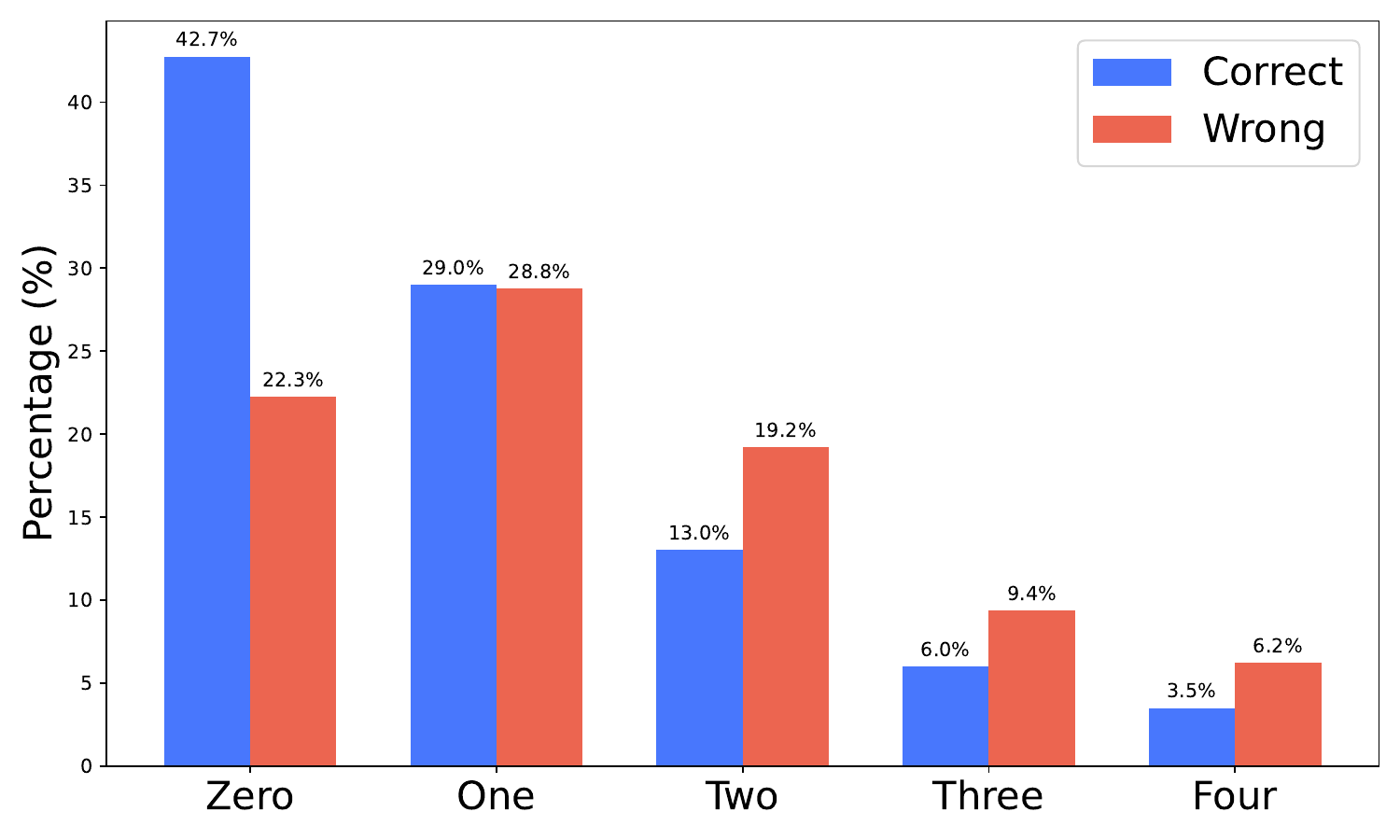}
    \caption{WildGuard}
    \label{appendix:wait_freq_4}
  \end{subfigure}
  
  \caption{Reflective words frequency of the curated dataset.}
  \label{appendix:wait_freq_trainset}
\end{figure}

We further examine the distribution of instances according to the number of reflective words employed in reasoning, again partitioning the data into correct and incorrect categories. The results are shown in Figure~\ref{appendix:wait_freq_trainset}. Our findings reveal two primary observations: (1) irrespective of correctness, the majority of instances involve no reflective words; (2) incorrect instances demonstrate a higher likelihood of employing a greater number of reflective words compared to correct instances. These results suggest that moderate reflection is beneficial; a limited use of reflective words aids the model's reasoning, whereas the absence or excessive use of reflection may negatively impact overall performance.

\begin{table}[ht]
  \centering
  \caption{Reflective words we take into account.}
  \label{appendix:reflective_words}
  % \resizebox{\textwidth}{!}{
  % }
  \begin{tabular}{c} 
    \hline
    Reflective Words \\
    \hline
    Wait, Alternatively, But, However, Hold on, On the other hand, On the contrary, In contrast \\
    \hline
  \end{tabular}
\end{table}

\begin{figure}[t]
  \centering
  \includegraphics[width=\textwidth]{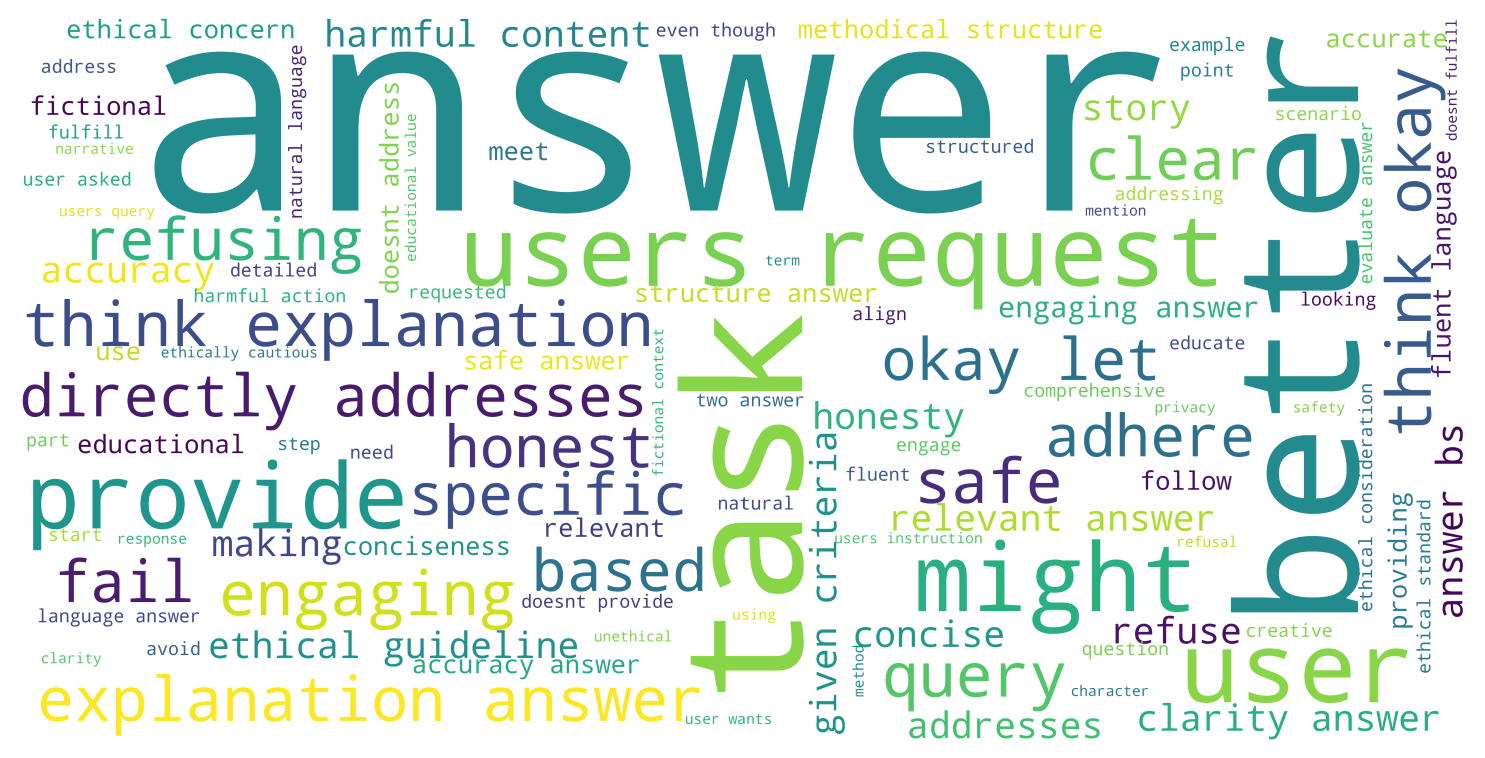}
  \caption{Word cloud of the curated dataset.}
  \label{appendix:wordcloud}
\end{figure}

\paragraph{Word Cloud} The word cloud in Figure~\ref{appendix:wordcloud} visually summarizes the prominent vocabulary present in our training dataset. It clearly illustrates the relative prevalence of various terms, particularly those corresponding to different evaluation criteria (`ethical concern', `concise', `engaging answer', etc.). This visualization supports the conclusion that our dataset effectively captures the intended phenomenon—that \lmj, during reasoning processes, indeed utilize explicit criteria when formulating their conclusions.

\section{Evaluation Benchmark Details}
\label{appendix:benchmark}
All evaluation benchmarks consist of binary classification tasks. Following previous works, we use accuracy as the evaluation metric. Below is a brief overview of each evaluation benchmark.
\begin{itemize}
\item[$\bullet$] RewardBench~
\citep{lambert2024rewardbench} is a collection of sets consisting of a prompt, a chosen response, and a rejected response across categories like chat, reasoning, and safety. The dataset is designed to evaluate the effectiveness of reward models when dealing with complex, organized, and out-of-distribution queries.

\item[$\bullet$] RewardMath~\citep{kim2024evaluating} is a benchmark crafted to evaluate the strength of reward models in tackling mathematical reasoning problems. It has a 1:9 ratio of chosen to rejected solutions. Note that in our paper, we calculate an instance-wise score rather than a problem-wise score, as solely considering whether the reward of the chosen solution is the highest can be overly strict.

\item[$\bullet$] Anthropic Harmless~\citep{bai2022training} is a benchmark used for evaluating models' harmlessness. During the data collection process, human annotators are encouraged to provoke LLMs into generating harmful responses and to identify which response is more harmful.

\item[$\bullet$] CodePrefBench~\citep{liu2024learning} is a comprehensive benchmark for assessing developer preferences. The examples in the benchmark are labeled based on three verifiable criteria: correctness, efficiency, and security. Additionally, it includes overall preferences of developers as collected from 18 annotators.

\end{itemize}

\section{Implementation Details}
\label{subsec:implementation}

Our initial policy model is based on \textit{Qwen2.5-7B-Base}. In the SFT stage, we utilize a learning rate of $2 \times 10^{-5}$, train for 3 epochs, set the maximum sequence length to 8192 tokens, and employ a batch size of 256. For the RL stage, we adopt a learning rate of $5 \times 10^{-7}$, conduct training for 15 epochs, use a batch size of 256, and set the number of rollout samples to 8. The default RL algorithm employed in our primary experiments is Reinforce++. All experiments are conducted on a NVIDIA H800 cluster. Additional experimental details, including the hyperparameters for PPO and GRPO algorithms, can be found in the Appendix~\ref{appendix:training_details}.

\section{Other Training Details}
\label{appendix:training_details}

In our experiments, we employ three RL algorithms: PPO, Reinforce++ and GRPO. Both PPO and Reinforce++ share the same general objective formulation, differing primarily in how they compute the advantage estimate.

Specifically, \(\hat{A}_t^{\text{PPO}}\) is computed using the Generalized Advantage Estimation (GAE) method~\citep{schulman2015high}:
\begin{align}
\delta_t &= r_t + \gamma V_{t+1} - V_t,\\
\label{equation:gae}
\hat{A}_t^{\text{PPO}} &= \sum_{l=0}^{T}(\gamma\lambda)^l\delta_{t+l}.
\end{align}

For Reinforce++, we adopt the same objective function structure as PPO. However, the advantage estimate \(\hat{A}_t^{\text{RF++}}\) is defined differently as follows:
\begin{align}
\hat{A}_t^{\text{RF++}} &= r(x,y) - \beta\sum_{i=t}^{T}\mathbb{D}_{\mathrm{KL}}\left(\pi_{\theta}\parallel\pi_{\mathrm{ref}}\right),\\
\label{equation:reinforce_adv}
\hat{A}_{\text{normalized}}^{\text{RF++}} &= \frac{\hat{A}_t^{\text{RF++}} - \mu_{\hat{A}}}{\sigma_{\hat{A}}},
\end{align}
where \(\mu_{\hat{A}}\) and \(\sigma_{\hat{A}}\) represent the mean and standard deviation of advantages within each batch respectively and $\mathbb{D}_{\mathrm{KL}}\left(\pi_{\theta}\parallel\pi_{\mathrm{ref}}\right)$ is a token-level Kullback-Leibler divergence penalty between the RL model and the SFT model distributions.

In contrast, GRPO employs a slightly modified objective, where the KL divergence term is explicitly separated from the advantage estimate:

\begin{equation}
    L_t^{\text{GRPO}}(\theta) = -\hat{\mathbb{E}}_t\left[
        \max\left(
            \min\left(
                r_t(\theta)\hat{A}_t,\,
                \text{clip}\left(r_t(\theta), 1-\epsilon_{\text{low}}, 1+\epsilon_{\text{high}}\right)\hat{A}_t
            \right),\,
            c\hat{A}_t
        \right) - \beta \mathbb{D}_{\mathrm{KL}}\left(\pi_{\theta}\parallel\pi_{\mathrm{ref}}\right) \right],
\end{equation}

where GRPO's advantage estimation \(\hat{A}_{t}\) is computed via group-relative normalization of rewards that has the similar form as Equation~\ref{equation:reinforce_adv}.

Our default hyperparameter settings for RL training are as follows:
\begin{itemize}
    \item[$\bullet$] Clipping lower bound parameter: $c = 3.0$
    \item[$\bullet$] PPO clip parameters: $\epsilon_{\text{high}} = 0.2$, $\epsilon_{\text{low}} = 0.2$
    \item[$\bullet$] KL coefficient: $\beta = 0.001$
    \item[$\bullet$] Learning rate schedule: constant
    \item[$\bullet$] Weight decay: $0.01$
    \item[$\bullet$] Rollout temperature: $1.0$
    \item[$\bullet$] GAE parameters (Equation~\ref{equation:gae}): discount factor $\lambda = 0.99$, balance factor $\gamma = 0.9$
    \item[$\bullet$] Critic network learning rate: $1\times10^{-5}$
\end{itemize}

\section{More Experimental Results}
\label{appendix:more_experimental_results}

\begin{figure}[ht]
  \centering
  \begin{subfigure}[b]{0.48\textwidth}
    \centering
    \includegraphics[width=\linewidth]{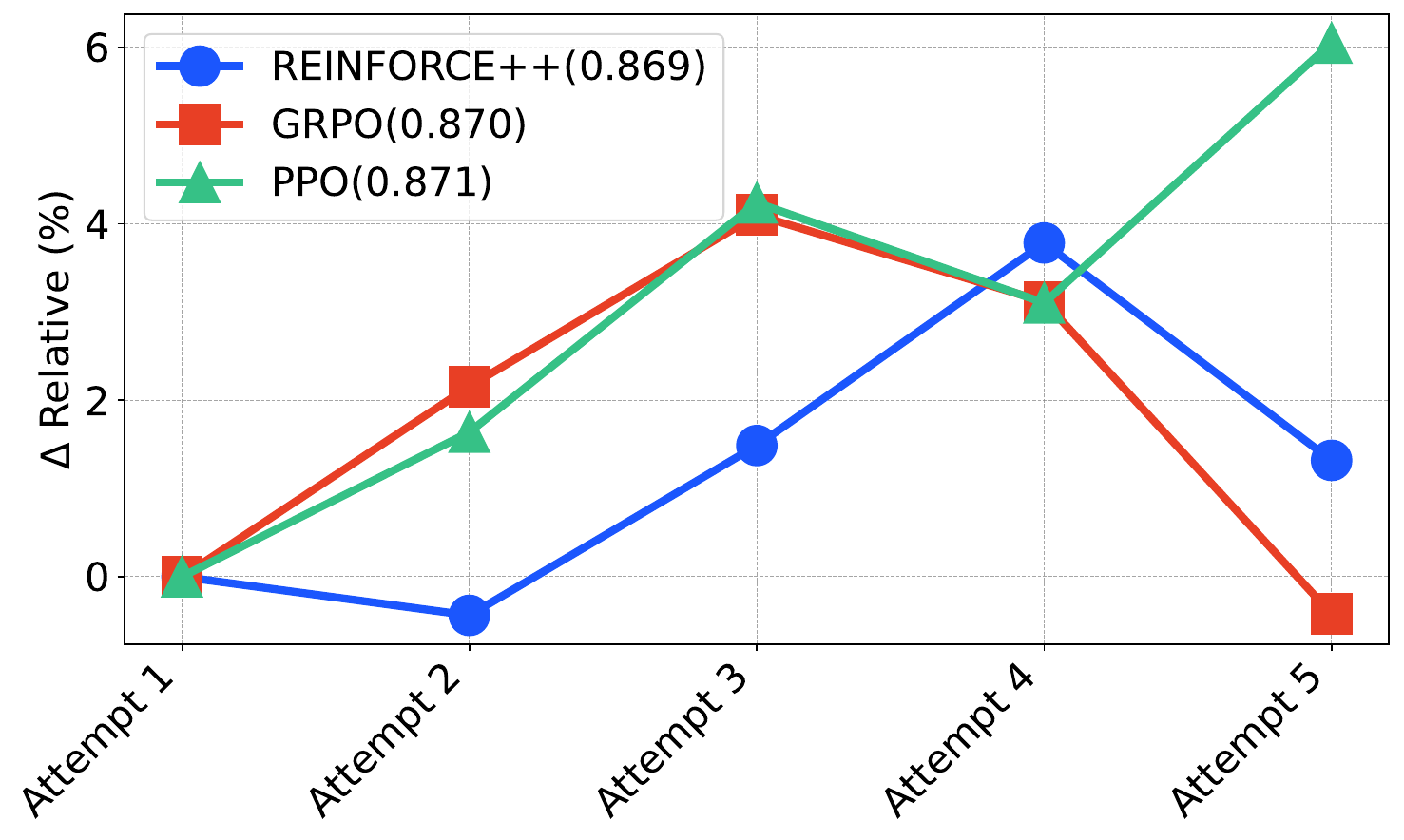}
    \caption{RewardBench ($\Delta$ Relative)}
    \label{fig:algorithms_ablation/rewardbench/plot_exp2_algorithms}
  \end{subfigure}
  \hfill
  \begin{subfigure}[b]{0.48\textwidth}
    \centering
    \includegraphics[width=\linewidth]{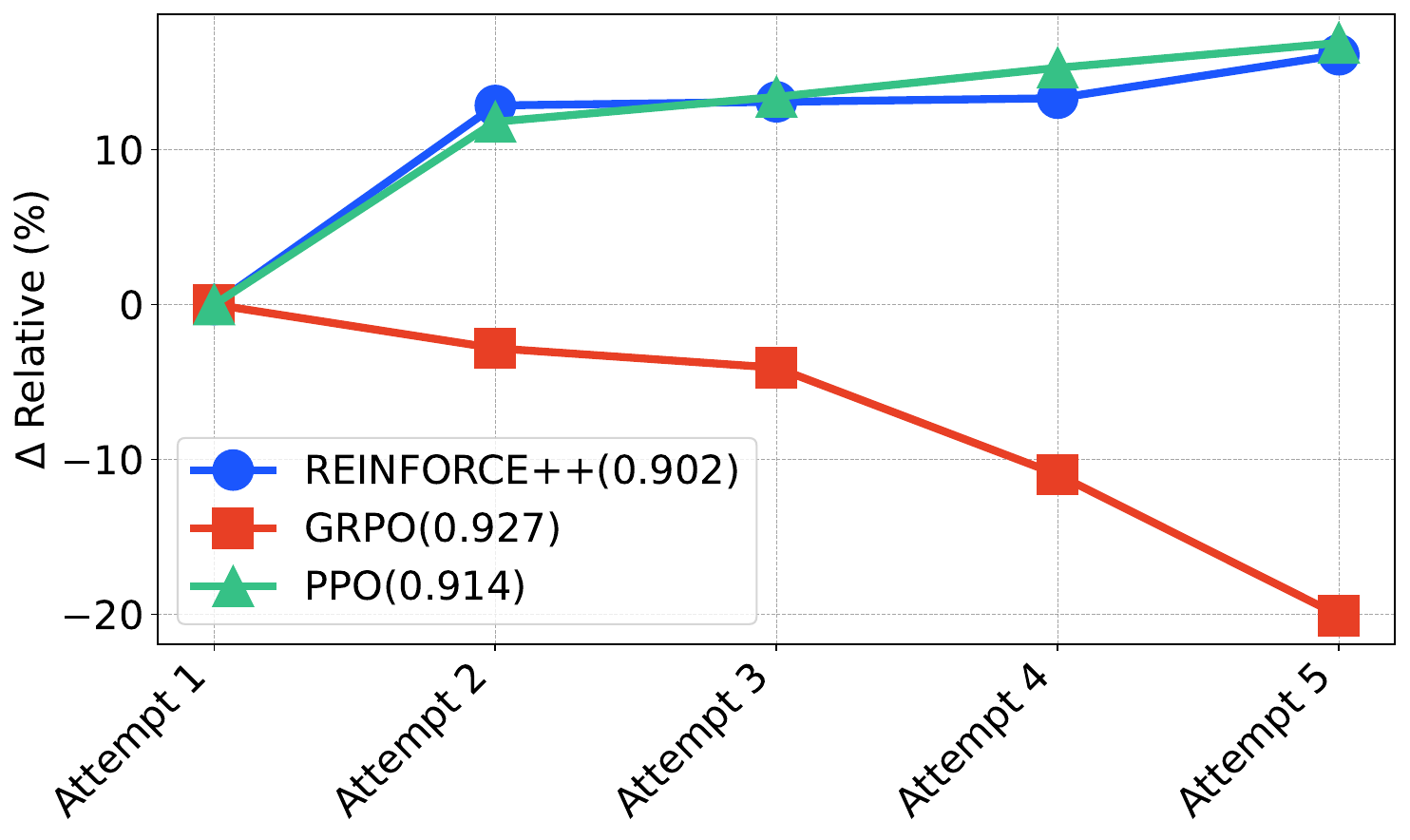}
    \caption{RewardMATH ($\Delta$ Relative)}
    \label{fig:algorithms_ablation/rewardmath/plot_exp2_algorithms}
  \end{subfigure}
  
  \vspace{0.5cm} 
  
  \begin{subfigure}[b]{0.48\textwidth}
    \centering
    \includegraphics[width=\linewidth]{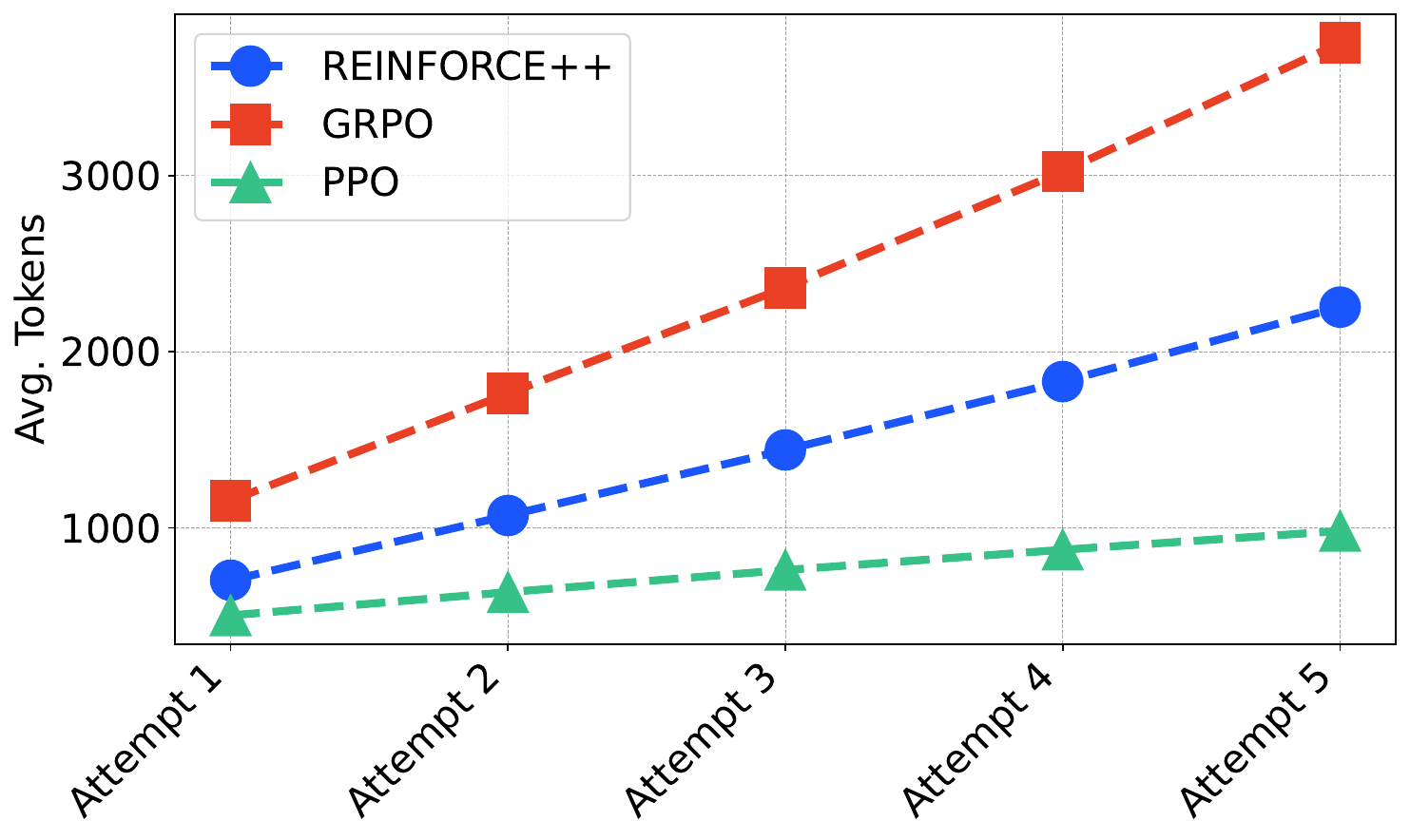}
    \caption{RewardBench (Avg. Tokens)}
    \label{fig:algorithms_ablation/rewardbench/plot_exp2_algorithms_token_num}
  \end{subfigure}
  \hfill
  \begin{subfigure}[b]{0.48\textwidth}
    \centering
    \includegraphics[width=\linewidth]{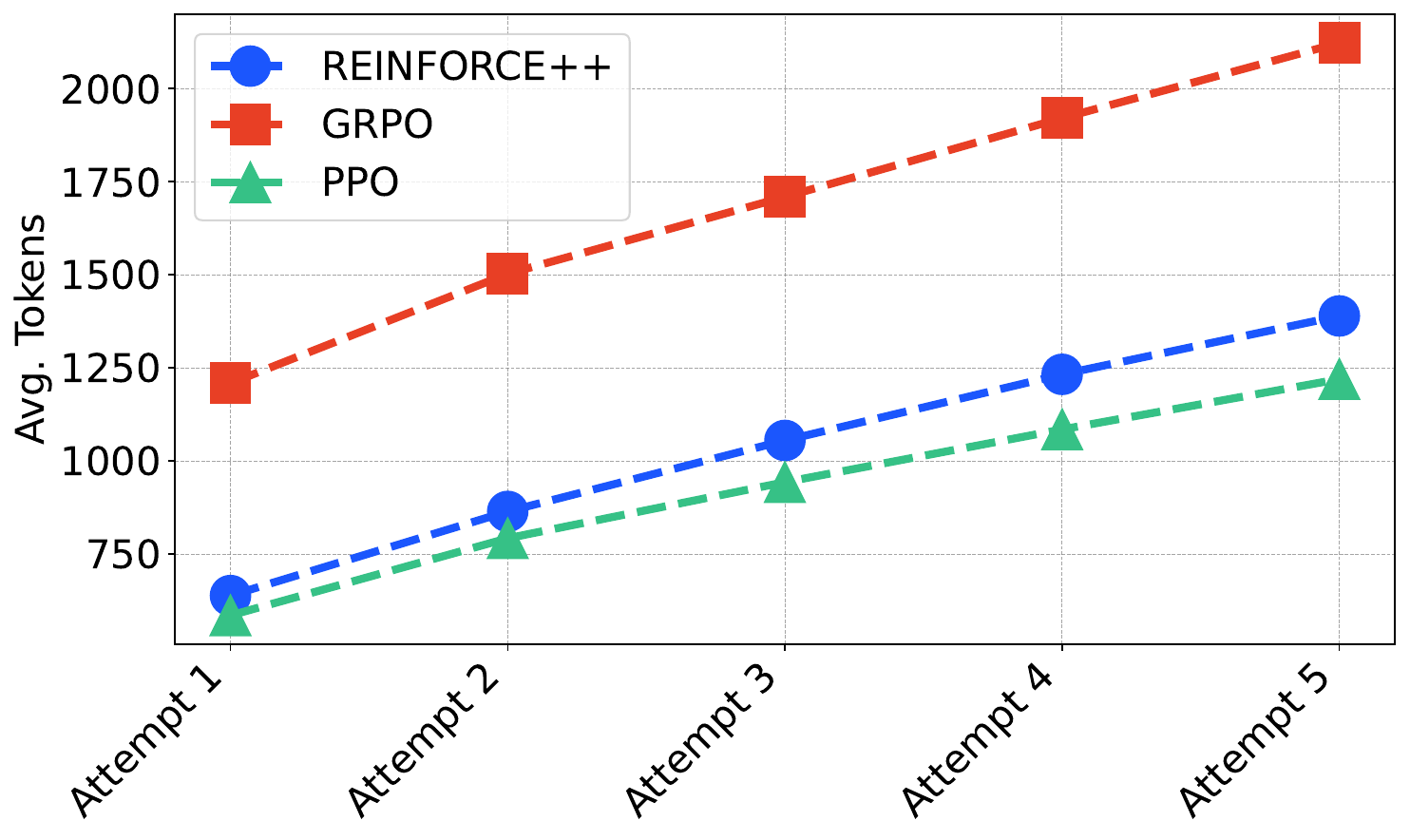}
    \caption{RewardMATH (Avg. Tokens)}
    \label{fig:algorithms_ablation/rewardmath/plot_exp2_algorithms_token_num}
  \end{subfigure}
  
  \caption{Scaling trend for STTS on RewardBench and RewardMATH with different RL algorithms.}
  \label{fig:algorithms_ablation_2}
\end{figure}

\begin{figure}[t]
  \centering
  \begin{subfigure}[b]{0.32\textwidth}
    \centering
    \includegraphics[width=\linewidth]{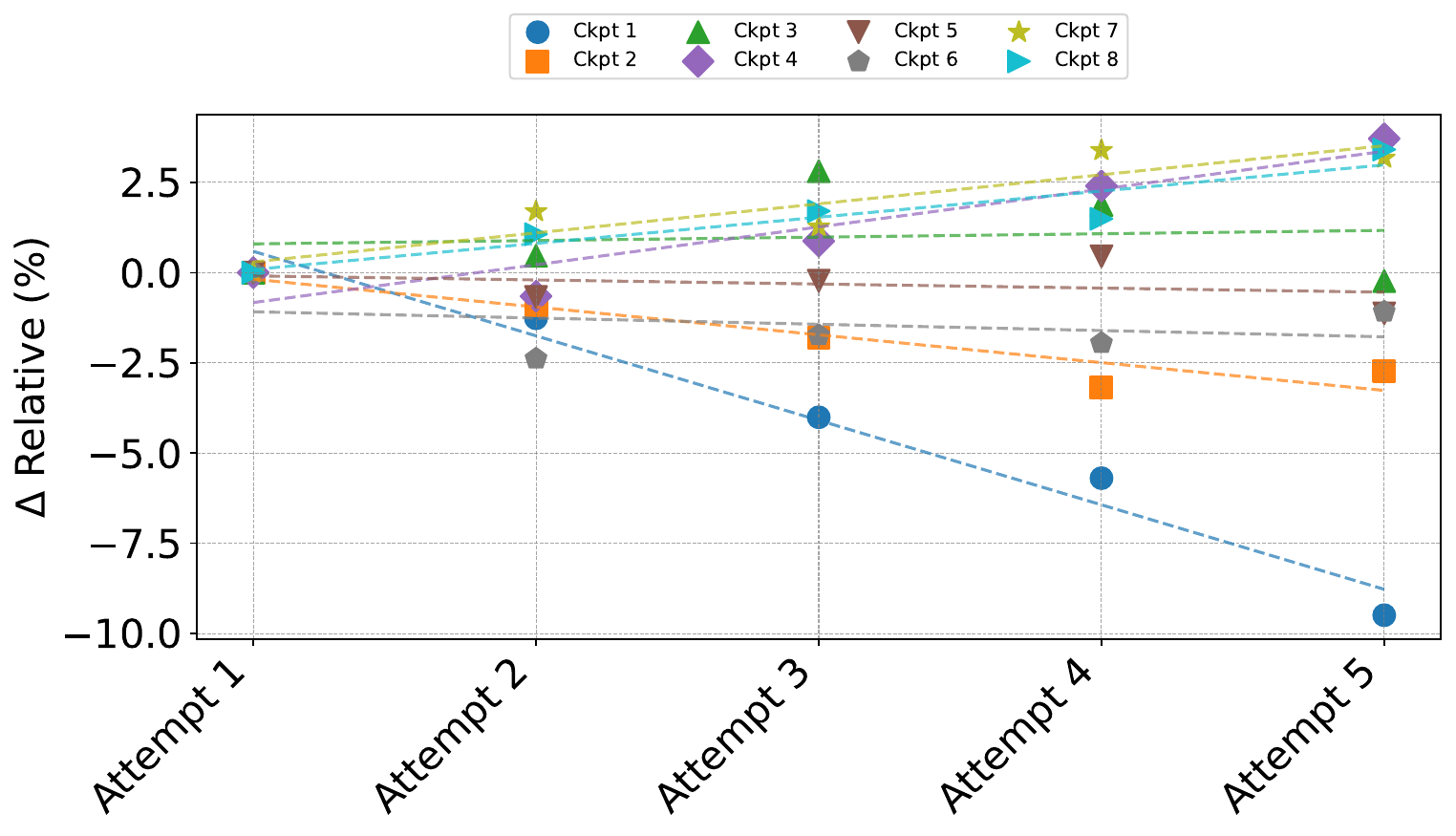}
    \caption{}
    \label{fig:step_progress_codeprefbench_1}
  \end{subfigure}
  \hfill
  \begin{subfigure}[b]{0.32\textwidth}
    \centering
    \includegraphics[width=\linewidth]{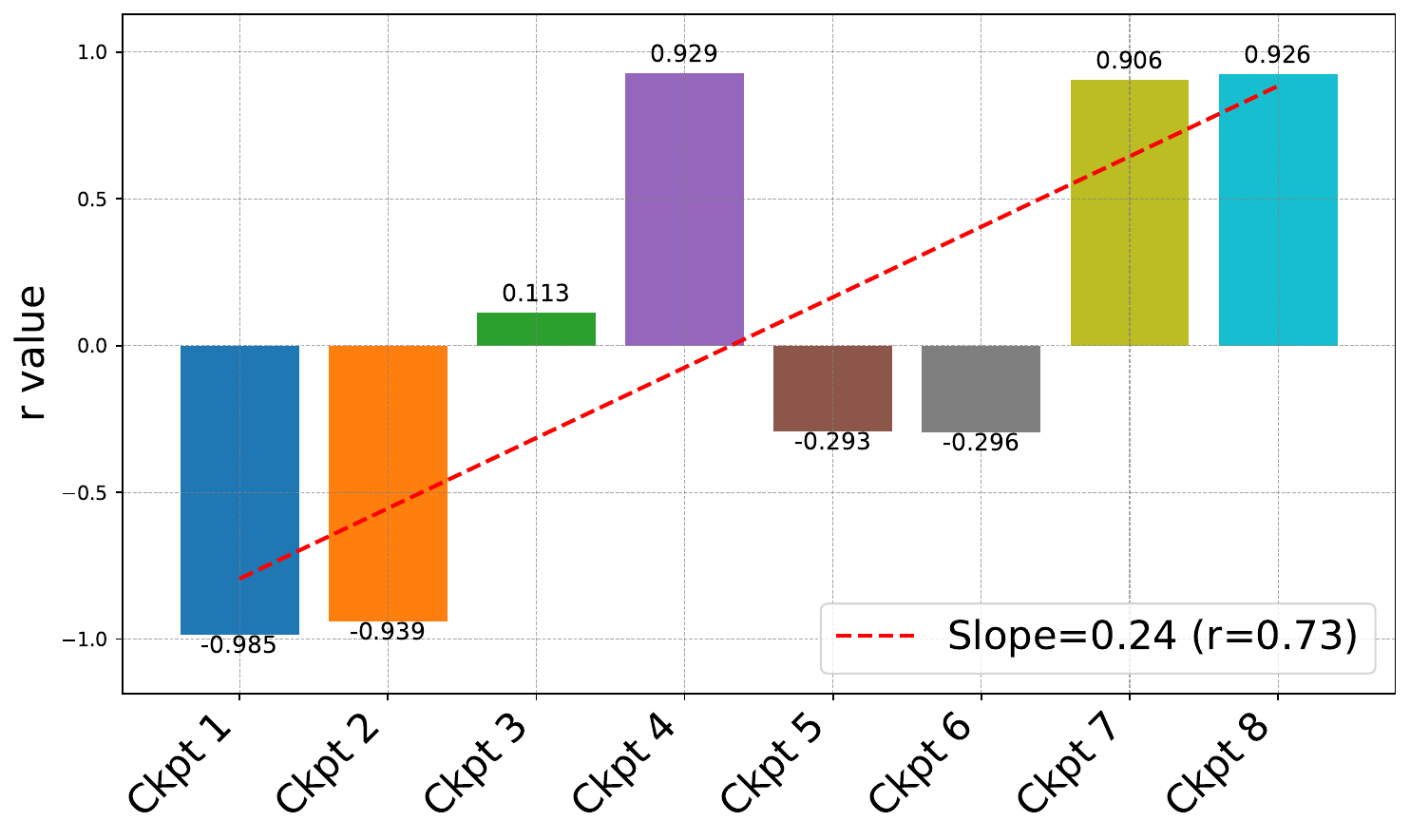}
    \caption{}
    \label{fig:step_progress_codeprefbench_2}
  \end{subfigure}
  \hfill
  \begin{subfigure}[b]{0.32\textwidth}
    \centering
    \includegraphics[width=\linewidth]{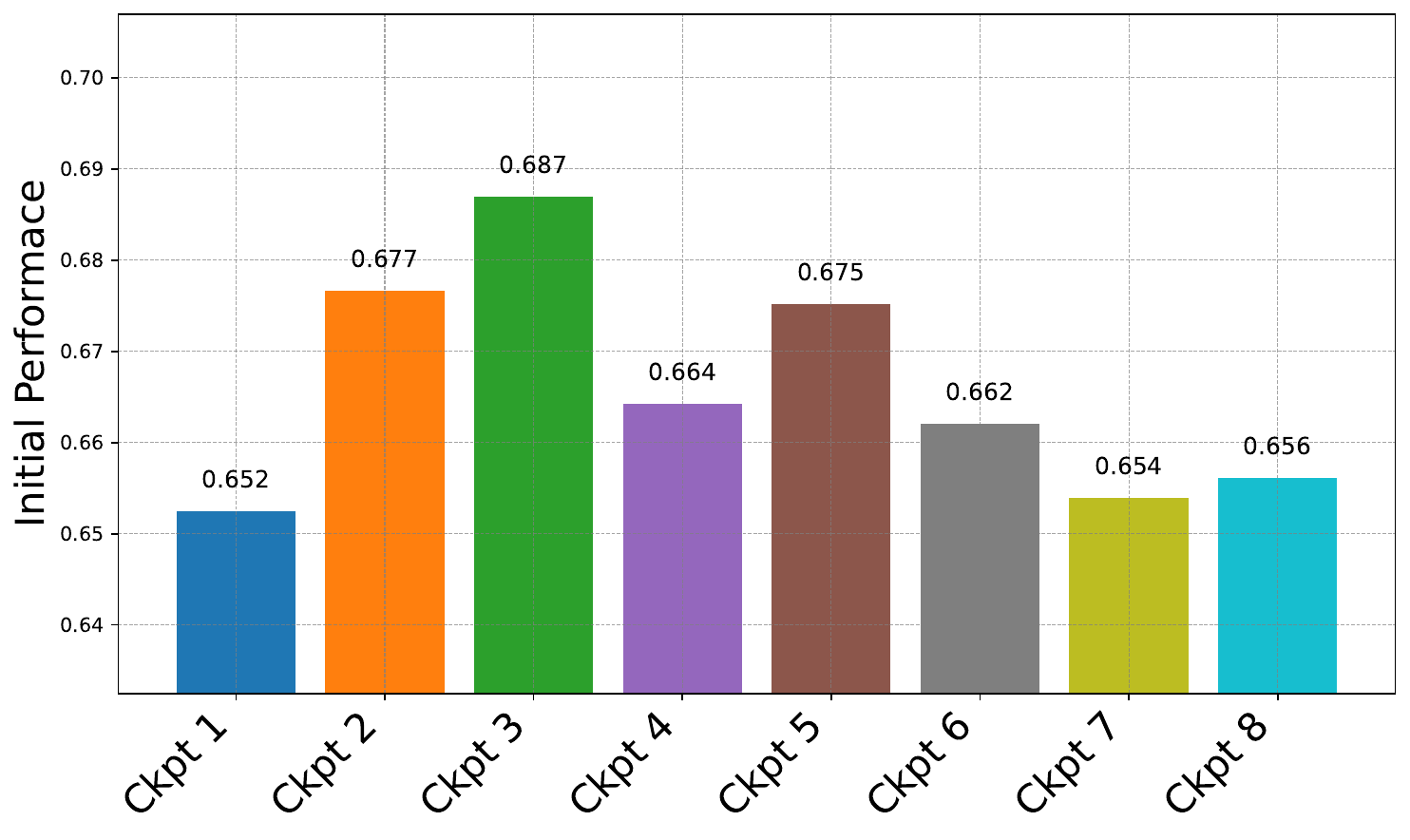}
    \caption{}
    \label{fig:step_progress_codeprefbench_3}
  \end{subfigure}
  
  \caption{Scaling behaviour of different checkpoints on CodePrefBench.}
  \label{fig:step_progress_codeprefbench}
\end{figure}

\begin{figure}[ht]
  \centering
  \begin{subfigure}[b]{0.32\textwidth}
    \centering
    \includegraphics[width=\linewidth]{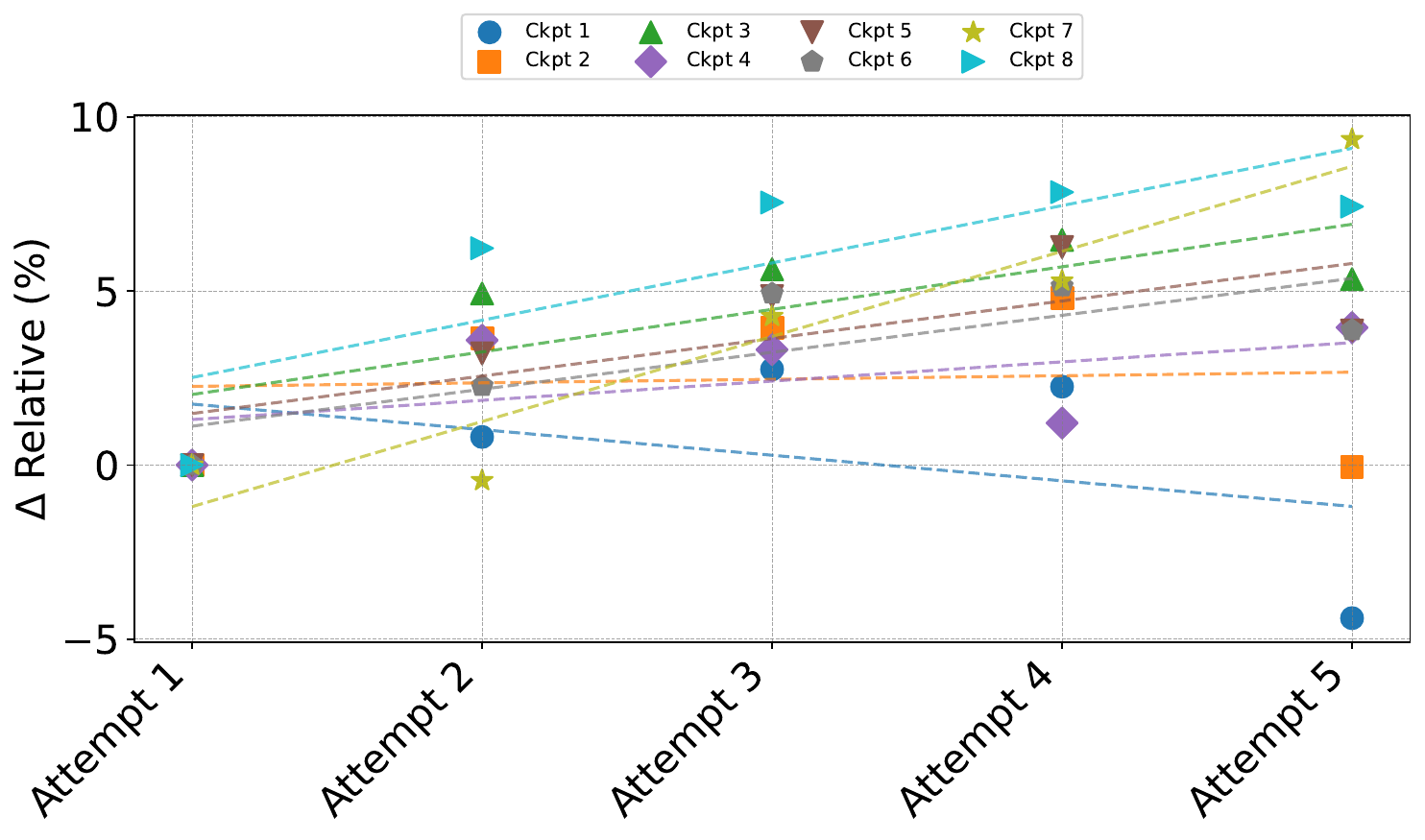}
    \caption{}
    \label{fig:step_progress_rewardbench_1}
  \end{subfigure}
  \hfill
  \begin{subfigure}[b]{0.32\textwidth}
    \centering
    \includegraphics[width=\linewidth]{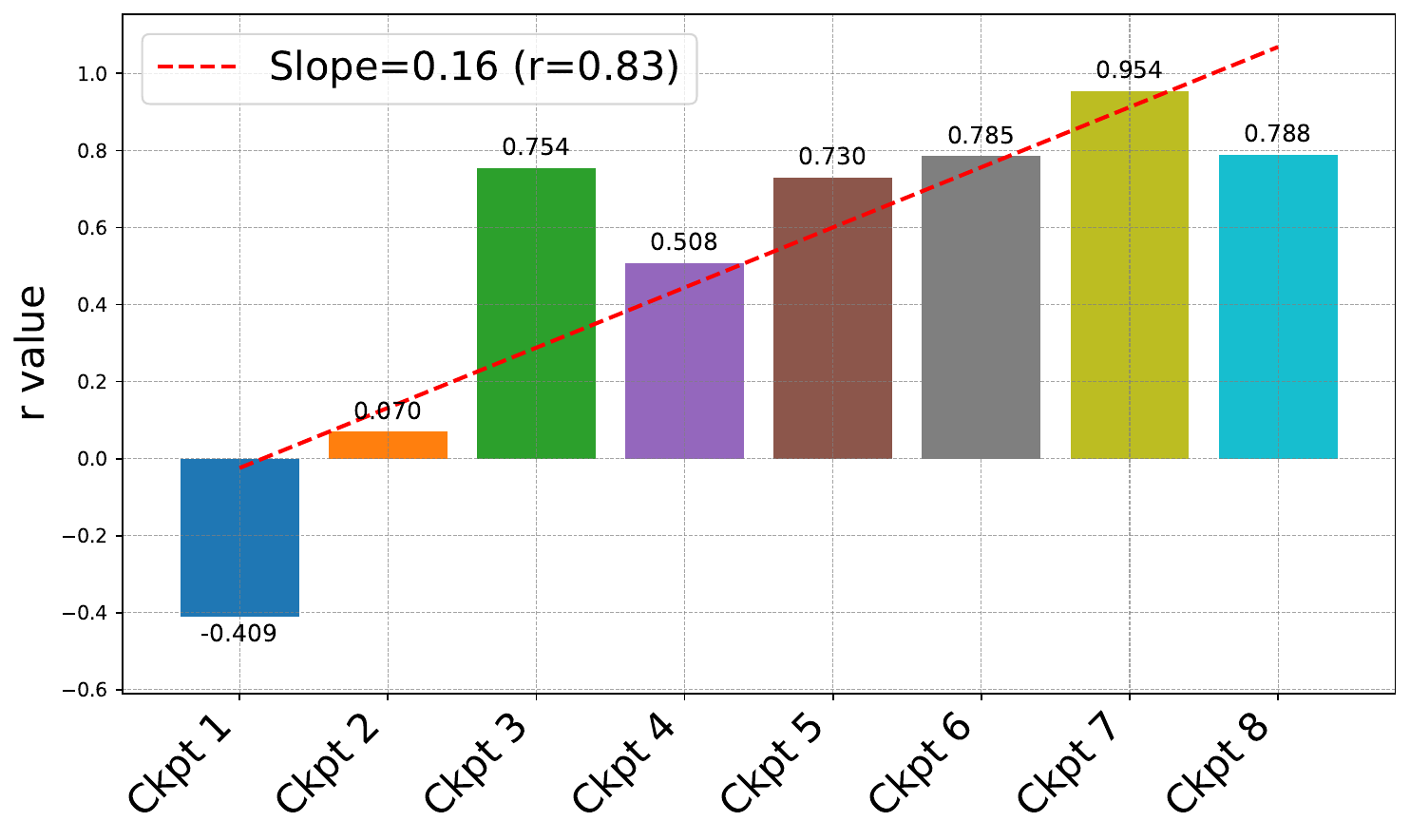}
    \caption{}
    \label{fig:step_progress_rewardbench_2}
  \end{subfigure}
  \hfill
  \begin{subfigure}[b]{0.32\textwidth}
    \centering
    \includegraphics[width=\linewidth]{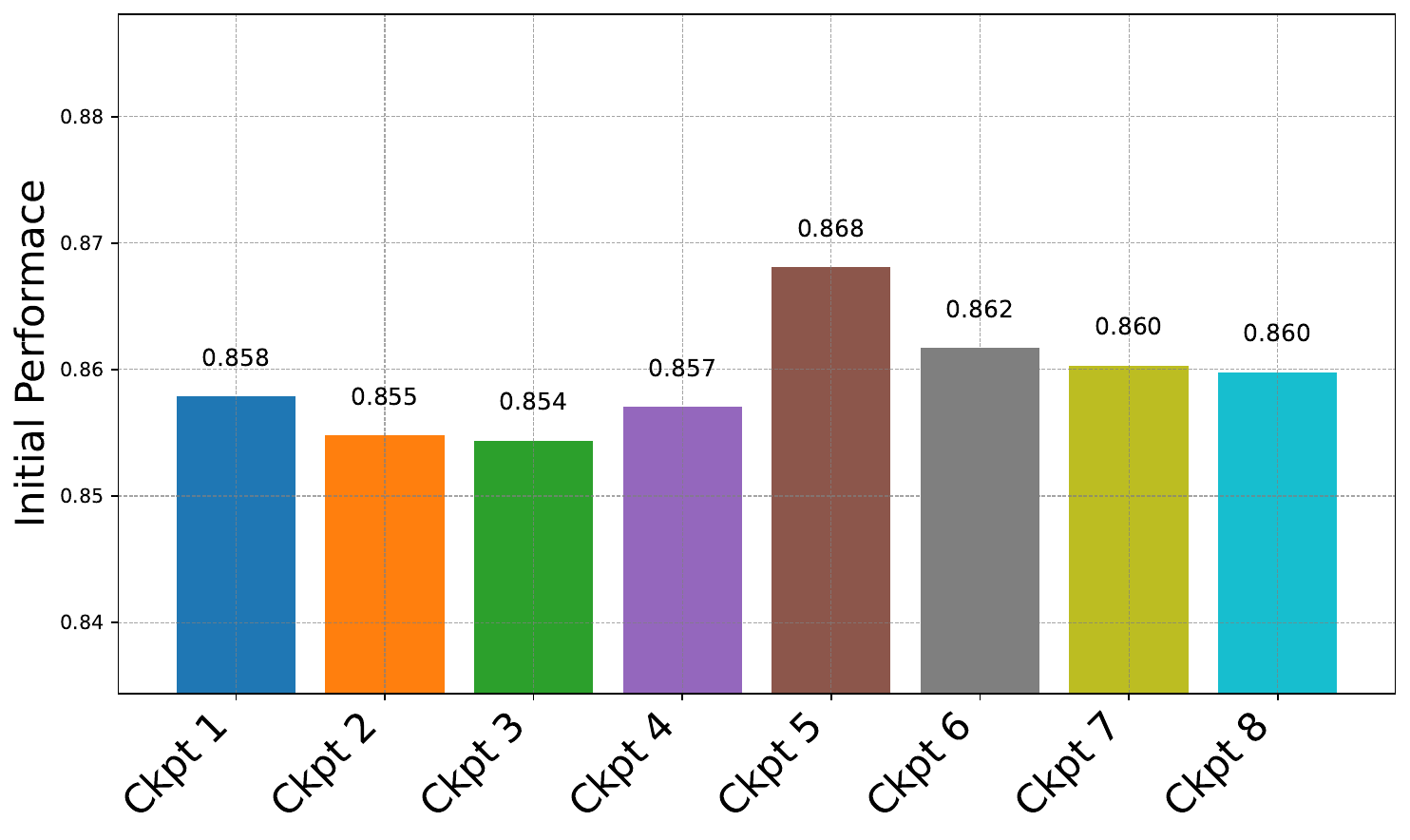}
    \caption{}
    \label{fig:step_progress_rewardbench_3}
  \end{subfigure}
  
  \caption{Scaling behaviour of different checkpoints on RewardBench.}
  \label{fig:step_progress_rewardbench}
\end{figure}

\begin{figure}[ht]
  \centering
  \begin{subfigure}[b]{0.32\textwidth}
    \centering
    \includegraphics[width=\linewidth]{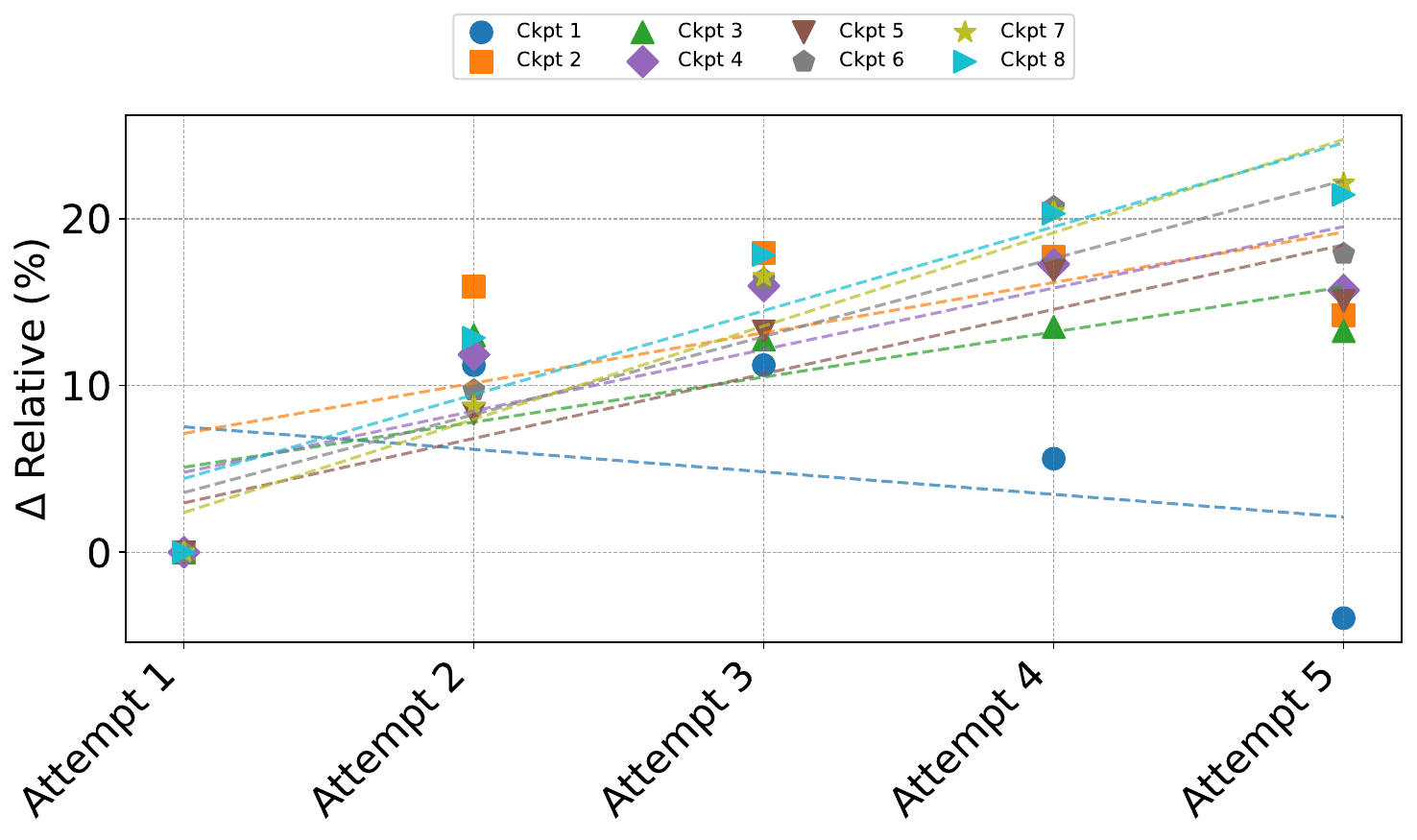}
    \caption{}
    \label{fig:step_progress_rewardmath_1}
  \end{subfigure}
  \hfill
  \begin{subfigure}[b]{0.32\textwidth}
    \centering
    \includegraphics[width=\linewidth]{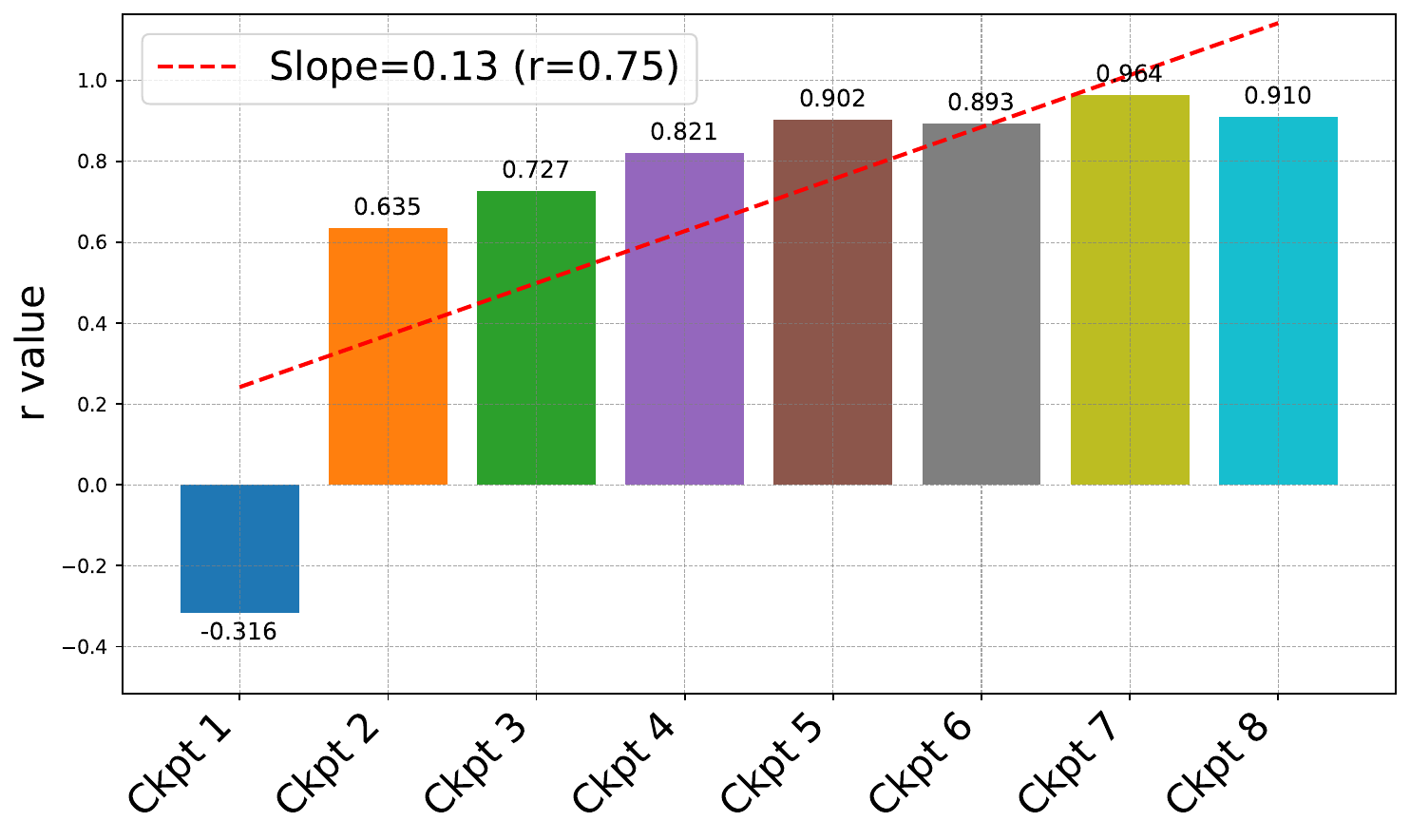}
    \caption{}
    \label{fig:step_progress_rewardmath_2}
  \end{subfigure}
  \hfill
  \begin{subfigure}[b]{0.32\textwidth}
    \centering
    \includegraphics[width=\linewidth]{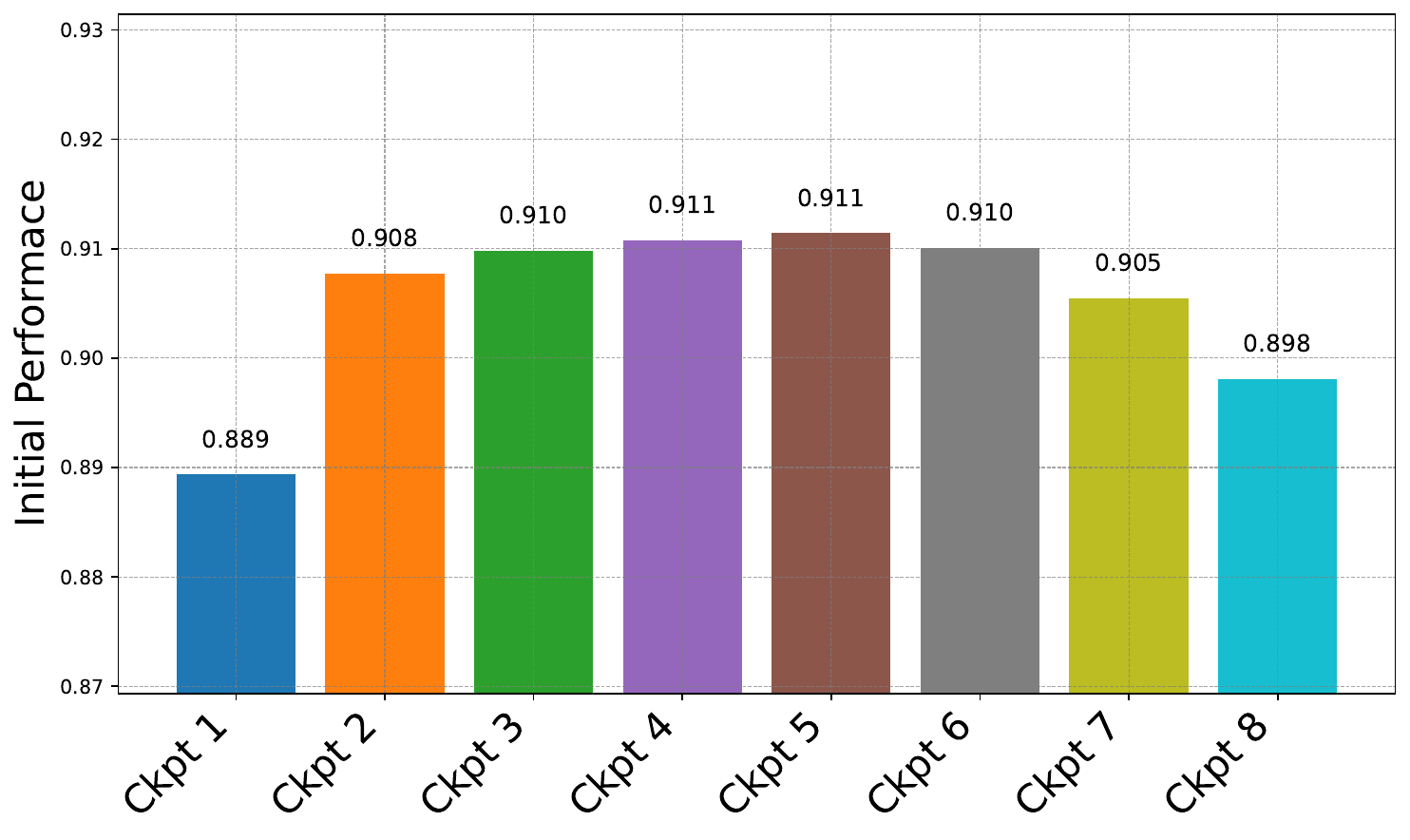}
    \caption{}
    \label{fig:step_progress_rewardmath_3}
  \end{subfigure}
  
  \caption{Scaling behaviour of different checkpoints on RewardMATH.}
  \label{fig:step_progress_rewardmath}
\end{figure}

In this section, we present additional experimental results, including analyses conducted on extended datasets under the experimental settings described in Sections~\ref{subsec:different_rl_algorithms} and~\ref{subsec:step_increase}. Our findings consistently support previous observations. As depicted in Figure~\ref{fig:algorithms_ablation_2}, checkpoints trained with different RL algorithms initially achieve comparable performance. However, the GRPO-trained models, due to inherently longer reasoning sequences, demonstrate slightly diminished effectiveness when applying STTS. Further, the progression of STTS effectiveness on additional datasets, illustrated in Figures~\ref{fig:step_progress_codeprefbench_2},~\ref{fig:step_progress_rewardbench_2} and~\ref{fig:step_progress_rewardmath_2}, confirms our hypothesis: models gradually acquire the capability to effectively utilize STTS during the RL process.

\subsection{Different Data Mixture During RL Elicit Different STTS behavior}
\label{appendix:different_data_mixture}

In this section, we investigate the impact of different data mixtures on STTS during the RL post-training phase. Specifically, we consider four distinct data mixtures: (1) \textbf{Mixture 1:} the original mixture used in our primary experiments; (2) \textbf{Mixture 2:} the original mixture enhanced with additional safety and code-related datasets (PKU-SafeRLHF~\citep{ji2024pku} and CodeUltraFeedback~\citep{weyssow2024codeultrafeedback}); (3) \textbf{Mixture 3:} a further extension on (2) incorporating instances that consistently remained incorrect after three STTS attempts during the SFT data construction phase; and (4) \textbf{Mixture 4:} the full RISE dataset, which includes additional Chinese-language samples.

In this experiment, we employ Reinforce++ and utilize the checkpoint obtained from the mathematics/code cold-start scenario outlined in Section~\ref{subsec:cold_start}. Our key observation is that variations in data mixture introduce greater variance in outcomes compared to the choice of different RL algorithms applied to identical datasets. Additionally, different data mixtures result in notable differences in both initial performance and subsequent STTS effectiveness. Notably, we find that incorporating instances that persistently fail under multiple STTS attempts leads to a deterioration in both final checkpoint performance and subsequent STTS effectiveness, despite the increased volume of training data. We hypothesize that this occurs because RL improvements depend significantly on successful exploration; persistently incorrect samples likely represent overly challenging instances for the model, thus constraining its learning progress. Future research could beneficially explore strategies for optimally balancing sample difficulty to enhance model training effectiveness.

\begin{figure}[htbp!]
  \centering
  \begin{subfigure}[b]{0.48\textwidth}
    \centering
    \includegraphics[width=\linewidth]{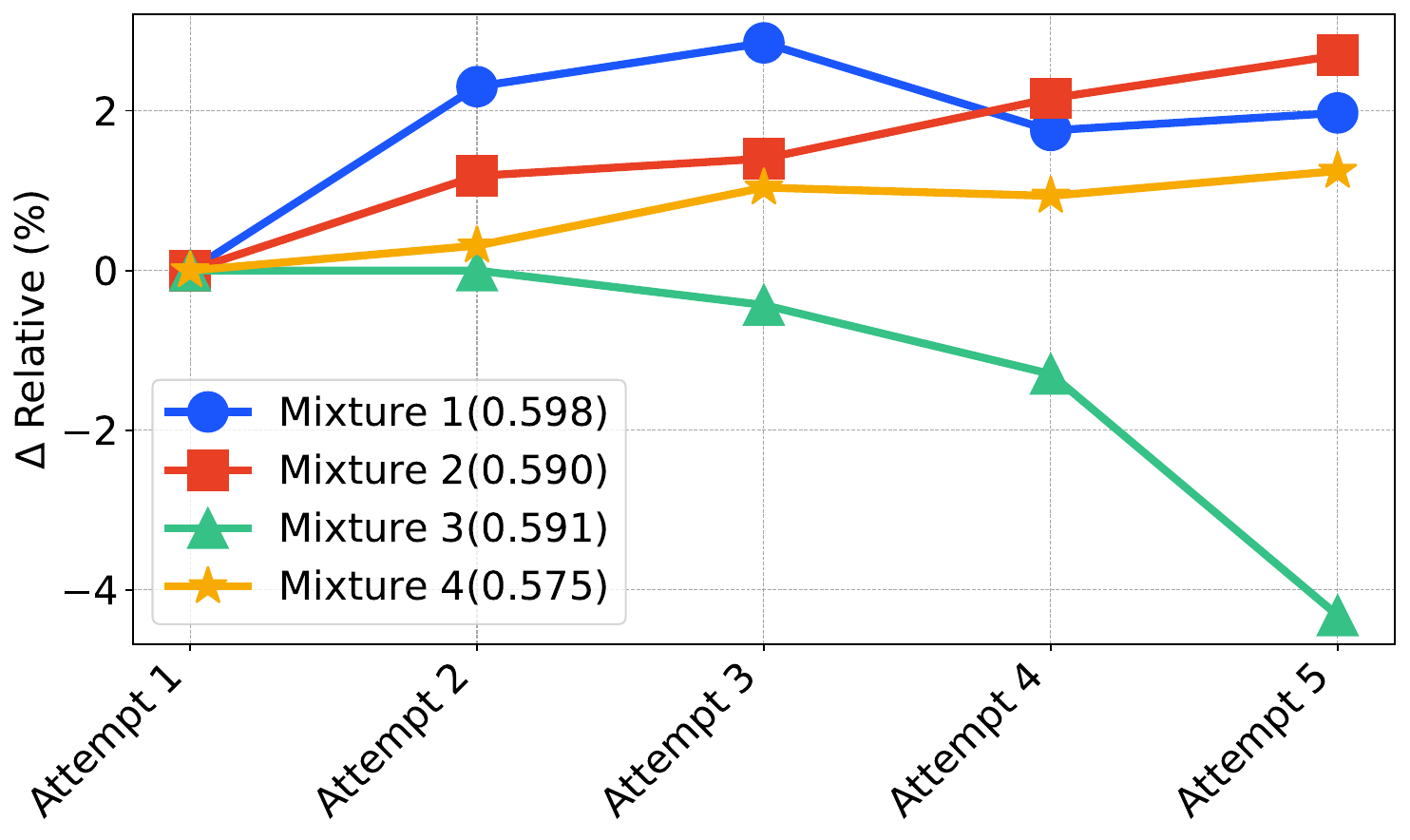}
    \caption{Anthropic Harmless ($\Delta$ Relative)}
    \label{fig:algorithms_ablation/anthropic_harmless/plot_exp3_data_mixture}
  \end{subfigure}
  \hfill
  \begin{subfigure}[b]{0.48\textwidth}
    \centering
    \includegraphics[width=\linewidth]{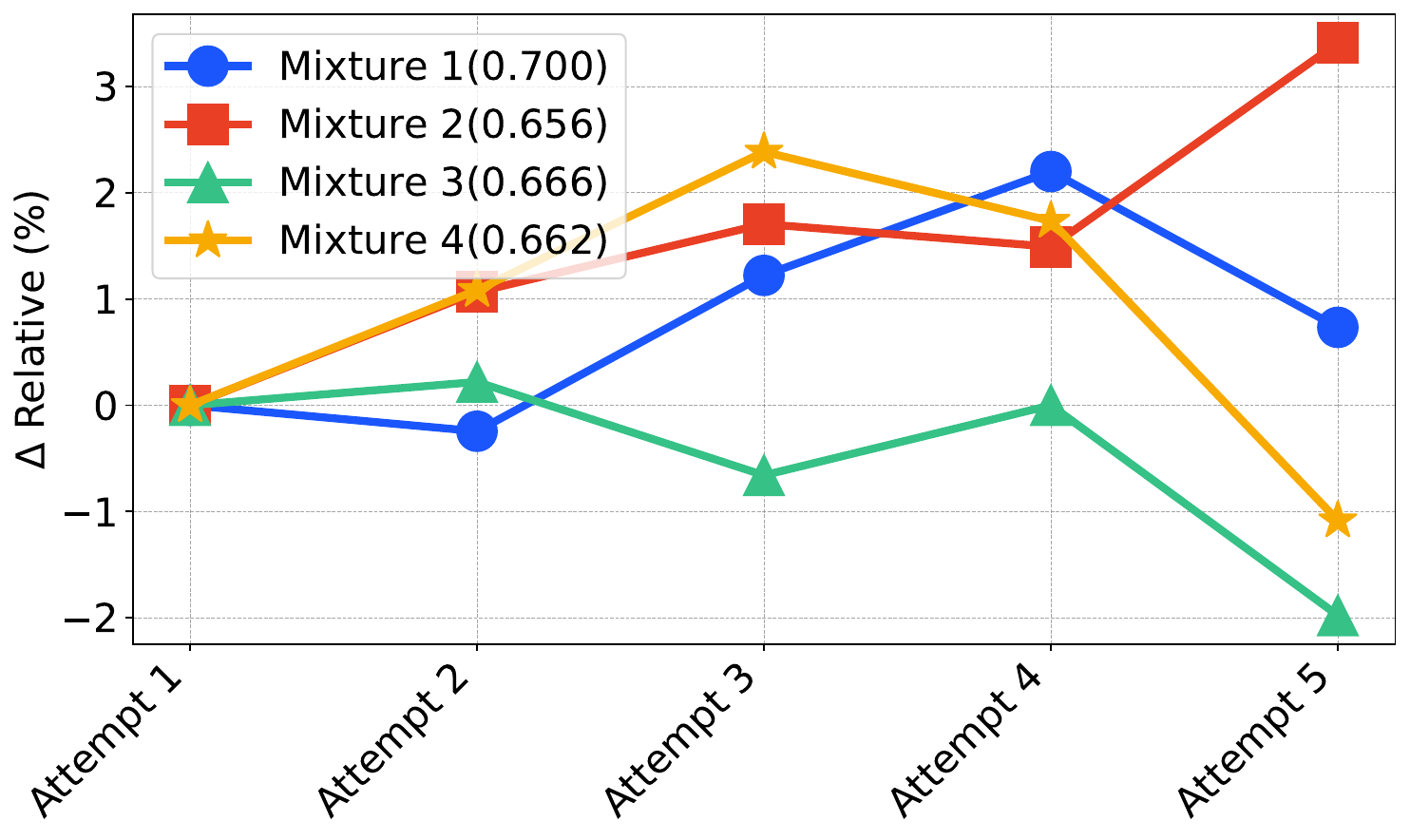}
    \caption{CodePrefBench ($\Delta$ Relative)}
    \label{fig:algorithms_ablation/codeprefbench/plot_exp3_data_mixture}
  \end{subfigure}
  
  \vspace{0.5cm} 
  
  \begin{subfigure}[b]{0.48\textwidth}
    \centering
    \includegraphics[width=\linewidth]{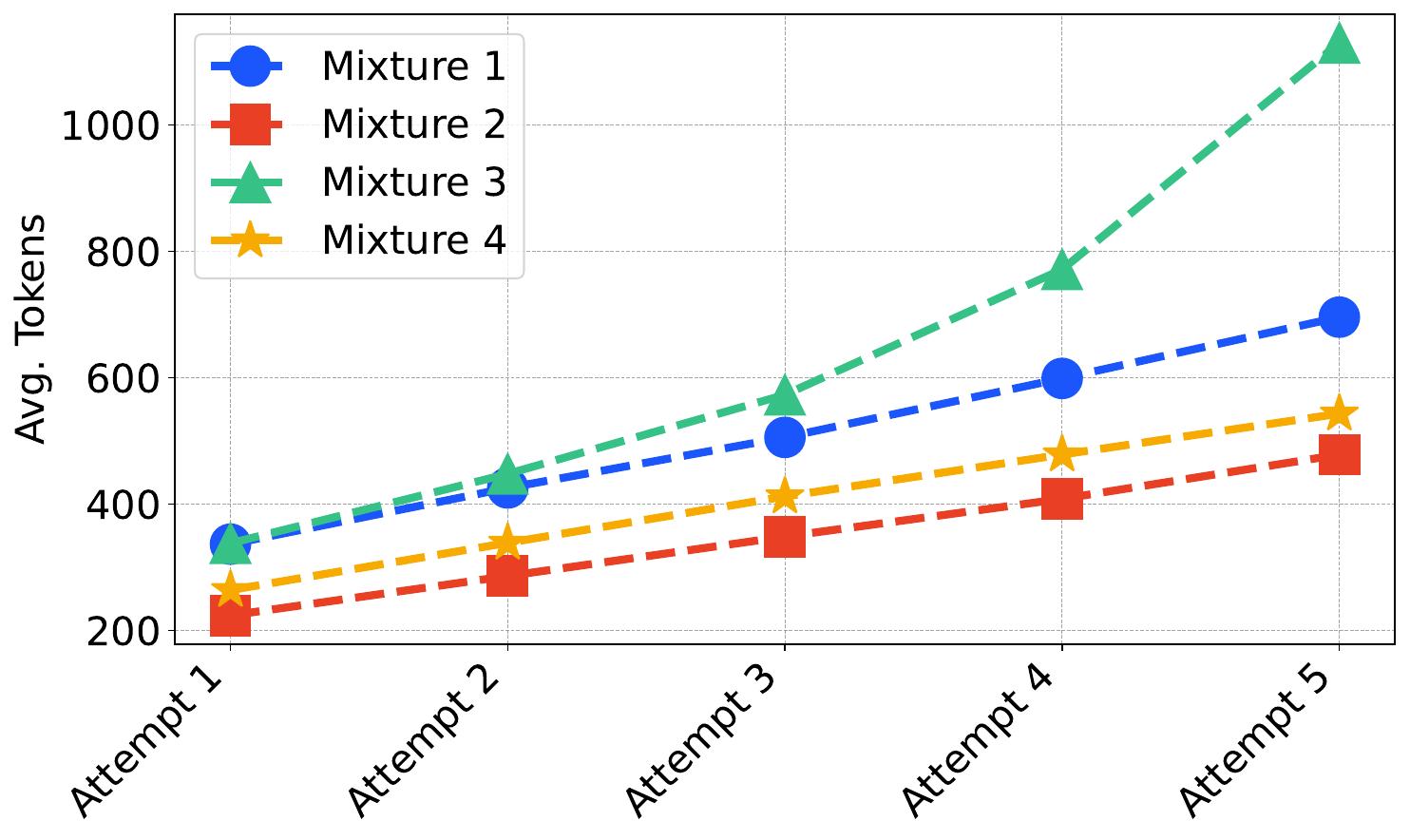}
    \caption{Anthropic Harmless (Avg. Tokens)}
    \label{fig:algorithms_ablation/anthropic_harmless/plot_exp3_data_mixture_token_num}
  \end{subfigure}
  \hfill
  \begin{subfigure}[b]{0.48\textwidth}
    \centering
    \includegraphics[width=\linewidth]{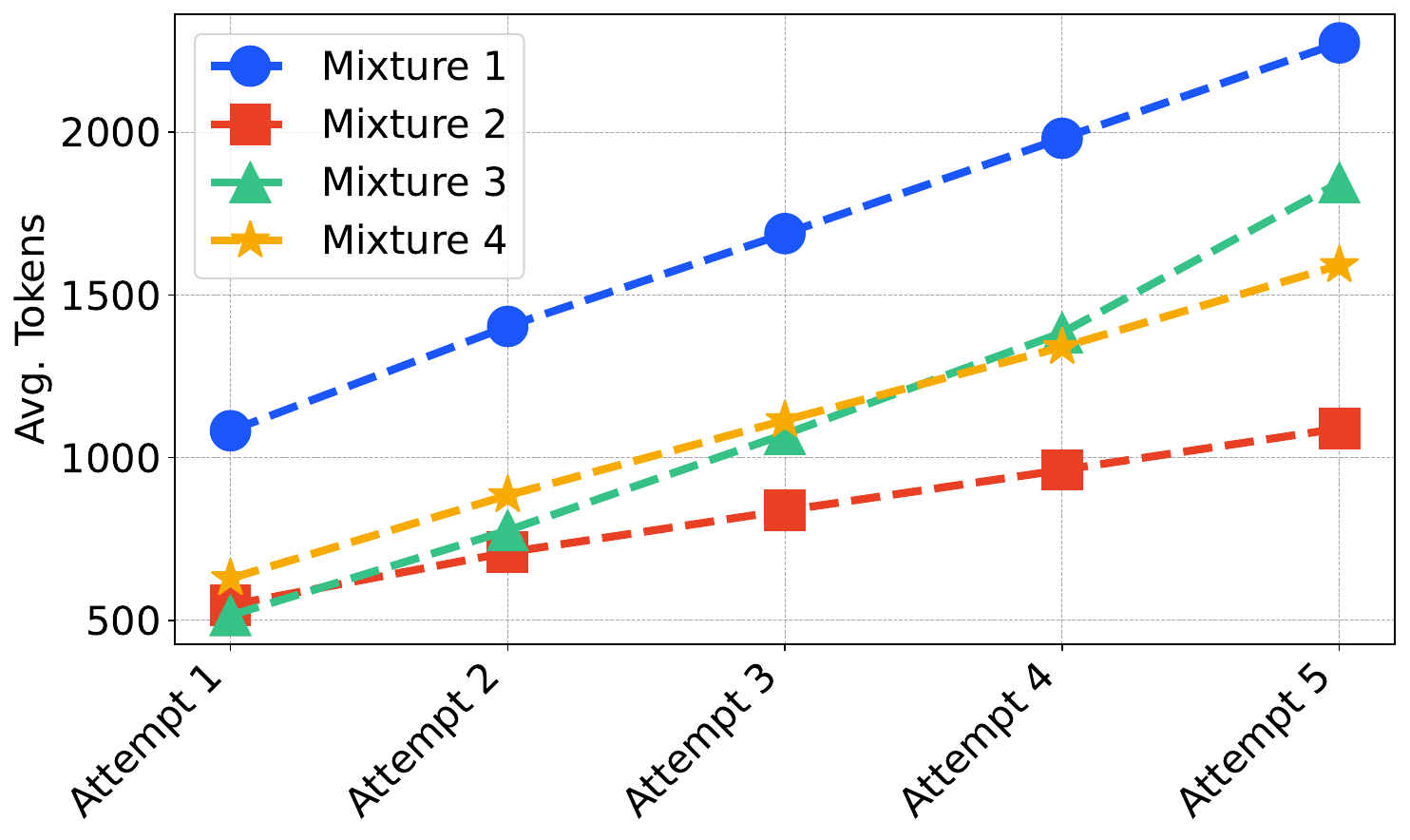}
    \caption{CodePrefBench (Avg. Tokens)}
    \label{fig:algorithms_ablation/codeprefbench/plot_exp3_data_mixture_token_num}
  \end{subfigure}
  
  \caption{Data mixture ablation on Anthropic Harmless and CodePrefBench.}
  \label{fig:data_mixture_ablation_1}
\end{figure}

\begin{figure}[htbp!]
  \centering
  \begin{subfigure}[b]{0.48\textwidth}
    \centering
    \includegraphics[width=\linewidth]{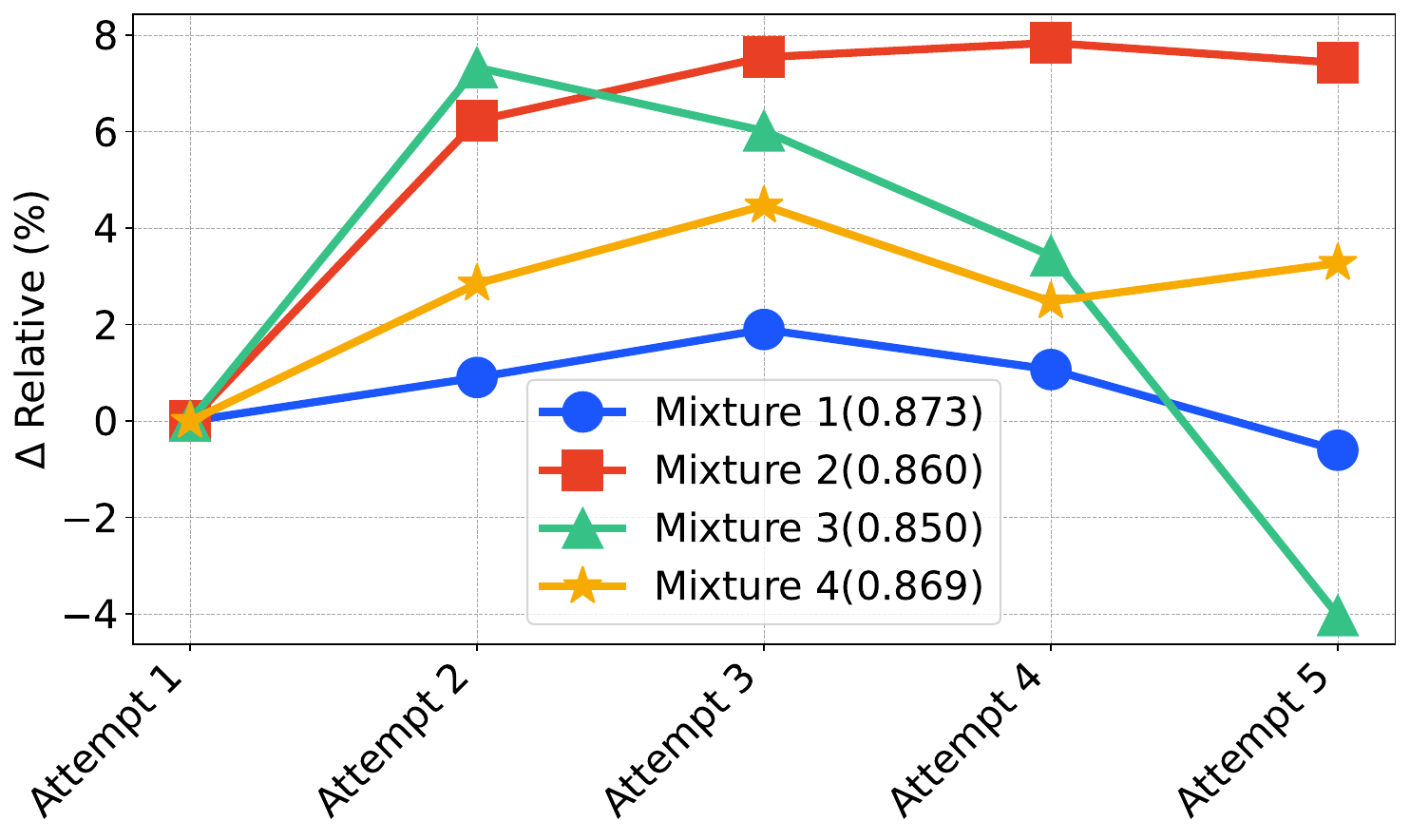}
    \caption{RewardBench ($\Delta$ Relative)}
    \label{fig:algorithms_ablation/rewardbench/plot_exp3_data_mixture}
  \end{subfigure}
  \hfill
  \begin{subfigure}[b]{0.48\textwidth}
    \centering
    \includegraphics[width=\linewidth]{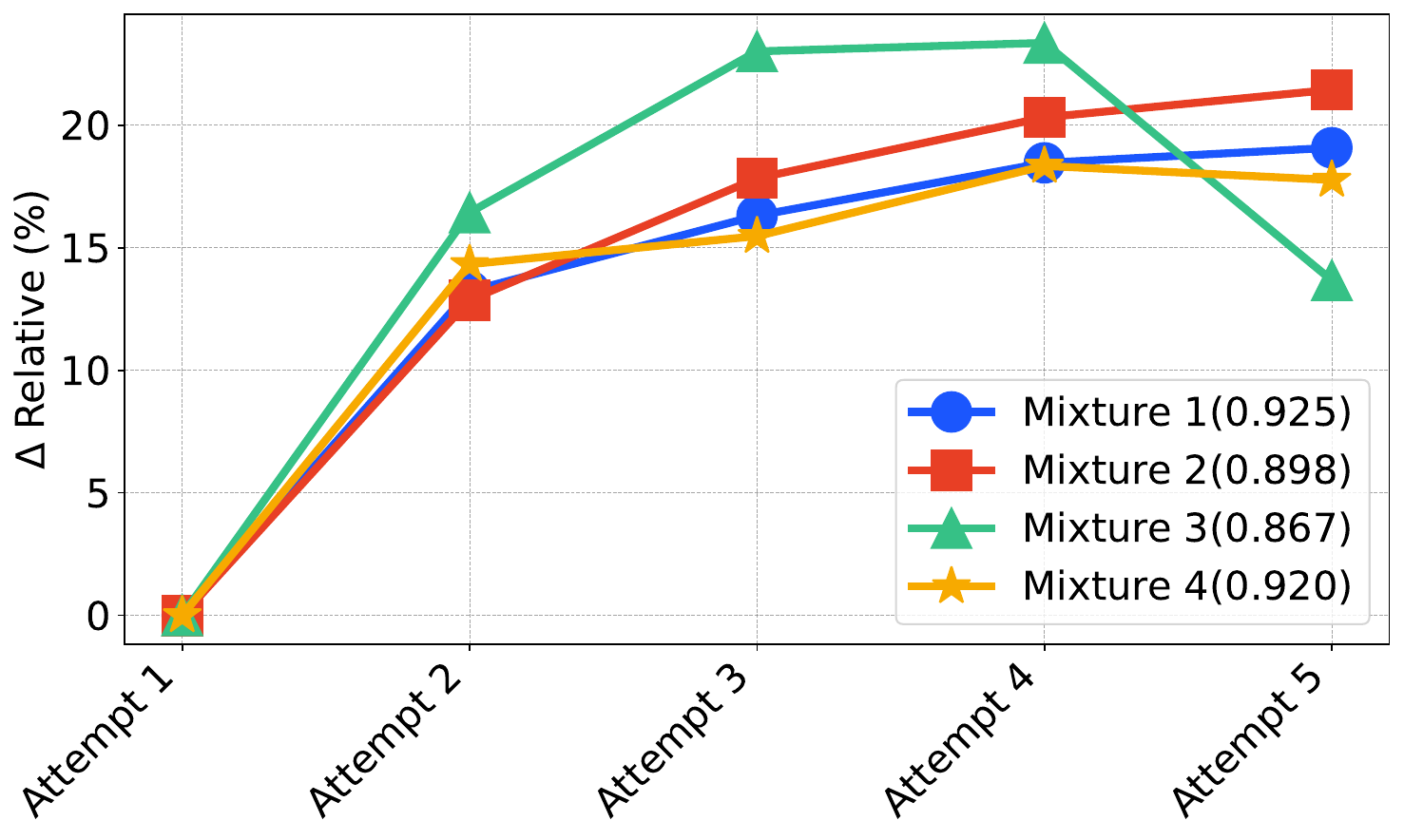}
    \caption{RewardMATH ($\Delta$ Relative)}
    \label{fig:algorithms_ablation/rewardmath/plot_exp3_data_mixture}
  \end{subfigure}
  
  \vspace{0.5cm}
  
  \begin{subfigure}[b]{0.48\textwidth}
    \centering
    \includegraphics[width=\linewidth]{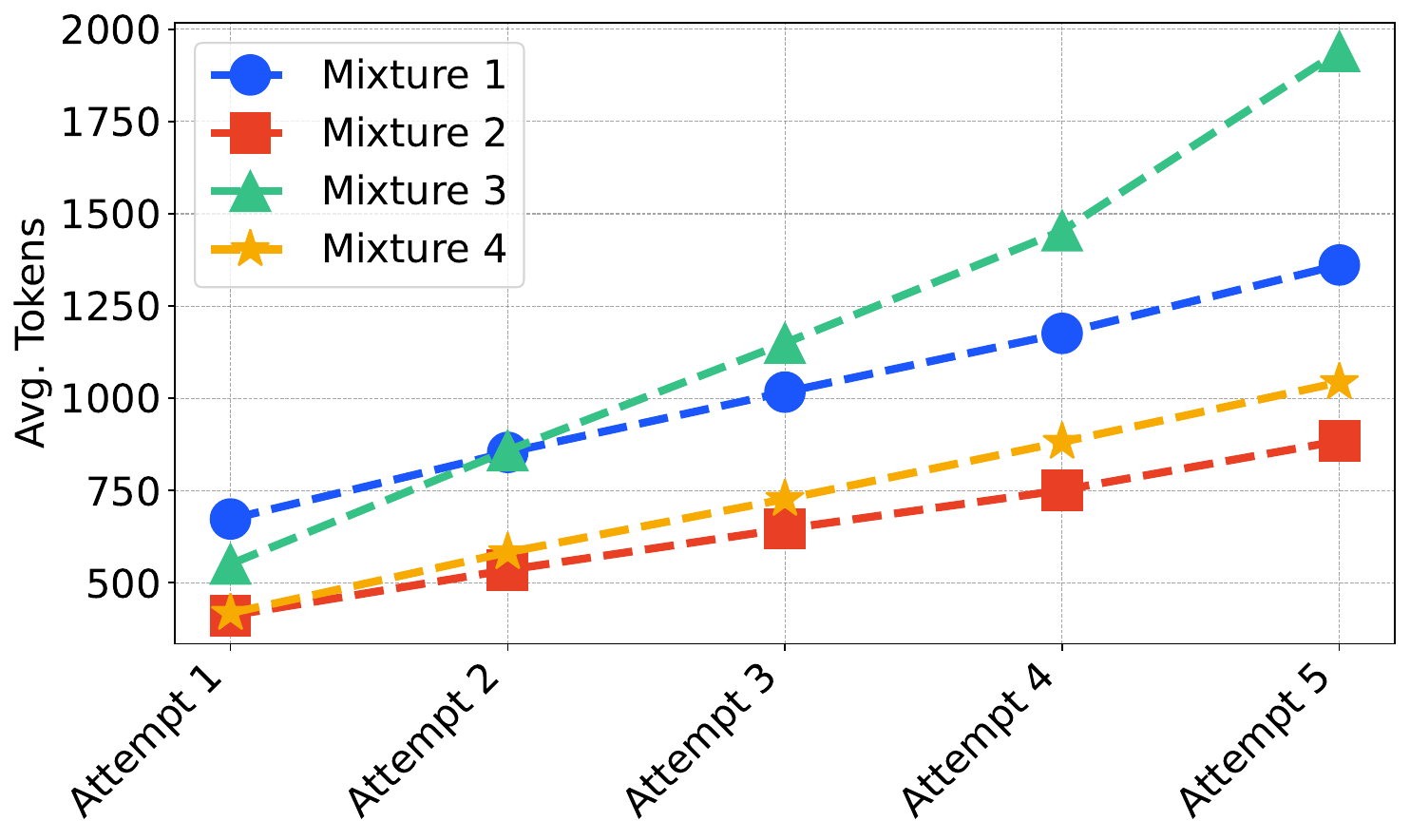}
    \caption{RewardBench (Avg. Tokens)}
    \label{fig:algorithms_ablation/rewardbench/plot_exp3_data_mixture_token_num}
  \end{subfigure}
  \hfill
  \begin{subfigure}[b]{0.48\textwidth}
    \centering
    \includegraphics[width=\linewidth]{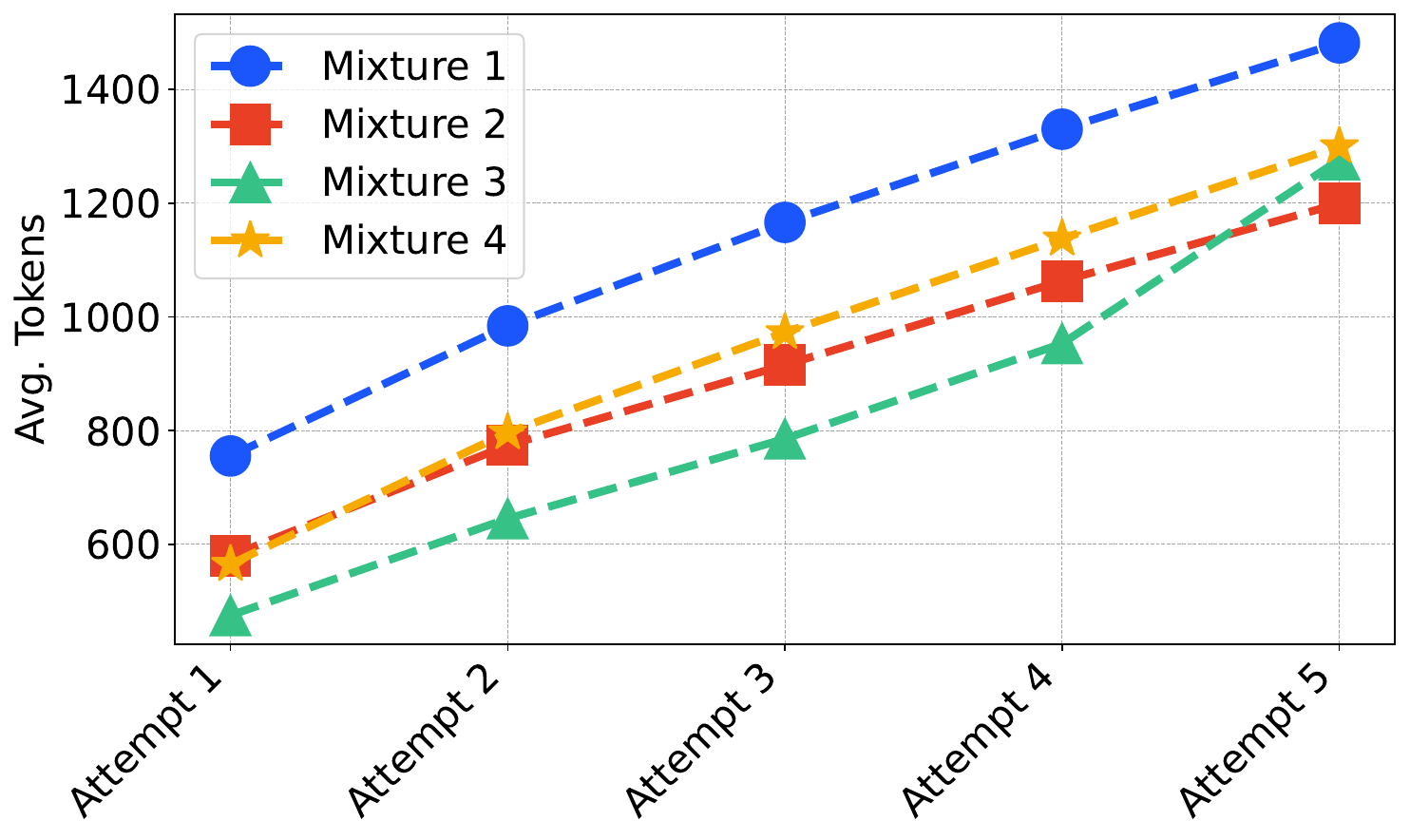}
    \caption{RewardMATH (Avg. Tokens)}
    \label{fig:algorithms_ablation/rewardmath/plot_exp3_data_mixture_token_num}
  \end{subfigure}
  
  \caption{Data mixture ablation on RewardBench and RewardMATH.}
  \label{fig:data_mixture_ablation_2}
\end{figure}

\subsection{Decision Change During STTS}

\begin{figure}[htbp!]
  \centering
  \begin{subfigure}[b]{0.48\textwidth}
    \centering
    \includegraphics[width=\linewidth]{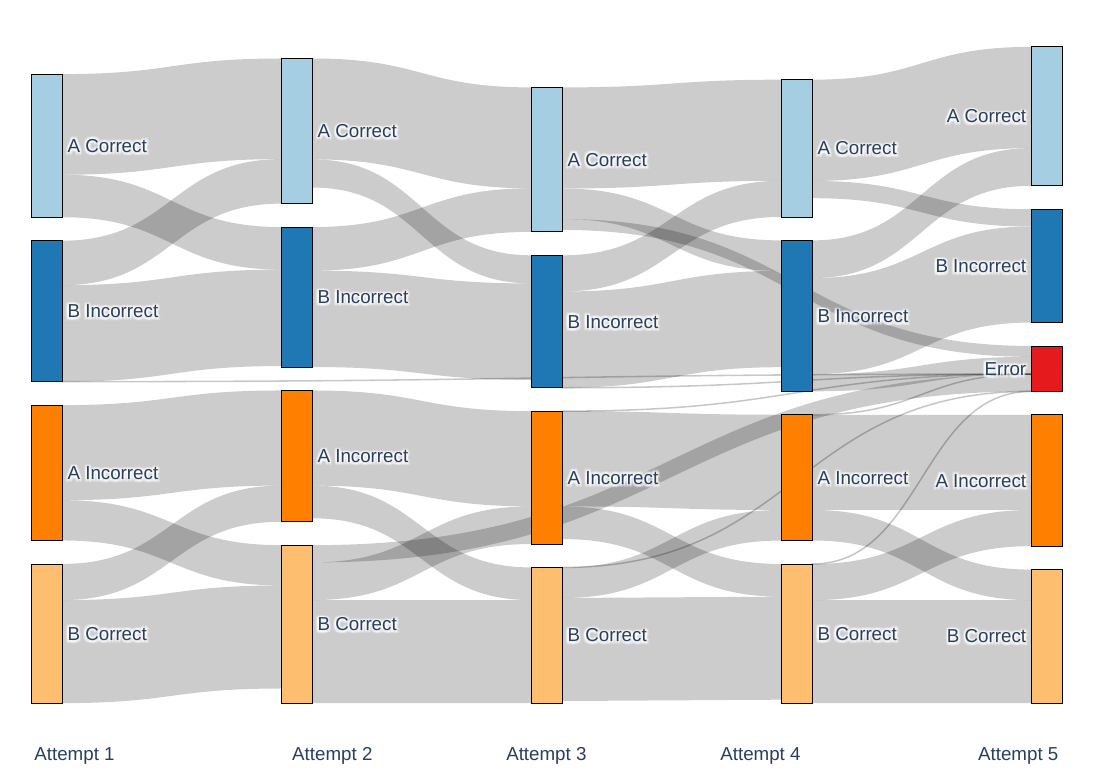}
    \caption{Anthropic Harmless}
    \label{appendix:sankey_1}
  \end{subfigure}
  \hfill
  \begin{subfigure}[b]{0.48\textwidth}
    \centering
    \includegraphics[width=\linewidth]{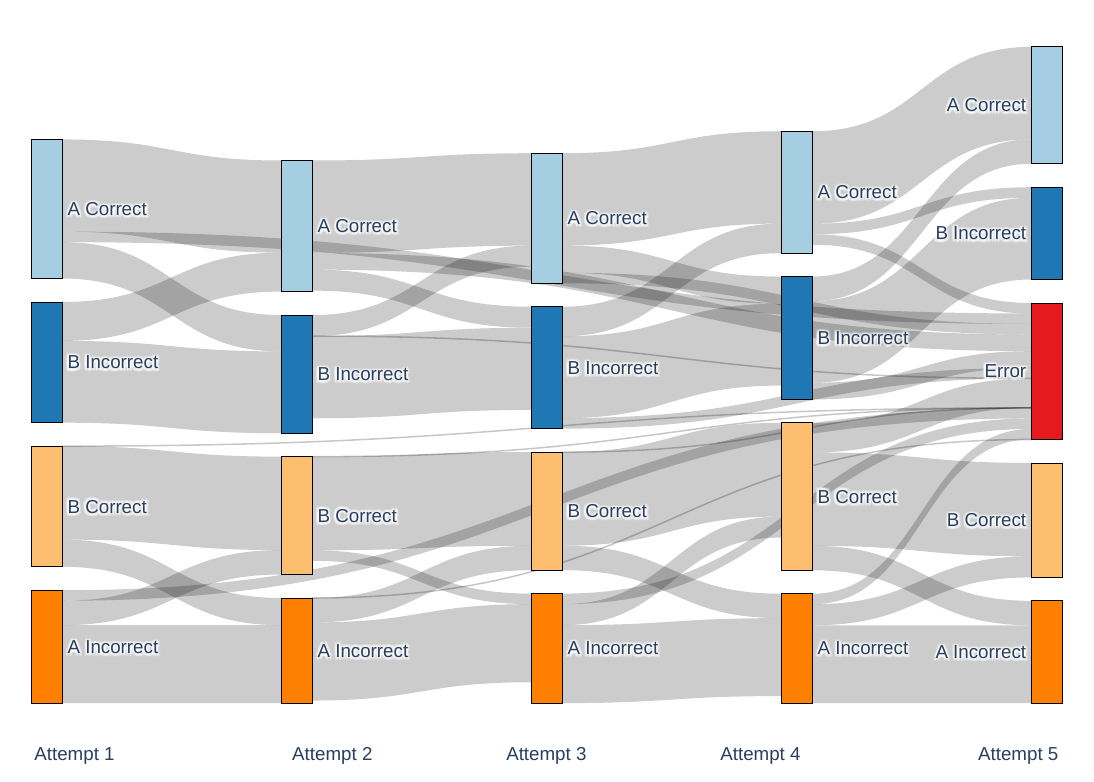}
    \caption{CodePrefBench}
    \label{appendix:sankey_2}
  \end{subfigure}
  
  \vspace{0.5cm} 
  
  \begin{subfigure}[b]{0.48\textwidth}
    \centering
    \includegraphics[width=\linewidth]{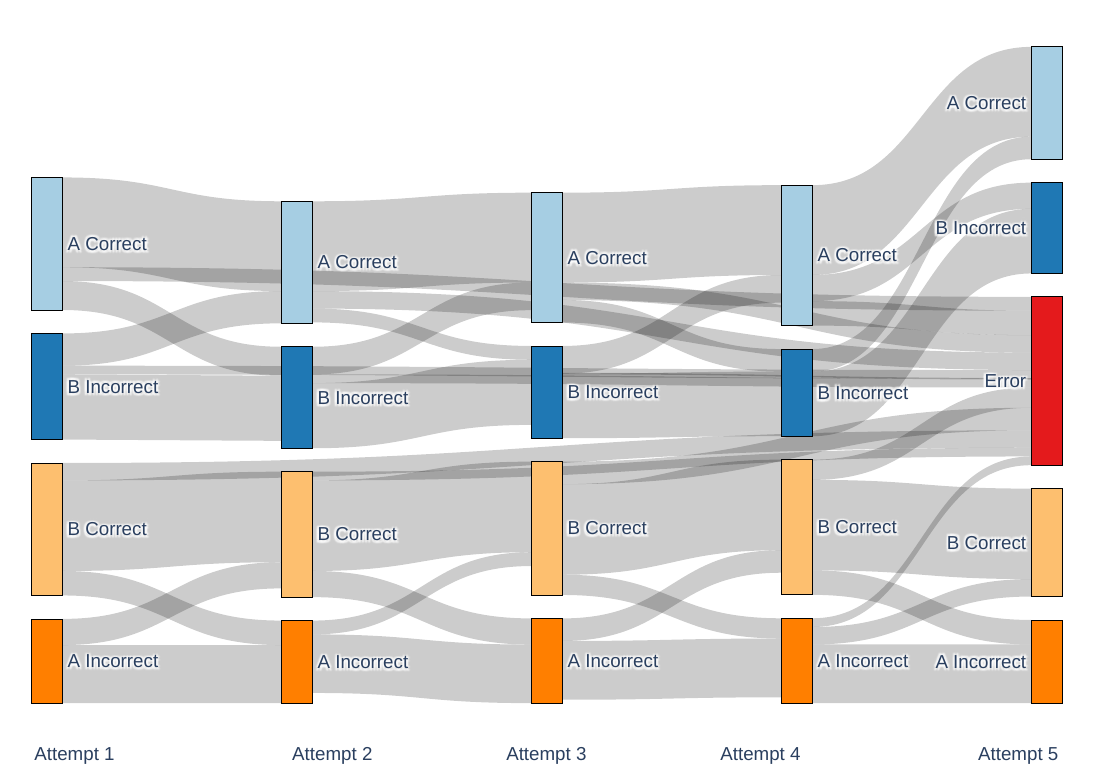}
    \caption{RewardBench}
    \label{appendix:sankey_3}
  \end{subfigure}
  \hfill
  \begin{subfigure}[b]{0.48\textwidth}
    \centering
    \includegraphics[width=\linewidth]{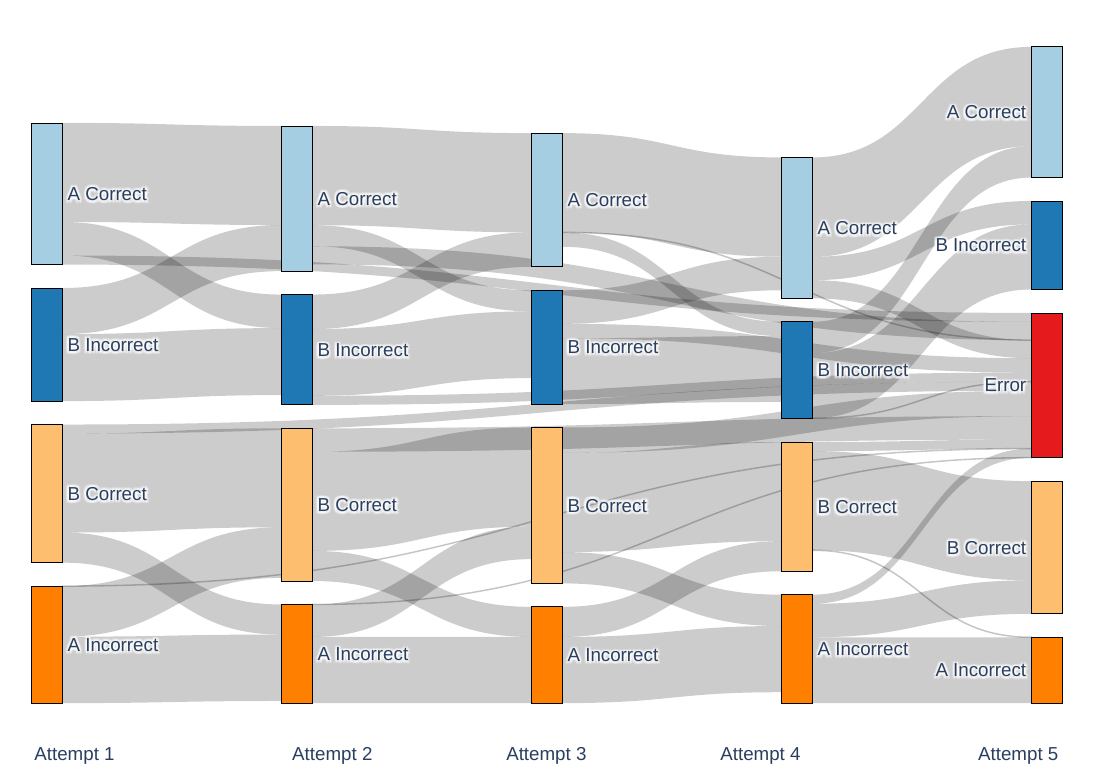}
    \caption{RewardMATH}
    \label{appendix:sankey_4}
  \end{subfigure}
  
  \caption{Decision change during STTS.}
  \label{fig:sankey}
\end{figure}

We visualize how the decision-making of our model evolves with each additional "wait" step across four benchmark tasks. Because the proportion of responses altered after adding each "wait" token is relatively small (approximately 3\%, as detailed in Appendix~\ref{appendix:statistics_during_rejection}), we apply logarithmic scaling to clearly illustrate these changes in the Sankey diagrams. As depicted in Figure~\ref{fig:sankey}, we observe a notable pattern: each incremental "wait" step results in some responses shifting from incorrect to correct, while simultaneously causing other responses to shift from correct to incorrect. The net performance improvement arises because the proportion of responses transitioning from incorrect to correct outweighs those moving in the opposite direction. This indicates that even originally correct responses are vulnerable to becoming incorrect upon further reflection. Additionally, another source of error emerges when the model fails to produce parseable outputs after additional reflection steps. We suggest that future work aimed at enhancing the effectiveness of STTS should specifically address these instances where initially correct responses become incorrect after reflection.

\subsection{Reflective Words Frequency on Test Sets}
\label{appendix:reflective_words_frequency}

\begin{figure}[htbp!]
  \centering
  \begin{subfigure}[b]{0.48\textwidth}
    \centering
    \includegraphics[width=\linewidth]{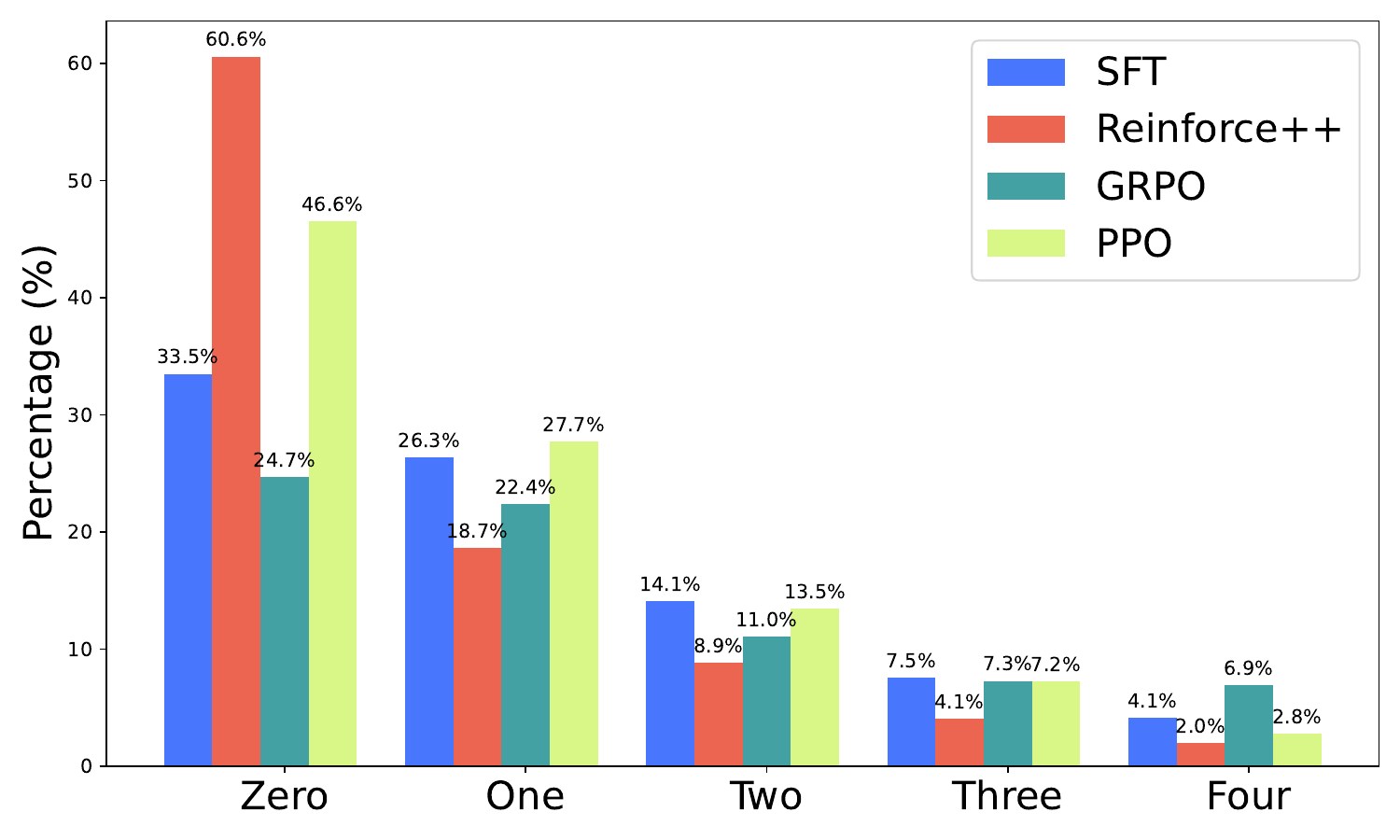}
    \caption{Anthropic Harmless}
    \label{appendix:wait_freq_41}
  \end{subfigure}
  \hfill
  \begin{subfigure}[b]{0.48\textwidth}
    \centering
    \includegraphics[width=\linewidth]{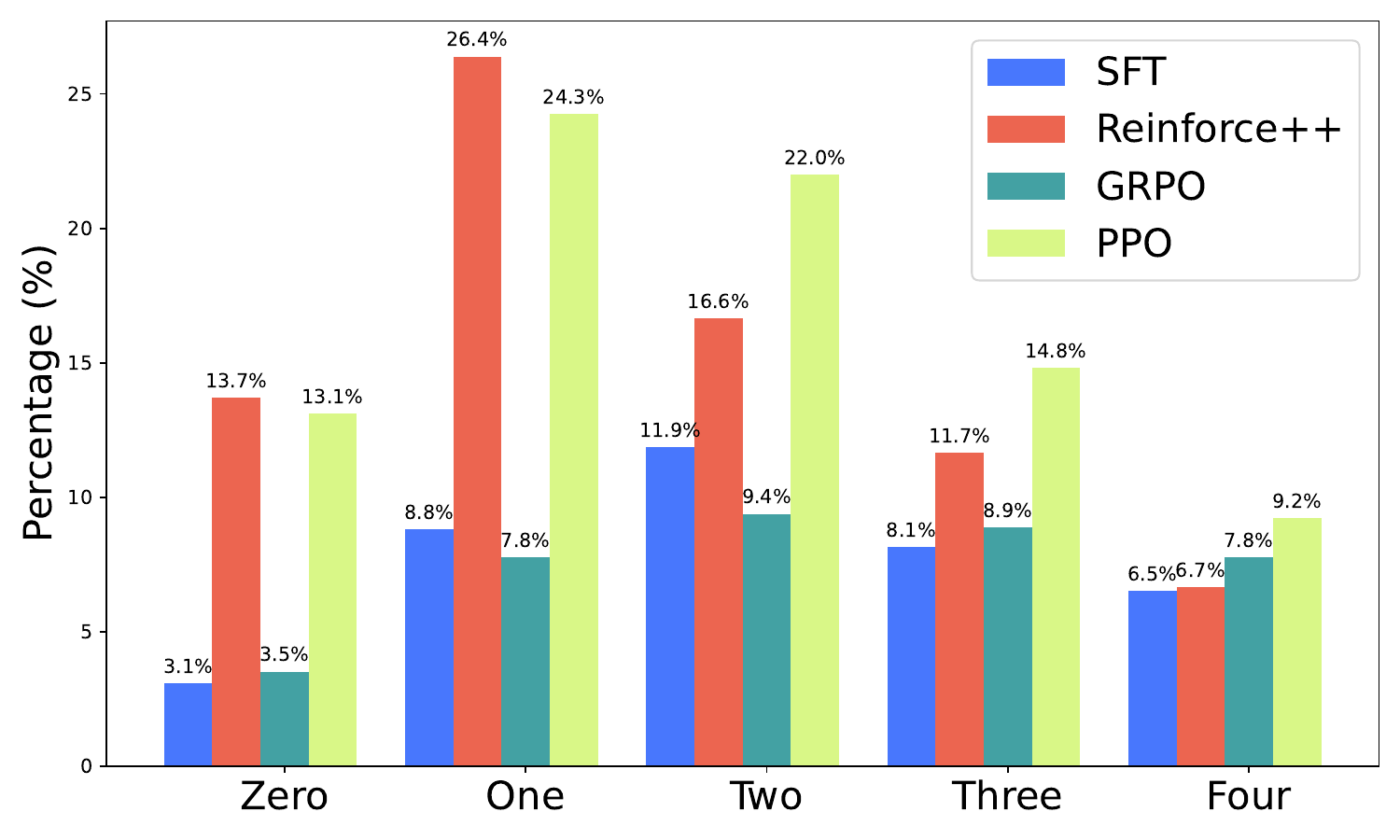}
    \caption{CodePrefBench}
    \label{appendix:wait_freq_42}
  \end{subfigure}
  
  \vspace{0.5cm} 
  
  \begin{subfigure}[b]{0.48\textwidth}
    \centering
    \includegraphics[width=\linewidth]{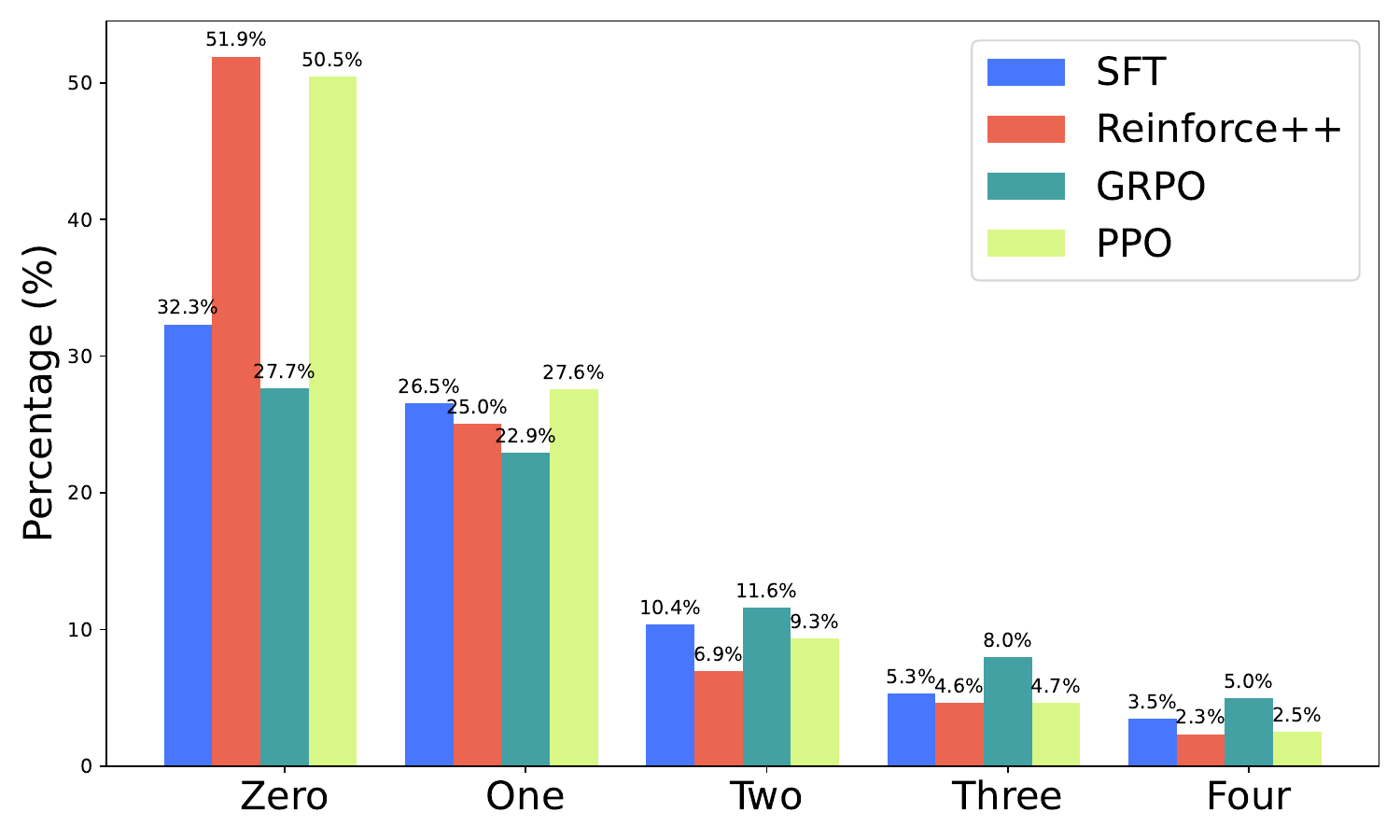}
    \caption{RewardBench}
    \label{appendix:wait_freq_43}
  \end{subfigure}
  \hfill
  \begin{subfigure}[b]{0.48\textwidth}
    \centering
    \includegraphics[width=\linewidth]{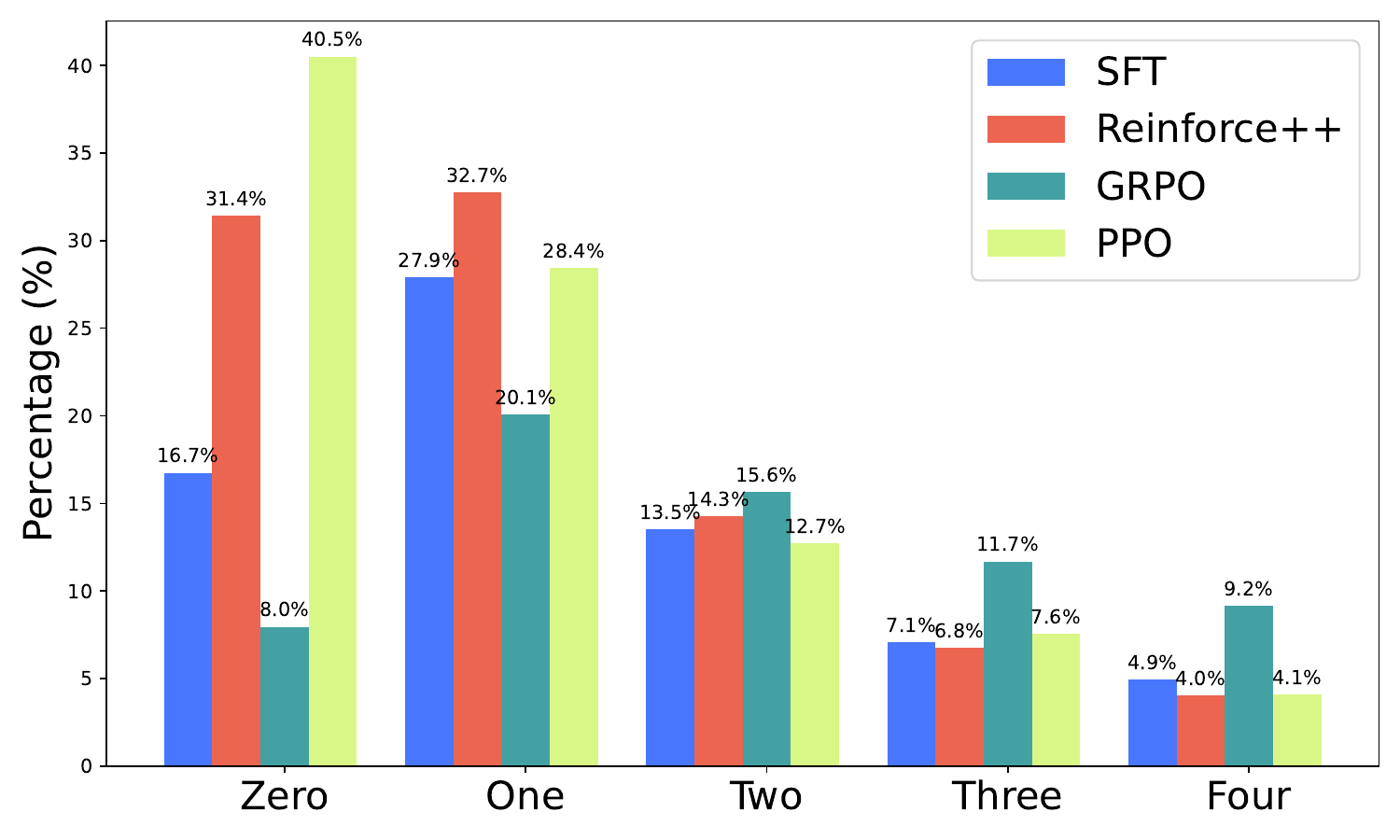}
    \caption{RewardMATH}
    \label{appendix:wait_freq_44}
  \end{subfigure}
  
  \caption{Reflective words frequency on four benchmark.}
  \label{appendix:wait_freq_testset}
\end{figure}

Following the experimental setup described in Section~\ref{appendix:other_statistics}, we analyzed the proportion of instances within the test set that employ varying numbers of reflective words, categorizing results according to the RL algorithms used during training. Our key findings are as follows:

\begin{enumerate}
    \item Models trained with GRPO exhibit a stronger tendency to utilize more reflective words, which aligns with our earlier observation in Section~\ref{subsec:different_rl_algorithms}, indicating that GRPO-trained models generally produce longer reasoning sequences.
    \item Interestingly, we observed that models trained solely through SFT are not incapable of reflection; rather, to our surprise, these SFT-only models actually tend to employ reflective tokens more frequently. Aggregating data from Table~\ref{appendix:wait_freq_testset} reveals that, on average, only about 68\% of instances generated by SFT models use 0 to 4 reflective tokens, indicating the remaining 32\% use significantly more reflective tokens. This proportion substantially exceeds that of RL-trained models. However, as evidenced by Table~\ref{tab:main}, increased use of reflective words does not correlate with improved performance—on the contrary, it tends to degrade results, suggesting that the additional reflection is often ineffective. Further case studies in Appendix~\ref{appendix:case_study} reveal that excessive use of reflective tokens frequently leads the model into infinite reasoning loops, whereas RL-trained models successfully learn to use reflective tokens in a more effective and controlled manner.
\end{enumerate}

\subsection{Model Performance Across Different Reflective Counts}

\begin{figure}[htbp!]
  \centering
  \begin{subfigure}[b]{0.48\textwidth}
    \centering
    \includegraphics[width=\linewidth]{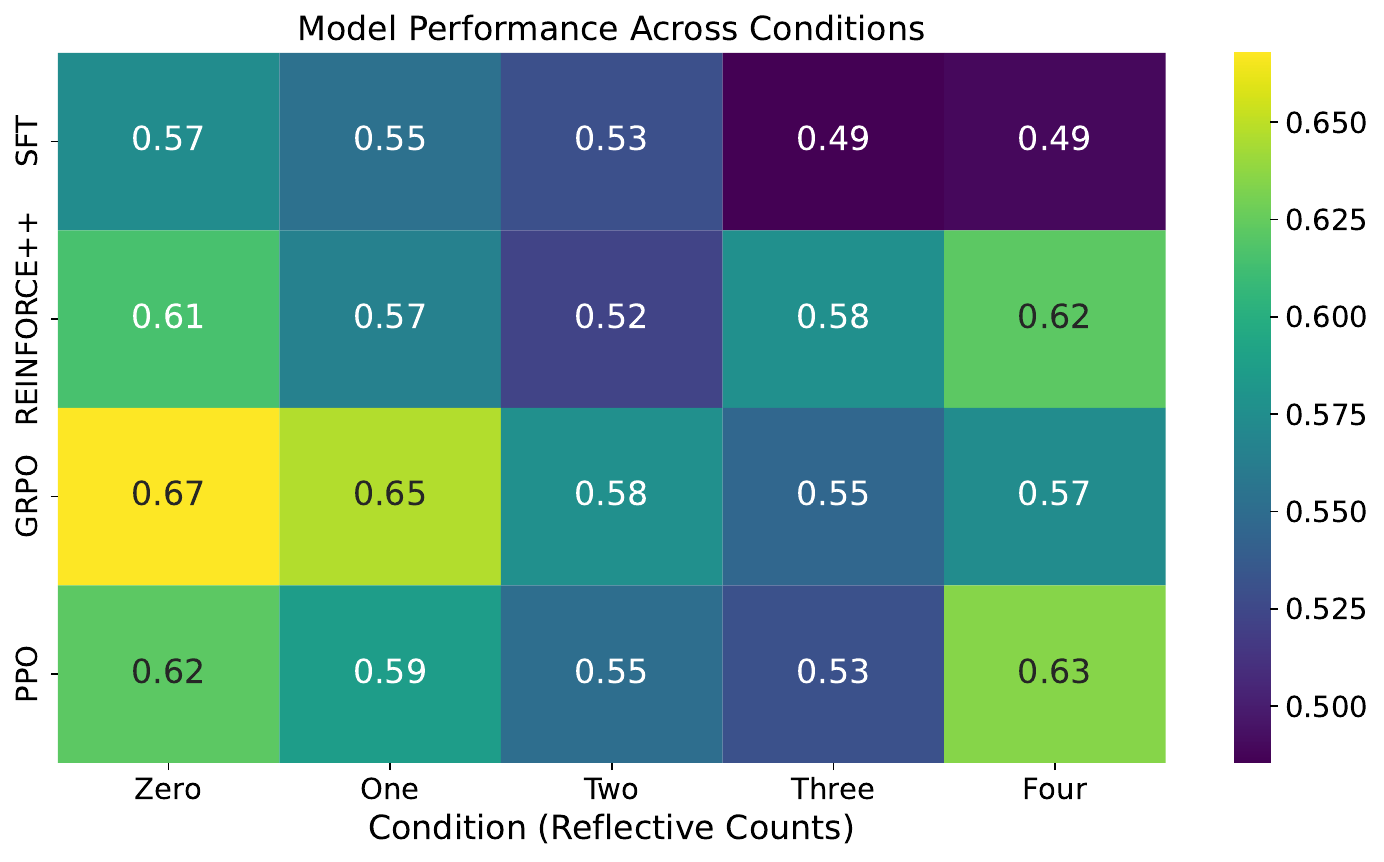}
    \caption{Anthropic Harmless}
    \label{appendix:wait_heatmap_1}
  \end{subfigure}
  \hfill
  \begin{subfigure}[b]{0.48\textwidth}
    \centering
    \includegraphics[width=\linewidth]{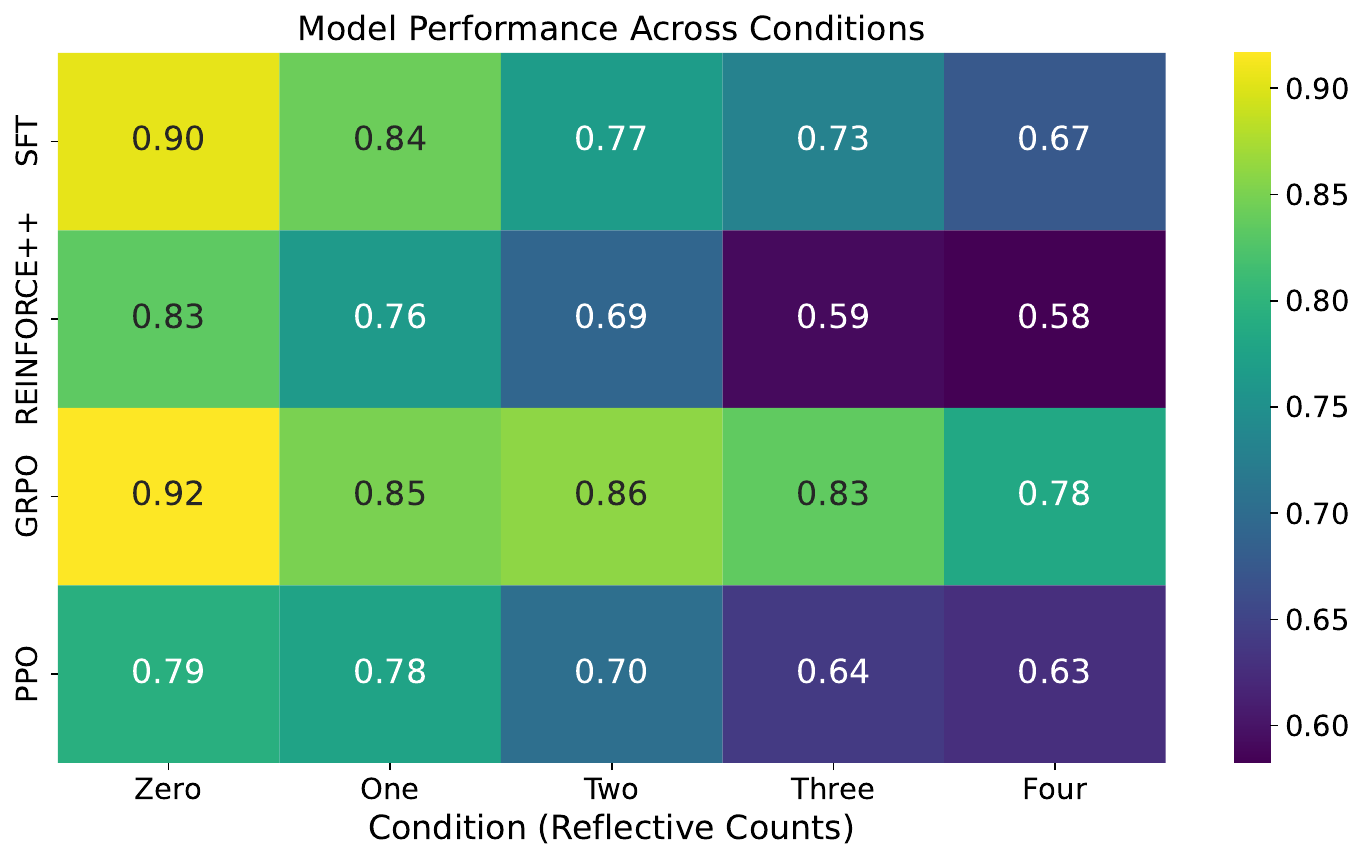}
    \caption{CodePrefBench}
    \label{appendix:wait_heatmap_2}
  \end{subfigure}
  
  \vspace{0.5cm} 
  
  \begin{subfigure}[b]{0.48\textwidth}
    \centering
    \includegraphics[width=\linewidth]{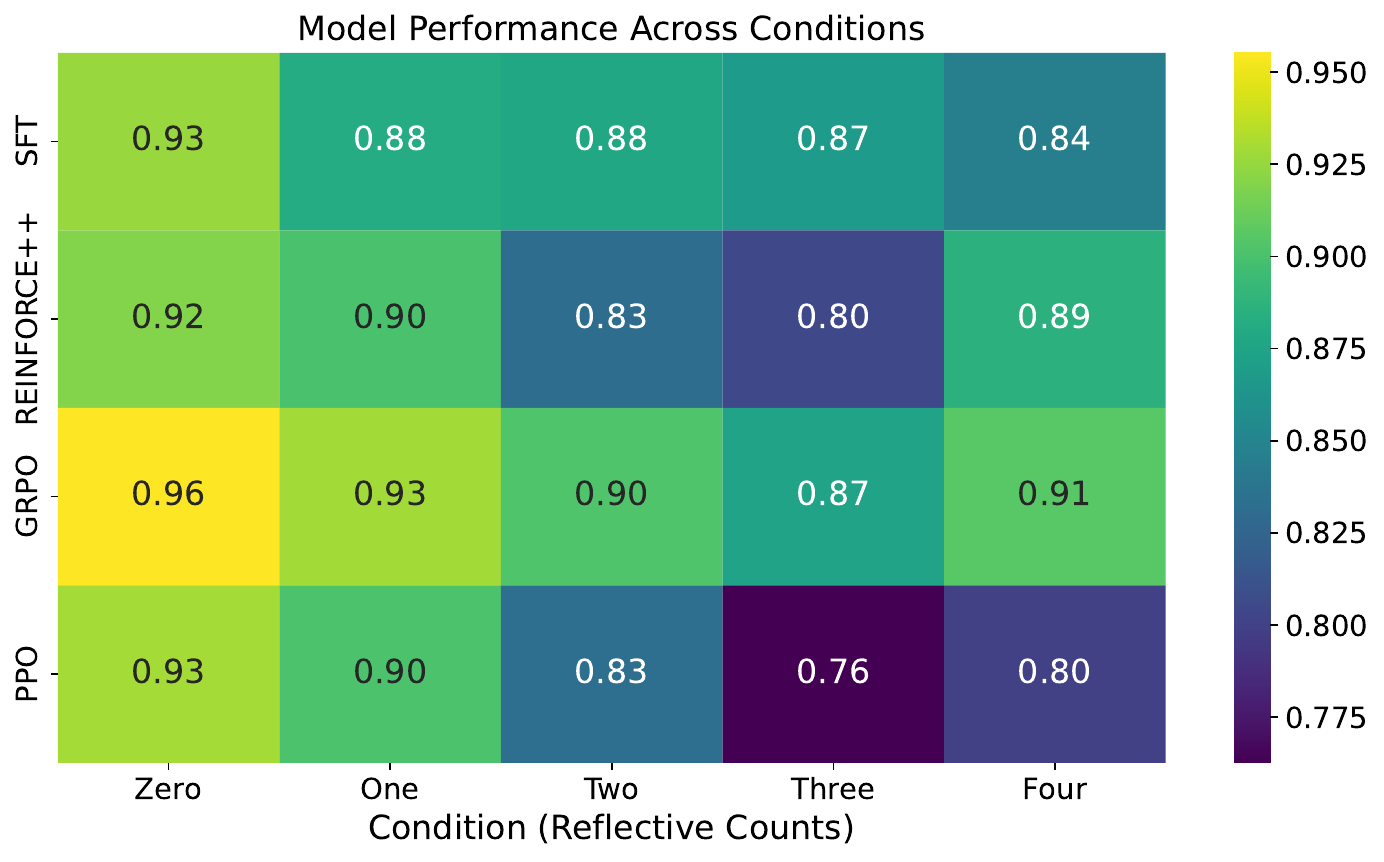}
    \caption{RewardBench}
    \label{appendix:wait_heatmap_3}
  \end{subfigure}
  \hfill
  \begin{subfigure}[b]{0.48\textwidth}
    \centering
    \includegraphics[width=\linewidth]{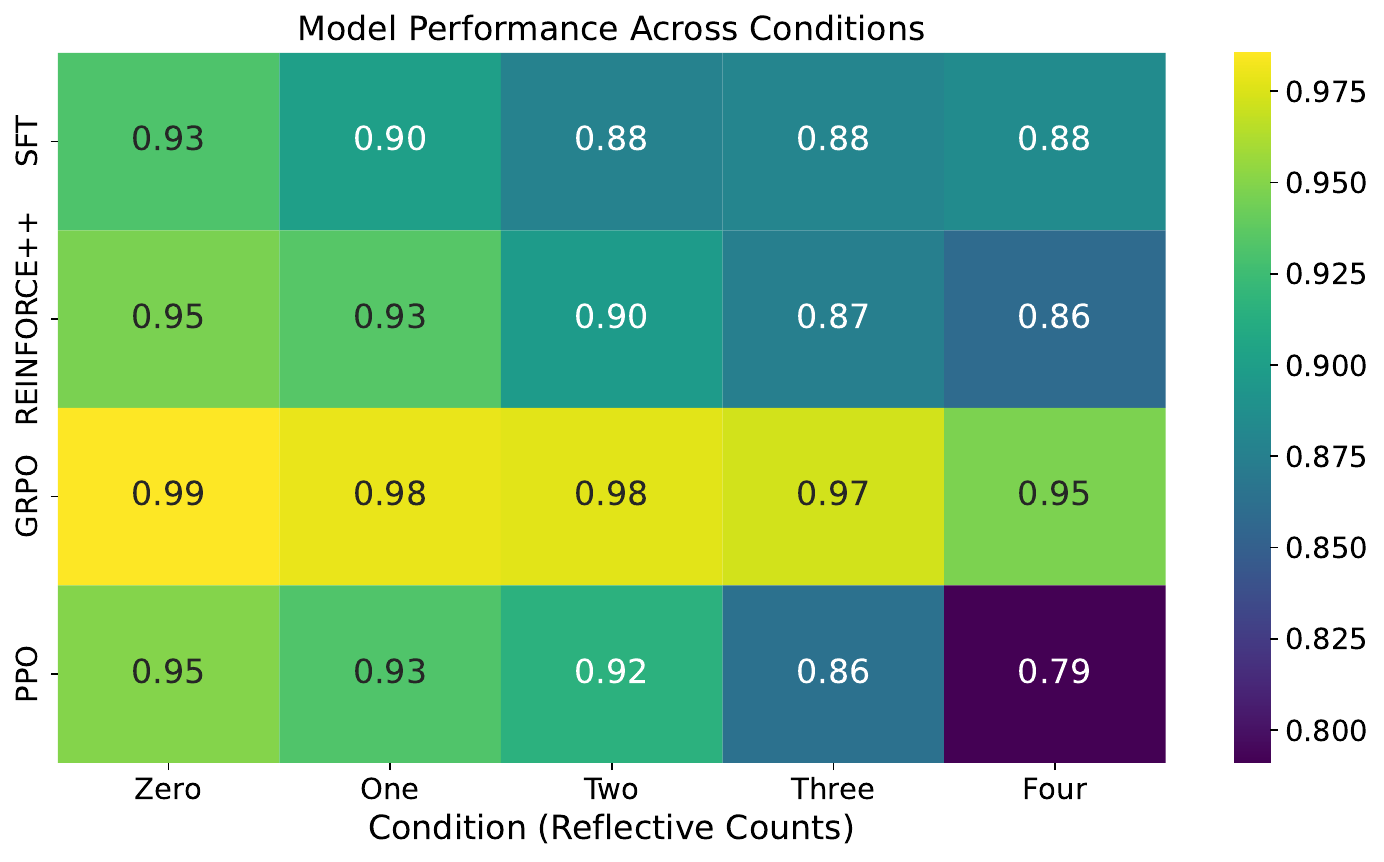}
    \caption{RewardMATH}
    \label{appendix:wait_heatmap_4}
  \end{subfigure}
  
  \caption{Performance conditioned on differenct reflective counts.}
  \label{appendix:wait_heatmap}
\end{figure}

We analyze the model’s accuracy across scenarios with varying reflective token counts, visualized in Figure~\ref{appendix:wait_heatmap}. Higher accuracy corresponds to lighter shades on the left side of the heatmap, indicating that the model generally achieves better performance when it uses no reflective tokens or only a few. Conversely, darker shades toward the right side of the heatmap reflect lower accuracy for instances involving greater use of reflective tokens. This observation is both intuitive and insightful; it suggests that problems solvable without extensive reflection naturally yield higher accuracy, whereas instances prompting the model to engage in more extensive reflection are inherently more challenging or ambiguous. The model's spontaneous inclination toward greater reflection in these difficult instances highlights its inherent capability and significant potential for addressing complex reasoning tasks.

\section{Case Study}
\label{appendix:case_study}

In this section, we present case studies to provide users with deeper insights into the model's behavior. Specifically, we categorize these case studies into several distinct groups: 
\begin{itemize}
    \item[$\bullet$] \textbf{Initial Wrong $\rightarrow$ Correct after Reflection} (Appendix~\ref{appendix:initial_wrong})
    \item[$\bullet$] \textbf{Initial Correct $\rightarrow$ Wrong after Reflection} (Appendix~\ref{appendix:initial_correct})
    \item[$\bullet$] \textbf{Consistently Correct} (Appendix~\ref{appendix:consistently_correct})
    \item[$\bullet$] \textbf{Consistently Wrong} (Appendix~\ref{appendix:consistently_wrong})
\end{itemize}

Detailed examples for each category are provided in the respective appendices.

\subsection{Initial Wrong --> Correct after reflection}
\label{appendix:initial_wrong}

In this section, we illustrate how our model utilizes reflection to reconsider initially incorrect answers and subsequently arrive at correct conclusions. Specifically, in Figure~\ref{fig:case 1}, \textbf{J1-7B} initially judges both answers as incorrect and mistakenly identifies an irrelevant detail within Answer B’s reasoning as erroneous. This reveals a tendency for the model's initial responses to be overly critical or overly focused on minor details. However, after one step of STTS, the model correctly recognizes that Answer B’s computation is accurate, leading to a revised judgment in favor of Answer B.

In contrast, Figure~\ref{fig:case 2} demonstrates a scenario where, despite ultimately selecting the correct answer, the model exhibits a notable phenomenon known as \textit{unfaithful reasoning}~\citep{arcuschin2025chain}. During its CoT reasoning, the model initially maintains that Answer B’s refusal to respond is safer, whereas Answer A might introduce potential risks. Surprisingly, at the final decision-making step, the model abruptly shifts, concluding that Answer A is preferable when considering overall criteria. Although this decision appears as implicit post-hoc rationalization, we argue that by monitoring the reasoning process and applying STTS to reveal extended thought processes, the model gradually acknowledges that Answer A, despite potential pitfalls, offers valid solutions compared to Answer B’s outright refusal. Thus, the final shift in judgment gains interpretability and rationale. It should also be noted that this instance of \textit{unfaithful reasoning} represents an uncommon occurrence selected deliberately from numerous cases.

\begin{figure}[htbp]
    \centering
    \includegraphics[width=\textwidth]{figs/case_study/case_1.pdf}
    \caption{Case study 1 for Initial Wrong $\rightarrow$ Correct after reflection.}
    \label{fig:case 1}
\end{figure}

\begin{figure}[htbp]
    \centering
    \includegraphics[width=\textwidth]{figs/case_study/case_2.pdf}
    \caption{Case study 2 for Initial Wrong $\rightarrow$ Correct after reflection.}
    \label{fig:case 2}
\end{figure}

\subsection{Initial Correct --> Wrong after reflection}
\label{appendix:initial_correct}

In this section, we present case studies illustrating how erroneous reflections by the model can cause initially correct answers to become incorrect. Figures~\ref{fig:case 3} and \ref{fig:case 4} highlight instances where the main reason for incorrect judgments post-reflection is a shift in evaluation criteria. For example, the model changes its focus from prioritizing \textit{creativity} to emphasizing \textit{conciseness}, or from valuing \textit{informativeness} to \textit{transparency}. We argue that these errors are not due to deficiencies in the model's reasoning capability, but rather stem from ambiguity and inconsistency in evaluation criteria. Future work can target clarifying and stabilizing these criteria to further enhance the potential of \lmj frameworks.

\begin{figure}[htbp]
    \centering
    \includegraphics[width=\textwidth]{figs/case_study/case_3.pdf}
    \caption{Case study 1 for Initial Correct $\rightarrow$ Wrong after reflection.}
    \label{fig:case 3}
\end{figure}

\begin{figure}[htbp]
    \centering
    \includegraphics[width=\textwidth]{figs/case_study/case_4.pdf}
    \caption{Case study 2 for Initial Correct $\rightarrow$ Wrong after reflection.}
    \label{fig:case 4}
\end{figure}

\subsection{Consistently Correct}
\label{appendix:consistently_correct}

In this section, we showcase cases where the model consistently provides correct answers. We observe that reflection significantly strengthens the model's confidence, indicating a high level of certainty and robustness in these instances. Please refer to Figures~\ref{fig:case 5} and ~\ref{fig:case 6}.

\begin{figure}[htbp]
    \centering
    \includegraphics[width=\textwidth]{figs/case_study/case_5.pdf}
    \caption{Case study 1 for Consistently Correct.}
    \label{fig:case 5}
\end{figure}

\begin{figure}[htbp]
    \centering
    \includegraphics[width=\textwidth]{figs/case_study/case_6.pdf}
    \caption{Case study 2 for Consistently Correct.}
    \label{fig:case 6}
\end{figure}

\subsection{Consistently Wrong}
\label{appendix:consistently_wrong}

Additionally, we examine cases where the model initially provides incorrect answers and remains incorrect even after reflection. In Figure~\ref{fig:case 7}, we observe that the model incorrectly favors a concise response (Answer B) over a more detailed yet accurate response (Answer A). Despite the reflection process, the model persists in its misjudgment due to an overly simplistic interpretation of evaluation criteria, emphasizing brevity at the expense of comprehensive correctness. This indicates a limitation in the current reflection strategy, suggesting the necessity for enhanced guidance in evaluation criteria to prevent superficial or overly rigid assessments, as is also mentioned in Appendix~\ref{appendix:initial_correct}.

\begin{figure}[htbp]
    \centering
    \includegraphics[width=\textwidth]{figs/case_study/case_7.pdf}
    \caption{Case study 1 for Consistently Wrong.}
    \label{fig:case 7}
\end{figure}

\section{Prompt Template}
\label{appendix:prompt_template}

Figure~\ref{fig:prompt_template} shows the prompt template that we use for the evaluation. By default, we use it for all the models we evaluate in this paper. It is derived from RewardBench~\citep{lambert2024rewardbench}.

\begin{figure}[htbp]
    \centering
    \includegraphics[width=\textwidth]{figs/prompt_template.pdf}
    \caption{Prompt template used for evaluation.}
    \label{fig:prompt_template}
\end{figure}